\documentclass{article}

\PassOptionsToPackage{numbers, compress}{natbib}

\usepackage[preprint]{neurips_2026}

\usepackage[utf8]{inputenc}
\usepackage[T1]{fontenc}
\usepackage{hyperref}
\usepackage{url}
\usepackage{booktabs}
\usepackage{amsfonts}
\usepackage{nicefrac}
\usepackage{microtype}
\usepackage{graphicx}
\usepackage{float}
\usepackage{subcaption}
\usepackage{multirow}
\usepackage{amsmath}
\usepackage{amssymb}
\usepackage{mathtools}
\usepackage{amsthm}
\usepackage[capitalize,noabbrev]{cleveref}

\usepackage[dvipsnames]{xcolor}
\usepackage{adjustbox}
\usepackage{xspace}
\usepackage{tikz}
\usepackage{pgfplots}
\pgfplotsset{compat=1.18}

\usepackage[skip=3pt]{caption}
\usepackage{dsfont}
\usepackage{etoolbox}
\usepackage{color}
\usepackage{amsmath}
\usepackage[normalem]{ulem}

\newif\ifshowedits

\newcommand{\addeditor}[3]{%
  \definecolor{#1color}{rgb}{#3}
  \expandafter\newcommand\csname #1\endcsname[1]{%
  \ifshowedits
    {\color{#1color} ##1}%
  \else
    {##1}%
  \fi
  }%
  \expandafter\newcommand\csname #1rmk\endcsname[1]{%
  \ifshowedits
    {\color{#1color} {\bf [#2: ##1]}}
  \fi
  }%
  \expandafter\newcommand\csname #1rpl\endcsname[2]{%
  \ifshowedits
    {\color{#1color} ##1 \sout{##2}}
  \else
    {##1}
  \fi
  }%
}

\newcommand{\createtextvar}[1]{
  \expandafter\newcommand\csname #1\endcsname{%
  {\text{#1}}
}%
}
\newcommand{\mycomment}[1]{}

\newcommand{\calL}{{\cal L}}

\newcommand{\bp}{{\bf p}}

\newcommand{\vcomment}[1]{}

\addeditor{vincent}{VL}{0.0, 0.5, 0.0}
\addeditor{thibaut}{TL}{1., 0.0, 1.}
\addeditor{guillaume}{GB}{1., 0.4, 0.0}
\showeditsfalse

\def\ourmethod{Gekko\xspace}

\definecolor{mygreen}{RGB}{182,215,168}
\definecolor{myorange}{RGB}{249,203,156}
\definecolor{mygrey}{RGB}{102,102,102}

\def\m#1{\ensuremath{\mathtt{#1}}}

\hypersetup{hidelinks}

\title{Revisiting Cross-View Completion: Self-Supervised\\Pre-Training via Reconstruction Error Comparison}

\theoremstyle{plain}

\theoremstyle{definition}

\theoremstyle{remark}

\author{
  Thibaut Loiseau$^{1}$ 
  \hspace{20pt}
  Guillaume Bourmaud$^{2}$
  \hspace{20pt}
  Vincent Lepetit$^{1}$  \\
  $^{1}$ LIGM, Ecole Nationale des Ponts et Chaussées, IP Paris, Univ Gustave Eiffel, CNRS, France \\
  $^{2}$ Univ. Bordeaux, CNRS, Bordeaux INP, IMS, UMR 5218, France \\
  {\tt\small \{thibaut.loiseau,vincent.lepetit\}@enpc.fr}
  \hspace{5pt}
  {\tt\small guillaume.bourmaud@u-bordeaux.fr}
}

\begin{document}

\maketitle
\vspace{-0.60cm}
\begin{figure}[h]
    \centering
    \input{figures/teaser/teaser}
\end{figure}

\begin{abstract}
  \vspace{-0.35cm}
  Self-supervised pre-training via cross-view completion learns strong features for 3D vision from co-visible regions of image pairs. However, the reference view provides little information for reconstructing non-co-visible patches, implicitly yielding a monocular training signal in these regions.
We introduce \textbf{\ourmethod}, which turns this limitation into a useful signal. The \emph{relative improvement} of the cross-view reconstruction error over a masked-autoencoder error is a self-supervised proxy for co-visibility: large improvements indicate co-visible regions, negligible ones non-co-visible areas. \ourmethod is a network, trained from scratch, that jointly performs cross-view completion, masked autoencoding, and per-pixel prediction of this relative improvement, providing an additional binocular signal for \emph{all} masked regions without any ground-truth 3D annotation.
Under identical architectures and training data, \ourmethod consistently outperforms CroCo on zero-shot correspondence estimation, relative pose estimation, and pointmap regression, with up to $6\times$ higher accuracy at the strictest relative-pose threshold and a $22\%$ drop in end-point error on ETH3D. The extra channel it learns is itself a strong co-visibility detector on unseen scenes, and \ourmethod's frozen features outperform released cross-view backbones of comparable or larger size. It can also be trained directly from raw videos with a simple stride-based curriculum, removing the cumbersome 3D preprocessing prior methods require while matching models trained on curated data. Code and pre-trained models are publicly available.

\end{abstract}

\section{Introduction}

Learning meaningful representations for 3D vision from visual data is a fundamental challenge in computer vision~\cite{chen2020simple,grill2020bootstrap,caron2021emerging,bao2022beit,he2022masked,zhou2022image,oquab2024dinov2,yang2025pursuit}, with applications spanning depth estimation~\cite{yang2024depth,lin2025depth}, visual localization~\cite{dong2024reloc3r}, and 3D reconstruction~\cite{wang2024dust3r,elflein2025light3r,yang2025fast3r,wang20243d}. Self-supervised pre-training has emerged as a powerful paradigm for acquiring these representations without expensive manual annotations. Among recent approaches, cross-view completion~\cite{weinzaepfel2022croco} extends Masked Autoencoders~(MAE)~\cite{he2022masked} to image pairs: patches in a target image are masked and reconstructed using both the remaining target patches and a reference view. The underlying principle is that reconstructing co-visible regions forces the network to establish correspondences and reason about geometry, yielding features that transfer effectively to downstream 3D tasks~\cite{an2025cross,wang2024dust3r,leroy2024grounding}.

However, cross-view completion faces a critical limitation. 
When a masked patch is not co-visible (\textit{e.g.}, due to occlusion), the reference view provides little information in those regions, implicitly yielding a MAE-like monocular training signal.
Consequently, cross-view completion pre-training does not benefit from image pairs with low overlap, where many masked patches lack a corresponding region in the reference view~\cite{weinzaepfel2022croco,loiseau2025alligat0r}.

We propose~\textbf{\ourmethod}, a fully self-supervised pre-training framework that turns this limitation into a strength. Our key insight is that the \emph{relative improvement} of the cross-view completion reconstruction error over the MAE reconstruction error is a reliable proxy for co-visibility (\Cref{fig:teaser}). In co-visible regions, incorporating the reference image substantially improves reconstruction quality, yielding a large relative improvement. In non-co-visible regions, both reconstructions perform similarly, and the relative improvement is near zero. This proxy requires no ground-truth depth maps or camera poses, making it applicable to any collection of image pairs, including uncalibrated data.

\ourmethod is a network that jointly performs three tasks during pre-training: (i) cross-view completion, (ii) masked autoencoding, and (iii) prediction of the relative improvement map. The architecture is identical to CroCo~\cite{weinzaepfel2022croco} up to a single additional output channel, enabling a direct and fair comparison. The relative improvement prediction provides an additional dense binocular self-supervised signal for \emph{all} masked regions. The cost of this third task is negligible: the encoder and the cross-attention decoder are byte-for-byte those of CroCo, and only the lightweight pixel head grows from three to four channels, so pre-training, fine-tuning and inference all keep CroCo's architecture.

Under identical architectures and training data, \ourmethod consistently outperforms CroCo across three downstream tasks: up to $6\times$ higher accuracy at the strictest threshold on relative metric pose estimation (ScanNet-1500), a 22\% drop in end-point error on zero-shot correspondence estimation (ETH3D), and lower error on pointmap regression, with gains that scale to larger architectures. We also enable fully self-supervised pre-training from raw video with a simple stride-based curriculum, removing the expensive 3D preprocessing prior methods require~\cite{weinzaepfel2022croco,loiseau2025alligat0r} while matching models trained on curated data. The extra channel is also directly interpretable: probed on its own, it detects co-visibility on scenes never seen during pre-training, and does so better than the cross-view reconstruction error it is derived from (\Cref{sec:covis_probe}).

Our contributions are as follows:
\begin{itemize}
    \item We show that the relative improvement of a cross-view completion error over a masked autoencoding error is a reliable self-supervised proxy for co-visibility, substantially outperforming the cross-view completion error alone, and we propose~\ourmethod, a network trained from scratch that jointly learns cross-view completion, masked autoencoding and relative improvement prediction. The relative improvement loss supplies a binocular signal for \emph{all} masked regions without any ground-truth 3D labels.
    \item We demonstrate that~\ourmethod consistently outperforms CroCo on three cross-view 3D vision tasks under identical architectures and training data, that its predicted channel is itself a strong co-visibility detector on scenes unseen during pre-training, and that its frozen features outperform released cross-view backbones of comparable or larger size.
    \item We show that~\ourmethod can be trained from raw videos with a simple stride-based curriculum, removing the need for overlap-based data preprocessing and matching the performance of models trained on curated data, enabling fully self-supervised pre-training.
\end{itemize}

\section{Related work}\label{sec:related_work}

\paragraph{Self-supervised visual representation learning.}
Masked image modeling~\cite{he2022masked,bao2022beit,zhou2022image} and self-distillation~\cite{caron2021emerging,oquab2024dinov2,assran2023self} have become dominant paradigms for learning visual features without labels. MAE~\cite{he2022masked} reconstructs randomly masked patches and learns rich visual features, while DINO~\cite{caron2021emerging} and I-JEPA~\cite{assran2023self} learn representations through self-distillation in pixel or latent space. DINOv2~\cite{oquab2024dinov2} combines both strategies at scale. These methods operate on single images and therefore do not explicitly capture multi-view geometric relationships. Our work builds on masked image modeling but extends it to image pairs, where the comparison between monocular and binocular reconstruction errors provides a geometric training signal absent from single-image methods.

\paragraph{Pre-training on image pairs.}
CroCo~\cite{weinzaepfel2022croco} pioneered cross-view completion for self-supervised pre-training of 3D vision tasks by extending masked image modeling to image pairs. CroCo v2~\cite{weinzaepfel2023croco} scaled this framework to large and diverse training data, demonstrating strong transfer to stereo matching and optical flow. P-Match~\cite{zhu2023pmatch} introduces a variant where both images are partially masked to pre-train an image matching model. Other variants include masked appearance transfer for object tracking~\cite{zhao2023representation,song2023compact}. Notably, the foundational models DUSt3R~\cite{wang2024dust3r} and MASt3R~\cite{leroy2024grounding}, and all subsequent works~\cite{yang2025fast3r,cabon2025must3r,zust2025panst3r,dong2024reloc3r,zhang2024monst3r,smart2024splatt3r,elflein2025light3r}, build upon CroCo's cross-view completion framework and have achieved remarkable success in dense 3D reconstruction and matching.
All these methods share a common limitation: the training signal is implicitly monocular for non-co-visible regions. Our method addresses this by explicitly modeling the relative improvement of cross-view completion over monocular reconstruction, providing an additional binocular signal for \emph{all} masked regions.
A parallel line of work extends cross-view completion beyond image pairs: MuM~\cite{nordstrom2025mum} masks arbitrarily many views uniformly and decodes them with inter-frame attention, and Muskie~\cite{li2025muskie} is a native multi-view backbone that learns view-invariant features by finding correspondences across views without 3D supervision. These extensions are orthogonal to our contribution (\ourmethod improves the training signal available within each pair), and combining the two is a promising direction, though not a straightforward one: with more than two views, the relative improvement might have to be defined against a \emph{set} of references rather than a single one (\Cref{sec:limitations}).

\paragraph{Co-visibility in image matching and pre-training.}
Ground-truth co-visibility masks, derived from depth maps and camera poses, are widely used in image matching methods~\cite{potje2024xfeat,edstedt2024dedode,lindenberger2023lightglue,zhao2023aliked,sun2021loftr,wang2024efficient,edstedt2024roma,chen2022aspanformer,wang2022matchformer,tangquadtree,edstedt2023dkm,fan2023occ,gleize2023silk,kim2025learning,germainvisual,jiang2021cotr,tan2022eco,sarlin2020superglue} to compute overlaps and select training pairs. Several approaches~\cite{sarlin2020superglue,edstedt2024roma,edstedt2025roma,truong2023pdc,edstedt2023dkm,germainvisual,bono2023end} also leverage these masks to learn matchability scores at test time.
Alligat0R~\cite{loiseau2025alligat0r} recently demonstrates that co-visibility prediction is an effective \emph{supervised} pre-training objective for relative pose, training on ground-truth co-visibility labels derived from depth maps and camera poses. While effective, this approach requires ground-truth 3D annotations and expensive overlap-based data preprocessing.
Our method pursues the same goal in a self-supervised manner: the relative improvement of cross-view completion over masked autoencoding serves as a proxy for co-visibility, eliminating the need for any ground-truth depth or pose information. Combined with a simple stride-based curriculum (\Cref{sec:curriculum}), \ourmethod can be trained directly from raw videos, removing the cumbersome 3D data preprocessing required by both CroCo and Alligat0R, thereby enabling fully self-supervised pre-training.
\section{Background and motivation}\label{sec:background}

\subsection{Background on CroCo and MAE}
CroCo~\cite{weinzaepfel2022croco} reconstructs a target image $\m{I}_{\text{T}}$ from a masked version of itself and a reference image $\m{I}_{\text{R}}$, using a ViT encoder, a cross-attention decoder, and a pixel decoder (full formalism in~\Cref{sec:croco_mae_formalism}). The per-pixel reconstruction error on masked pixels is:
\begin{equation}
    \ell_{\text{CroCo}}\left( \bp \right) = \left\Vert \m{I}_{\text{T}}\left( \bp \right) - \hat{\m{I}}_{\text{T}|\text{R}}\left( \bp \right) \right\Vert_2^2.\label{eq:lossCroco}
\end{equation}
In co-visible regions CroCo transfers information from the reference, implicitly learning correspondences~\cite{an2025cross}; elsewhere the reference adds little and the reconstruction degrades (\Cref{fig:croco_rec}).

A Masked Autoencoder (MAE)~\cite{he2022masked} reconstructs $\m{I}_{\text{T}}$ from a masked version of itself \emph{only}, with per-pixel error:
\begin{equation}
    \ell_{\text{MAE}}\left( \bp \right) = \left\Vert \m{I}_{\text{T}}\left( \bp \right) - \hat{\m{I}}_{\text{T}}\left( \bp \right)\right\Vert_2^2.\label{eq:lossMAE}
\end{equation}
MAE cannot exploit the reference view at all: it inpaints from unmasked target patches alone.

\subsection{Relative improvement of CroCo over MAE}\label{sec:analysis_rel_imp}
\begin{figure}[ht]
    \centering
    \begin{minipage}[t]{0.485\linewidth}
        \centering
        \begin{tikzpicture}
        \begin{axis}[
            width=\linewidth,
            height=4.3cm,
            title={\small CroCo's error alone, AP $0.57$},
            xlabel={$\ell_{\text{CroCo}}\left( \bp \right)$},
            ylabel={Density},
            ymin=0,
            xmin=0,
            xmax=1,
            legend pos=north east,
            legend style={font=\scriptsize},
            grid=major,
            grid style={gray!30},
            area style,
        ]
        \addplot+[ybar interval, fill=red!40, draw=red!70, opacity=0.7] coordinates {
            (0.0150, 11.7242) (0.0450, 4.6226) (0.0750, 2.9271) (0.1050, 2.0945) (0.1350, 1.5978) (0.1650, 1.2646) (0.1950, 1.0275) (0.2250, 0.8564) (0.2550, 0.7241) (0.2850, 0.6220) (0.3150, 0.5414) (0.3450, 0.4780) (0.3750, 0.4183) (0.4050, 0.3724) (0.4350, 0.3335) (0.4650, 0.3021) (0.4950, 0.2726) (0.5250, 0.2474) (0.5550, 0.2241) (0.5850, 0.2106) (0.6150, 0.1913) (0.6450, 0.1762) (0.6750, 0.1618) (0.7050, 0.1480) (0.7350, 0.1392) (0.7650, 0.1300) (0.7950, 0.1205) (0.8250, 0.1115) (0.8550, 0.1036) (0.8850, 0.0972) (0.9150, 0.0887) (0.9450, 0.0857) (0.9750, 0.0794) (1.0050, 0.0745) (1.0350, 0.0699) (1.0650, 0.0666) (1.0950, 0.0626) (1.1250, 0.0588) (1.1550, 0.0572) (1.1850, 0.0528) (1.2150, 0.0496) (1.2450, 0.0473) (1.2750, 0.0444) (1.3050, 0.0425) (1.3350, 0.0396) (1.3650, 0.0386) (1.3950, 0.0373) (1.4250, 0.0334) (1.4550, 0.0329) (1.4850, 0.0315) 
        };
        \addlegendentry{Non co-visible}
        \addplot+[ybar interval, fill=blue!40, draw=blue!70, opacity=0.7] coordinates {
            (0.0150, 15.7365) (0.0450, 5.0077) (0.0750, 2.8322) (0.1050, 1.8692) (0.1350, 1.3313) (0.1650, 1.0037) (0.1950, 0.7797) (0.2250, 0.6239) (0.2550, 0.5096) (0.2850, 0.4247) (0.3150, 0.3570) (0.3450, 0.3042) (0.3750, 0.2611) (0.4050, 0.2276) (0.4350, 0.1987) (0.4650, 0.1748) (0.4950, 0.1545) (0.5250, 0.1383) (0.5550, 0.1238) (0.5850, 0.1110) (0.6150, 0.0999) (0.6450, 0.0898) (0.6750, 0.0830) (0.7050, 0.0745) (0.7350, 0.0687) (0.7650, 0.0636) (0.7950, 0.0579) (0.8250, 0.0534) (0.8550, 0.0493) (0.8850, 0.0458) (0.9150, 0.0415) (0.9450, 0.0391) (0.9750, 0.0370) (1.0050, 0.0345) (1.0350, 0.0316) (1.0650, 0.0292) (1.0950, 0.0272) (1.1250, 0.0257) (1.1550, 0.0239) (1.1850, 0.0226) (1.2150, 0.0209) (1.2450, 0.0202) (1.2750, 0.0189) (1.3050, 0.0179) (1.3350, 0.0170) (1.3650, 0.0156) (1.3950, 0.0151) (1.4250, 0.0141) (1.4550, 0.0133) (1.4850, 0.0128)
        };
        \addlegendentry{Co-visible}
        \end{axis}
        \end{tikzpicture}
    \end{minipage}\hfill
    \begin{minipage}[t]{0.485\linewidth}
        \centering
        \begin{tikzpicture}
        \begin{axis}[
            width=\linewidth,
            height=4.3cm,
            title={\small Relative improvement, AP $0.74$},
            xlabel={$\m{C}\left( \bp \right)$},
            ylabel={Density},
            ymin=0,
            xmin=0,
            xmax=1,
            legend pos=north east,
            legend style={font=\scriptsize},
            grid=major,
            grid style={gray!30},
            area style,
        ]
        \addplot+[ybar interval, fill=red!40, draw=red!70, opacity=0.7] coordinates {
            (0.0100, 6.9866) (0.0300, 5.1941) (0.0500, 3.9127) (0.0700, 3.0417) (0.0900, 2.4173) (0.1100, 2.0128) (0.1300, 1.7129) (0.1500, 1.4894) (0.1700, 1.3028) (0.1900, 1.1629) (0.2100, 1.0427) (0.2300, 0.9338) (0.2500, 0.8499) (0.2700, 0.7769) (0.2900, 0.7200) (0.3100, 0.6782) (0.3300, 0.6340) (0.3500, 0.6031) (0.3700, 0.5752) (0.3900, 0.5552) (0.4100, 0.5347) (0.4300, 0.5169) (0.4500, 0.4998) (0.4700, 0.4927) (0.4900, 0.4759) (0.5100, 0.4624) (0.5300, 0.4554) (0.5500, 0.4465) (0.5700, 0.4329) (0.5900, 0.4306) (0.6100, 0.4183) (0.6300, 0.4210) (0.6500, 0.4214) (0.6700, 0.4230) (0.6900, 0.4180) (0.7100, 0.4190) (0.7300, 0.4222) (0.7500, 0.4213) (0.7700, 0.4283) (0.7900, 0.4391) (0.8100, 0.4496) (0.8300, 0.4585) (0.8500, 0.4658) (0.8700, 0.4781) (0.8900, 0.4832) (0.9100, 0.4842) (0.9300, 0.4721) (0.9500, 0.4594) (0.9700, 0.4087) (0.9900, 0.2589) 
        };
        \addlegendentry{Non co-visible}
        \addplot+[ybar interval, fill=blue!40, draw=blue!70, opacity=0.7] coordinates {
            (0.0100, 1.5654) (0.0300, 1.4493) (0.0500, 1.2803) (0.0700, 1.1693) (0.0900, 1.0856) (0.1100, 1.0198) (0.1300, 0.9563) (0.1500, 0.9113) (0.1700, 0.8717) (0.1900, 0.8358) (0.2100, 0.8039) (0.2300, 0.7798) (0.2500, 0.7578) (0.2700, 0.7416) (0.2900, 0.7361) (0.3100, 0.7298) (0.3300, 0.7253) (0.3500, 0.7215) (0.3700, 0.7164) (0.3900, 0.7151) (0.4100, 0.7197) (0.4300, 0.7265) (0.4500, 0.7293) (0.4700, 0.7370) (0.4900, 0.7520) (0.5100, 0.7660) (0.5300, 0.7782) (0.5500, 0.7976) (0.5700, 0.8185) (0.5900, 0.8438) (0.6100, 0.8663) (0.6300, 0.8917) (0.6500, 0.9169) (0.6700, 0.9476) (0.6900, 0.9817) (0.7100, 1.0172) (0.7300, 1.0557) (0.7500, 1.0988) (0.7700, 1.1418) (0.7900, 1.1867) (0.8100, 1.2376) (0.8300, 1.2897) (0.8500, 1.3492) (0.8700, 1.4112) (0.8900, 1.4751) (0.9100, 1.5279) (0.9300, 1.5605) (0.9500, 1.5491) (0.9700, 1.3842) (0.9900, 0.8706)
        };
        \addlegendentry{Co-visible}
        \end{axis}
        \end{tikzpicture}
    \end{minipage}
    \caption{\textbf{The relative improvement of CroCo over MAE separates co-visibility.}
    Histograms split by ground-truth co-visibility, computed on the same 100 ScanNet-50 pairs (\Cref{sec:details_hist}).
    \textbf{(left)} CroCo's reconstruction error $\ell_{\text{CroCo}}\left( \bp \right)$ (Eq.~\ref{eq:lossCroco}) alone. Errors are on average lower in co-visible regions (blue) than in non-co-visible ones (red), but the two distributions largely overlap, giving an average precision of only $0.57$.
    \textbf{(right)} The relative improvement $\m{C}\left( \bp \right)$ (Eq.~\ref{eq:RelImp}). Values concentrate near zero for non-co-visible regions and spread across $[0,1]$ for co-visible ones, raising the average precision to $0.74$.
    }
    \label{fig:hist}
\end{figure}
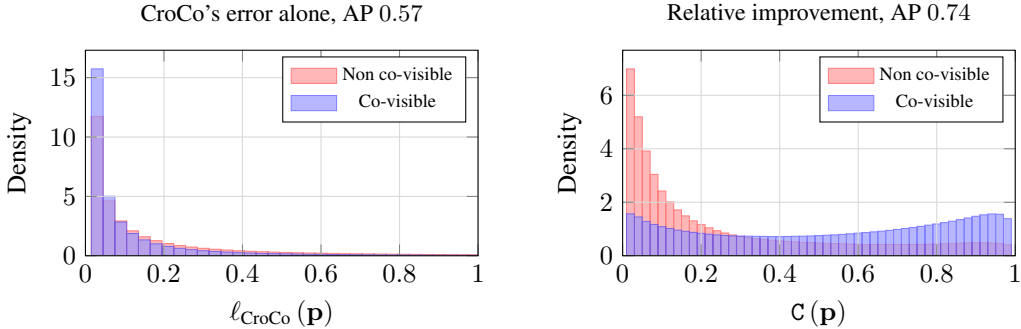

A natural question is whether co-visibility can be predicted from CroCo's reconstruction error alone. \Cref{fig:hist}~(left) shows histograms of $\ell_{\text{CroCo}}(\bp)$ split by ground-truth co-visibility labels. The distributions largely overlap, indicating that CroCo's error alone is an unreliable predictor.

MAE provides a reconstruction error baseline that does not exploit cross-view cues and is independent of the co-visibility between the reference and target images.

This raises another question: \emph{can co-visibility be predicted by comparing CroCo's and MAE's reconstruction errors?}

We define the relative improvement of CroCo over MAE:
\begin{equation}
    \m{C}\left( \bp \right) = \frac{\ell_{\text{MAE}}\left( \bp \right) - \ell_{\text{CroCo}}\left( \bp \right)}{\ell_{\text{MAE}}\left( \bp \right)}.\label{eq:RelImp}
\end{equation}
In co-visible regions, $\ell_{\text{CroCo}}\left( \bp \right) \ll \ell_{\text{MAE}}\left( \bp \right)$ and consequently $\m{C}\left( \bp \right) \approx 1$, because the network has access to the reference view. In non-co-visible regions, both networks perform similarly, leading to $\m{C}\left( \bp \right) \approx 0$.

\Cref{fig:hist} (right) reports histograms of $\m{C}\left( \bp \right)$ computed using ground-truth co-visibility labels. Ideally, co-visible regions would concentrate near 1. The distribution does spread over $[0,1]$, and the reason is structural: $\m{C}(\bp)$ is a ratio whose denominator is the MAE error, so it is uninformative wherever that error is already small. This happens in uniform regions but equally in repeated or self-similar structure: the target alone already predicts the pixel well, the reference cannot improve on it, and $\m{C}(\bp)$ stays near zero whether or not the pixel is co-visible. $\m{C}(\bp)$ therefore \emph{approximates} co-visibility rather than classifying it, and $\ell_{\text{MAE}}(\bp)$ is itself the detector of when it fails, which is what~\Cref{sec:training_objective} exploits.
Nevertheless, compared to CroCo's error alone (\Cref{fig:hist} (left)), the two distributions overlap substantially less, reaching an average precision of $0.74$ compared to $0.57$ for co-visibility prediction. The relative improvement $\m{C}(\bp)$ is thus a more reliable indicator of co-visibility. In~\Cref{sec:covis_probe} we show that the channel~\ourmethod actually \emph{learns} is a substantially better co-visibility classifier still, on scenes never seen during pre-training.

\section{Method}\label{sec:method}
The relative improvement $\m{C}(\bp)$ from~\Cref{sec:analysis_rel_imp} provides a pixelwise, albeit noisy, self-supervised co-visibility signal. Predicting co-visibility requires geometric reasoning, making it a useful training objective for 3D vision~\cite{loiseau2025alligat0r}. However, existing co-visibility objectives rely on ground-truth depth maps and camera poses, limiting their applicability.

We introduce \ourmethod, a self-supervised pre-training method that uses $\m{C}(\bp)$ as pseudo co-visibility labels to improve CroCo's pre-training without any 3D annotations (\Cref{fig:teaser}).

\subsection{Pre-training strategy}

\ourmethod is a network $F_{\text{\ourmethod}}$, \emph{trained from scratch}, and used in three complementary ways during pre-training.

Given a target image $\m{I}_{\text{T}}$ and a reference image $\m{I}_{\text{R}}$ of the same scene from different viewpoints, \ourmethod performs three forward passes:

\begin{enumerate}
    \item A cross-view completion pass that reconstructs $\m{I}_{\text{T}}$ from a masked version of itself \emph{and} $\m{I}_{\text{R}}$: 
    \begin{equation}
    \hat{\m{I}}_{\text{T}|\text{R}} = F_{\text{\ourmethod}}\left( \m{M} \odot  \m{I}_{\text{T}}, \m{I}_{\text{R}}  \right),
    \end{equation}
    where $\m{M}$ masks out a random subset (90\%) of the target input patches.
    \item A masked autoencoder pass that reconstructs $\m{I}_{\text{T}}$ from a masked version of itself \emph{only}:
    \begin{equation}
    \hat{\m{I}}_{\text{T}} = F_{\text{\ourmethod}}\left( \m{M} \odot  \m{I}_{\text{T}}  \right),
    \end{equation}
    where $\m{M}$ is the \emph{same} mask as the one used for the cross-view completion pass.
    \item A relative improvement prediction pass that predicts the relative improvement map ${\m{C}}$ (Eq.~\ref{eq:RelImp}) of CroCo's reconstruction error (first pass) over MAE's reconstruction error (second pass), from $\m{I}_{\text{T}}$ (unmasked) and $\m{I}_{\text{R}}$:
    \begin{equation}
    \hat{\m{C}} = F_{\text{\ourmethod}}\left(\m{I}_{\text{T}}, \m{I}_{\text{R}}  \right).
    \end{equation}
\end{enumerate}

\subsection{Network architecture}\label{sec:architecture}
The architecture $F_{\text{\ourmethod}}$ follows CroCo~\cite{weinzaepfel2022croco}: a ViT-based encoder, a cross-attention based decoder, and a lightweight pixel decoder.
In its original version, the pixel decoder outputs three channels corresponding to the RGB values of the reconstructed target image. In \ourmethod, a fourth channel is added to predict the relative improvement map $\hat{\m{C}}$.
This architecture handles all three passes: in the MAE pass (second pass), since there is no reference image, the cross-attention layers operate as self-attention with the unmasked target patches and dedicated mask tokens, as in classical MAE.
The fact that CroCo and~\ourmethod share the same architecture (up to the extra output channel) and the same training data enables fair comparison in~\Cref{sec:experiments}.

\subsection{Training objective}\label{sec:training_objective}
The network $F_{\text{\ourmethod}}$ is trained \emph{from scratch}, in a fully self-supervised manner, minimizing:
\begin{equation}
    \calL_{\text{\ourmethod}} = \sum_{\bp\in \Omega_{\m{M}}} \ell_{\text{MAE}}\left( \bp \right) + \ell_{\text{CroCo}}\left( \bp \right) + \ell_{\text{RI}}\left( \bp \right)
\end{equation}
where $\Omega_{\m{M}}$ is the set of masked target pixel locations, $\ell_{\text{MAE}}$ and $\ell_{\text{CroCo}}$ are defined in Eq.~\ref{eq:lossMAE} and Eq.~\ref{eq:lossCroco} respectively, and the relative improvement loss is:
\begin{equation}
    \ell_{\text{\text{RI}}}\left( \bp \right) =
     \left(\textit{sg}\left[ \ell_{\text{MAE}}\left( \bp \right) - \ell_{\text{CroCo}}\left( \bp \right)\right] - \textit{sg}\left[\ell_{\text{MAE}}\left( \bp \right)\right]\hat{\m{C}}\left( \bp \right)\right)^2,\label{eq:loss_rel}
\end{equation}
where $\textit{sg}\left[\cdot\right]$ denotes the stop-gradient operator.
By construction, when $\hat{\m{C}}\left( \bp \right) = \frac{\ell_{\text{MAE}}\left( \bp \right) - \ell_{\text{CroCo}}\left( \bp \right)}{\ell_{\text{MAE}}\left( \bp \right)}$, the loss satisfies $\ell_{\text{RI}}\left( \bp \right) = 0$.
Note that Eq.~\ref{eq:loss_rel} differs slightly from a direct regression of $\m{C}(\bp)$; this formulation down-weights pixels where the MAE loss is low (see~\Cref{sec:loss_ablation}), which often correspond to uniform regions that yield noisy pseudo-labels. This down-weighting leads to significantly improved performance and sharper co-visibility maps (see~\Cref{sec:ablations} and~\Cref{fig:heatmaps}).
As in CroCo~\cite{weinzaepfel2022croco}, reference and target images are normalized such that each patch has zero mean and unit variance.

In~\Cref{fig:heatmaps}, \ourmethod's relative improvement maps $\hat{\m{C}}$, while less sharp in uniform regions, are strongly correlated with ground-truth co-visibility labels.
\section{Experiments}\label{sec:experiments}
We evaluate~\ourmethod on three complementary tasks: zero-shot correspondence estimation directly after pre-training following~\cite{an2025cross}, relative metric pose estimation after fine-tuning with a lightweight prediction head on top of frozen features, and pointmap regression with a DPT head~\cite{ranftl2021vision}. All comparisons against CroCo use identical architectures and training data to ensure fairness.
The appendix reports a broader set of experiments: generality across architectures (\Cref{sec:arch_generality}), single-image probes on depth, semantic segmentation and classification (\Cref{sec:mono}), full-network fine-tuning on pose and optical flow (\Cref{sec:fullft}), complete pointmap metrics (\Cref{sec:pointmap_full}), data scaling on the raw-video mix (\Cref{sec:data_scaling}), a per-row ablation analysis (\Cref{sec:detailed_ablation}), and additional qualitative results (\Cref{sec:more_quali}).

\subsection{Experimental setup}
\paragraph{Pre-training datasets.}
All models are pre-trained on the indoor scenes ScanNet split of the Cub3 dataset~\cite{loiseau2025alligat0r}, which consists of image pairs extracted from ScanNet~\cite{dai2017scannet}. Two overlap variants test whether~\ourmethod benefits from low-overlap pairs, which standard cross-view completion does not: at least 50\% overlap (\emph{ScanNet-50}) and at least 5\% (\emph{ScanNet-all}). In~\Cref{sec:curriculum}, models are also pre-trained on DL3DV~\cite{ling2024dl3dv} to evaluate generalization to outdoor and diverse scenes.
\Cref{sec:covis_probe} and~\Cref{sec:vitl} instead use a 12-source mix of raw video, consumed with the stride curriculum and no 3D preprocessing (\Cref{sec:mix}); those models are written \ourmethod-B$^{\text{mix}}$ and \ourmethod-L$^{\text{mix}}$.

\paragraph{Model architectures.}
Following CroCo~\cite{weinzaepfel2022croco}, all models use a ViT-based architecture~\cite{dosovitskiy2020image} in Base and Large sizes (details in~\Cref{sec:archi_details}). As described in~\Cref{sec:architecture}, \ourmethod adds a single output channel to CroCo's architecture to predict the relative improvement map~$\hat{\m{C}}$, enabling direct and fair comparison. All pre-training and fine-tuning hyperparameters are given in~\Cref{sec:training_details}. Every CroCo result reported in this paper is our own pre-training from scratch under identical architecture, data, augmentation and schedule; no released checkpoint is used, with the single exception of the publicly released backbones probed in~\Cref{tab:vitl}.

\subsection{Zero-shot correspondence estimation}
Features are first evaluated immediately after pre-training, without any fine-tuning, on dense correspondence estimation. Following~\cite{an2025cross}, we evaluate on ETH3D~\cite{schops2017multi} and report the Average End-Point Error (AEPE, $\downarrow$) for correspondences read from either the encoder or the decoder, both of which cross-view completion models implicitly learn.
\ourmethod cuts AEPE by 22\% for encoder features and 21\% for decoder features over the CroCo baseline (\Cref{tab:eth3d}), so the extra training signal improves features for downstream tasks relying on either part of the network.

\begin{table}[ht]
\centering
\begin{minipage}[t]{0.32\linewidth}
    \centering
    \caption{\textbf{Zero-shot correspondences on ETH3D.} AEPE ($\downarrow$). All models are Base, pre-trained on ScanNet-50.}
    \label{tab:eth3d}
    \begin{tabular}{lc}
        \toprule
        Model & {AEPE ($\downarrow$)} \\
        \midrule
        \multicolumn{2}{l}{\textit{Encoder}} \\
        CroCo-B & 78.4 \\
        \ourmethod-B (ours) & \textbf{60.8} \\
        \midrule
        \multicolumn{2}{l}{\textit{Decoder}} \\
        CroCo-B & 101.9 \\
        \ourmethod-B (ours) & \textbf{80.8} \\
        \bottomrule
    \end{tabular}
\end{minipage}%
\hfill%
\begin{minipage}[t]{0.65\linewidth}
    \centering
    \caption{\textbf{Curriculum pre-training on ScanNet-1500.} Percentage of pairs ($\uparrow$) within thresholds. ``Curriculum'' denotes training from raw videos with a stride-based schedule (\Cref{sec:curriculum}). Best in \textbf{bold}.}
    \label{tab:curriculum}
    \resizebox{\linewidth}{!}{%
    \begin{tabular}{llccc}
        \toprule
        \multirow{2}{*}{Pre-training data} & \multirow{2}{*}{Model} & \multicolumn{3}{c}{ScanNet-1500} \\
        \cmidrule(lr){3-5}
        & & 10$^\circ$/0.25m & 10$^\circ$/0.5m & 10$^\circ$/1m \\
        \midrule
        \multicolumn{5}{l}{\textit{ScanNet (curriculum)}} \\
        \multirow{2}{*}{\quad 50} & CroCo-B & 20.1 & 34.4 & 40.2 \\
        & \ourmethod-B & 31.2 & 45.0 & 49.0 \\
        \multirow{2}{*}{\quad all} & CroCo-B & 18.9 & 35.1 & 43.9 \\
        & \ourmethod-B & \textbf{39.4} & \textbf{56.5} & \textbf{61.1} \\
        \midrule
        \multicolumn{5}{l}{\textit{DL3DV (curriculum)}} \\
        \multirow{2}{*}{\quad 50} & CroCo-B & 8.5 & 19.9 & 27.2 \\
        & \ourmethod-B & 29.0 & 48.4 & 54.3 \\
        \multirow{2}{*}{\quad all} & CroCo-B & 15.6 & 29.3 & 36.8 \\
        & \ourmethod-B & \textbf{34.9} & \textbf{51.1} & \textbf{57.1} \\
        \bottomrule
    \end{tabular}%
    }
\end{minipage}
\end{table}

\subsection{Relative metric pose estimation}
%
We next fine-tune a lightweight MLP head on frozen features for relative \emph{metric} pose estimation on ScanNet-1500~\cite{sarlin2020superglue}, a task that requires reasoning about both viewpoint and scene geometry.
\ourmethod-B outperforms CroCo-B by nearly $6\times$ at the strictest threshold (\Cref{tab:pose}), and the gap widens on ScanNet-all ($43.7\%$ vs.\ $6.6\%$): \ourmethod benefits from low-overlap pairs where CroCo does not.

The advantage carries over to Large models: on ScanNet-all, \ourmethod-L reaches 35.9\% at $10^\circ$/0.25m against 7.5\% for CroCo-L, a $4.8\times$ improvement. Both Large models are pre-trained by us for 100k steps at the same global batch size, so the comparison isolates the training objective at both scales. The benefits of the relative improvement loss thus scale to larger architectures.

\begin{table}[ht]
    \centering
    \caption{\textbf{Relative metric pose estimation on ScanNet-1500.} Percentage of pairs ($\uparrow$) within rotation/translation thresholds. Every row is our own pre-training from scratch; within each of the Base and Large groups the two rows share architecture, pre-training data, global batch size (768) and number of steps (100k), so the only difference is the training objective. Best result per column in \textbf{bold}.}
    \label{tab:pose}
    \resizebox{\linewidth}{!}{%
    \begin{tabular}{lcccccc}
        \toprule
        \multirow{2}{*}{Model} & \multicolumn{3}{c}{ScanNet-50} & \multicolumn{3}{c}{ScanNet-all} \\
        \cmidrule(lr){2-4} \cmidrule(lr){5-7}
        & 10$^\circ$/0.25m & 10$^\circ$/0.5m & 10$^\circ$/1m & 10$^\circ$/0.25m & 10$^\circ$/0.5m & 10$^\circ$/1m \\
        \midrule
        CroCo-B & 5.5 & 12.9 & 18.5 & 6.6 & 15.7 & 23.0 \\
        \ourmethod-B (ours) & \textbf{29.1} & \textbf{43.3} & \textbf{47.5} & 43.7 & 54.4 & 57.6 \\
        \midrule
        CroCo-L & -- & -- & -- & 7.5 & 16.1 & 23.0 \\
        \ourmethod-L (ours) & -- & -- & -- & \textbf{35.9} & \textbf{50.6} & \textbf{56.3} \\
        \bottomrule
    \end{tabular}%
    }
\end{table}

\subsection{Curriculum pre-training}\label{sec:curriculum}
%
3D preprocessing~\cite{weinzaepfel2023croco,loiseau2025alligat0r} is expensive and limits scalability to new datasets. We therefore train directly from raw video with a simple stride-based curriculum.

\paragraph{Curriculum strategy.}
Pairs start at a stride of 1 between consecutive frames; every $N$ steps the maximum stride $k_{\max}$ grows by $\Delta k$, and each pair draws its stride uniformly from $[1, k_{\max}]$. Two variants mimic the overlap distributions of the preprocessed datasets: a \emph{-50} variant producing mostly high-overlap pairs, and an \emph{-all} variant that reaches low-overlap pairs earlier. Schedule constants are given in~\Cref{sec:training_details}.

\paragraph{Results.}
\Cref{tab:curriculum} reports pose estimation on ScanNet-1500. Trained from raw video with no 3D annotation and no overlap-based preprocessing, \ourmethod matches its curated-data counterpart (31.2\% vs.\ 29.1\% at $10^\circ$/0.25m; the exact training pairs differ between the two settings). The curriculum helps CroCo considerably on its own (20.1\% vs.\ 5.5\%), so progressive stride scheduling is useful independently of our objective. Yet \ourmethod keeps a clear margin under it, on DL3DV too. Better sampling does not substitute for the relative improvement signal.

\subsection{Pointmap regression}\label{sec:pointmap}
%
Following DUSt3R~\cite{wang2024dust3r}, a DPT head on frozen features regresses per-pixel 3D pointmaps in the first image's frame, evaluated in-domain (ScanNet) and out-of-domain (DL3DV, ETH3D). \ourmethod outperforms CroCo on all three benchmarks and both pre-training variants (\Cref{tab:pointmap}), reducing Chamfer error by 10\% in-domain and by 5--7\% out-of-domain, so the relative improvement signal produces features that generalize beyond the fine-tuning distribution.

\begin{table}[ht]
    \centering
    \caption{\textbf{Pointmap regression on ScanNet, DL3DV and ETH3D.} Chamfer Overall ($\downarrow$) for models pre-trained on DL3DV with curriculum and fine-tuned on ScanNet-all. \ourmethod outperforms CroCo on all three benchmarks, including out-of-domain evaluation on DL3DV and ETH3D. Accuracy and Completeness are broken out in~\Cref{tab:pointmap_full}. Best per group in \textbf{bold}.}
    \label{tab:pointmap}
    \begin{tabular}{llccc}
        \toprule
        Pre-training & Model & ScanNet & DL3DV & ETH3D \\
        \midrule
        \multirow{2}{*}{DL3DV-50} & CroCo-B & 0.126 & 1.430 & 0.399 \\
        & \ourmethod-B & \textbf{0.113} & \textbf{1.359} & \textbf{0.373} \\
        \midrule
        \multirow{2}{*}{DL3DV-all} & CroCo-B & 0.122 & 1.409 & 0.398 \\
        & \ourmethod-B & \textbf{0.115} & \textbf{1.366} & \textbf{0.373} \\
        \bottomrule
    \end{tabular}
\end{table}

\subsection{Does the predicted channel recover co-visibility?}\label{sec:covis_probe}
\Cref{sec:analysis_rel_imp} showed that the \emph{analytic} relative improvement separates co-visible pixels better than $\ell_{\text{CroCo}}$ alone. We now test what the network actually predicts (\Cref{tab:covis_probe,tab:covis_tertile}): we run~\ourmethod once per pair with \emph{no masking} and threshold $\hat{\m{C}}$ as a per-pixel co-visibility classifier, against the same ground truth as~\Cref{sec:analysis_rel_imp}. Both models are Base, 200k steps on the mix of~\Cref{sec:mix}; neither benchmark contributes a pre-training scene. Protocol in~\Cref{sec:covis_protocol}.

$\hat{\m{C}}$ reaches $0.763$ AP on ScanNet-1500 against $0.576$ for $\ell_{\text{CroCo}}$ ($+0.26$ over the $0.502$ chance level against $+0.07$), and it separates the classes at one global threshold with $0.691$ balanced accuracy (\Cref{tab:covis_probe}). On 7-Scenes, where $73\%$ of pixels are co-visible and AP compresses, ROC-AUC is clearer: $0.702$ against $0.603$. The margin is widest where co-visibility is hardest: on the lowest-overlap tertile $\hat{\m{C}}$ gains $+0.32$ AP over chance against $+0.04$ (\Cref{tab:covis_tertile}). It even beats the pseudo-label it was trained on: recomputing $\m{C}(\bp)$ from the same model under its 90\% mask gives only $0.555$ AP, since a noisy per-mask estimate is a worse signal than the quantity regressed from it.

\begin{table}[t]
    \centering
    \begin{minipage}[t]{0.63\linewidth}
    \centering
    \caption{\textbf{Co-visibility classification from the predicted channel $\hat{\m{C}}$.} One unmasked forward pass, scored against depth-and-pose ground truth ($\uparrow$; \emph{b.acc.} = balanced accuracy at a single fitted threshold). Both models Base, 200k steps on the mix of~\Cref{sec:mix}. Protocol in~\Cref{sec:covis_protocol}.}
    \label{tab:covis_probe}
    \resizebox{\linewidth}{!}{%
    \begin{tabular}{lcccccc}
        \toprule
        \multirow{2}{*}{Score} & \multicolumn{3}{c}{ScanNet-1500 (chance AP $0.502$)} & \multicolumn{3}{c}{7-Scenes (chance AP $0.726$)} \\
        \cmidrule(lr){2-4} \cmidrule(lr){5-7}
        & AP & ROC-AUC & b.acc. & AP & ROC-AUC & b.acc. \\
        \midrule
        $\ell_{\text{CroCo}}$ (CroCo-B$^{\text{mix}}$) & 0.576 & 0.577 & 0.539 & 0.799 & 0.603 & 0.554 \\
        $\hat{\m{C}}$ (\ourmethod-B$^{\text{mix}}$, ours) & \textbf{0.763} & \textbf{0.737} & \textbf{0.691} & \textbf{0.859} & \textbf{0.702} & \textbf{0.658} \\
        \bottomrule
    \end{tabular}%
    }
    \end{minipage}%
    \hfill%
    \begin{minipage}[t]{0.34\linewidth}
    \centering
    \caption{\textbf{AP by overlap tertile} on ScanNet-1500. Chance differs per tertile.}
    \label{tab:covis_tertile}
    \resizebox{\linewidth}{!}{%
    \begin{tabular}{lccc}
        \toprule
        AP ($\uparrow$) & low & mid & high \\
        \midrule
        \emph{chance} & \emph{0.285} & \emph{0.506} & \emph{0.714} \\
        $\ell_{\text{CroCo}}$ & 0.326 & 0.570 & 0.777 \\
        $\hat{\m{C}}$ (ours) & \textbf{0.609} & \textbf{0.760} & \textbf{0.849} \\
        \bottomrule
    \end{tabular}%
    }
    \end{minipage}
\end{table}

\subsection{Comparison with released cross-view backbones}\label{sec:vitl}
The comparisons above hold architecture, data and schedule fixed, which isolates the training signal but says nothing about publicly available models. We therefore probe four frozen Large backbones under one protocol, changing only the backbone (\Cref{tab:vitl}). The self-supervised rows differ in corpus and budget, so this is \emph{not} a controlled comparison of objectives (\Cref{tab:pose} remains the matched evidence), and VGGT is a fully supervised ${\sim}1$B reference whose corpus contains ScanNet and DL3DV.
\ourmethod-L$^{\text{mix}}$ leads both self-supervised checkpoints on every cross-view benchmark: $+13.5$ points of pose accuracy at $10^\circ$/0.25m over CroCo~v2-L, $29\%$ lower Chamfer error in-domain and $20\%$ on ETH3D, and less than half the frozen flow error. The gains remain specific to cross-view geometry (\Cref{sec:mono}).

\begin{table}[t]
    \centering
    \caption{\textbf{Frozen Large backbones under a single probing protocol.} Identical head, fine-tuning data and schedule per column; only the frozen backbone changes. Pose: ScanNet-1500, \% of pairs ($\uparrow$) within $10^\circ$ and the given translation threshold. Pointmap: Chamfer Overall ($\downarrow$). Flow: MPI-Sintel \emph{clean} AEPE ($\downarrow$), frozen, so not comparable to published full fine-tuning figures (\Cref{sec:fullft}). The self-supervised rows are \emph{not} data-matched: they use the released checkpoints, and~\Cref{tab:pose} is the matched comparison. \emph{VGGT is a fully supervised reference, not a baseline.} Details in~\Cref{sec:vitl_details}. Best self-supervised in \textbf{bold}.}
    \label{tab:vitl}
    \resizebox{\linewidth}{!}{%
    \begin{tabular}{llcccccccc}
        \toprule
        \multirow{2}{*}{Frozen backbone} & \multirow{2}{*}{Sup.} & \multicolumn{3}{c}{ScanNet-1500 pose ($\uparrow$)} & \multicolumn{3}{c}{Pointmap Overall ($\downarrow$)} & Sintel ($\downarrow$) \\
        \cmidrule(lr){3-5} \cmidrule(lr){6-8} \cmidrule(lr){9-9}
        & & 0.25m & 0.5m & 1m & ScanNet & ETH3D & DL3DV & AEPE clean \\
        \midrule
        \ourmethod-L$^{\text{mix}}$ (ours) & self & \textbf{33.2} & \textbf{53.3} & \textbf{60.1} & \textbf{0.077} & \textbf{0.317} & \textbf{1.122} & \textbf{6.17} \\
        CroCo~v2-L~\cite{weinzaepfel2023croco} & self & 19.7 & 36.2 & 45.3 & 0.109 & 0.394 & 1.359 & 13.08 \\
        MuM-L~\cite{nordstrom2025mum} & self & 19.9 & 32.9 & 39.4 & 0.106 & 0.377 & 1.321 & 22.44 \\
        \midrule
        \emph{VGGT}~\cite{wang2025vggt} & \emph{full} & \emph{92.3} & \emph{98.3} & \emph{99.0} & \emph{0.030} & \emph{0.240} & \emph{1.065} & \emph{--} \\
        \bottomrule
    \end{tabular}%
    }
\end{table}

\subsection{Ablation studies}\label{sec:ablations}
Ablation studies are conducted by pre-training several Base models on ScanNet to analyze the key design choices in~\ourmethod. The different losses are defined in~\Cref{sec:loss_ablation}. All models are evaluated on ScanNet-1500 for relative metric pose estimation. Results are summarized in~\Cref{tab:ablations}.

\begin{table}[t]
    \centering
    \caption{\textbf{Ablation studies on ScanNet-1500.} Pose estimation accuracy ($\uparrow$) at three thresholds. All models are Base, pre-trained on ScanNet. Best results in \textbf{bold}. Loss definitions in~\Cref{sec:loss_ablation}.}
    \label{tab:ablations}
    \resizebox{\linewidth}{!}{%
    \begin{tabular}{lcccccc}
        \toprule
        \multirow{2}{*}{Configuration} & \multicolumn{3}{c}{ScanNet-50} & \multicolumn{3}{c}{ScanNet-all} \\
        \cmidrule(lr){2-4} \cmidrule(lr){5-7}
        & 10$^\circ$/.25m & 10$^\circ$/.5m & 10$^\circ$/1m & 10$^\circ$/.25m & 10$^\circ$/.5m & 10$^\circ$/1m \\
        \midrule
        $\ell_{\text{CroCo}}$ (CroCo-B) & 5.5 & 12.9 & 18.5 & 6.6 & 15.7 & 23.0 \\
        \midrule
        $\ell_{\text{CroCo}}+\ell_{\text{MAE}}$ & 6.7 & 15.3 & 20.9 & 6.2 & 14.5 & 21.2 \\
        $\ell_{\text{conf}}$ & 26.7 & 38.7 & 45.8 & 2.2 & 6.1 & 11.2 \\
        \midrule
         $\ell_{\text{CroCo}}+\ell_{\text{MAE}}+\ell_{\text{AI}}$ & 14.3 & 23.0 & 25.2 & 23.3 & 33.6 & 39.4 \\
         $\ell_{\text{CroCo}}+\ell_{\text{MAE}}+\ell_{\text{AI}}$, no norm. & 20.9 & 38.1 & 42.3 & 14.2 & 21.9 & 25.1 \\
         $\ell_{\text{CroCo}}+\ell_{\text{MAE}}+\ell_{\text{RI ratio}}$ & 27.2 & 39.0 & 42.7 & 25.7 & 39.6 & 44.4 \\
         $\ell_{\text{CroCo}}+\ell_{\text{MAE}}+\ell_{\text{RI}}$ (\ourmethod-B) & \textbf{29.1} & \textbf{43.3} & \textbf{47.5} & \textbf{43.7} & \textbf{54.4} & \textbf{57.6} \\
        \bottomrule
    \end{tabular}%
    }
\end{table}
\begin{figure}[t]
    \centering
    \setlength{\tabcolsep}{0pt}
    \begin{tabular}{@{}c@{}}
        \makebox[0.125\linewidth][c]{\tiny Target}\makebox[0.125\linewidth][c]{\tiny Reference}\makebox[0.125\linewidth][c]{\tiny $\ell_{\text{CroCo}}$~\eqref{eq:lossCroco}}\makebox[0.125\linewidth][c]{\tiny $\ell_{\text{conf}}$~\eqref{eq:loss_conf}}\makebox[0.125\linewidth][c]{\tiny $\ell_{\text{AI}}$~\eqref{eq:loss_rel_2}}\makebox[0.125\linewidth][c]{\tiny $\ell_{\text{AI}}$ \scriptsize{no n.}}\makebox[0.125\linewidth][c]{\tiny $\ell_{\text{RI ratio}}$~\eqref{eq:loss_rel_3}}\makebox[0.125\linewidth][c]{\tiny $\ell_{\text{RI}}$ \ourmethod} \\
        \includegraphics[width=\linewidth]{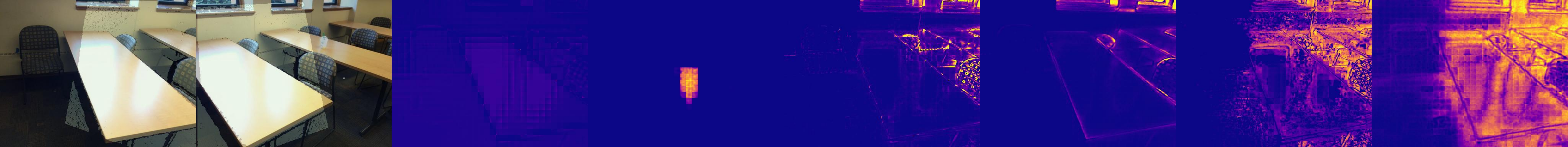} \\
        \includegraphics[width=\linewidth]{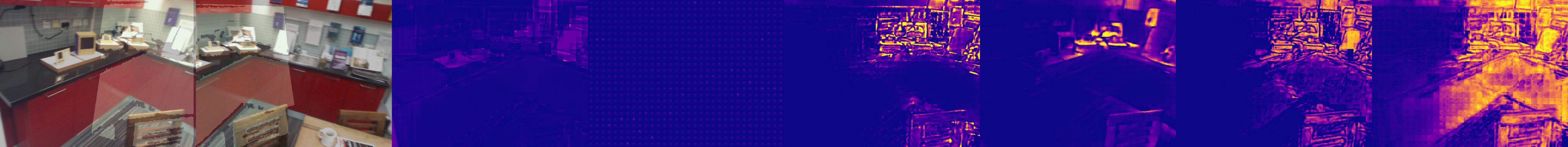} \\
    \end{tabular}
    \caption{\textbf{Qualitative comparison of relative improvement maps $\hat{\m{C}}$ across ablation configurations.} The relative improvement loss $\ell_{\text{RI}}$ (Eq.~\ref{eq:loss_rel}) produces sharper maps that better delineate \textcolor{mygreen}{co-visible} regions, while other configurations yield unreliable predictions. The first row is ScanNet, the second 7-Scenes, which no model here saw during pre-training. More examples in~\Cref{sec:more_quali}.}
    \label{fig:heatmaps}
\end{figure}
Adding MAE without the relative improvement loss (row~2) barely moves the result, so the gain comes from predicting the relative improvement, not from multi-task training. A DUSt3R-style confidence loss~\cite{wang2024dust3r} (row~3) does well on high-overlap pairs and collapses on low-overlap ones. The relative formulation beats the absolute one, patch normalization is critical on ScanNet-all, and down-weighting low-MAE pixels (Eq.~\ref{eq:loss_rel}) is worth $18.0$ points while yielding sharper maps (\Cref{fig:heatmaps}). Per-row analysis in~\Cref{sec:detailed_ablation}.

\subsection{Limitations}\label{sec:limitations}
\paragraph{Training cost.}
\ourmethod is more expensive to pre-train than CroCo. The extra full-resolution pass that predicts $\hat{\m{C}}$ roughly doubles activation memory and so halves the per-device batch (48 vs.\ 96 images for Base models), which means twice the GPUs for the same global batch of 768. A matched pair of 100k-step Base pre-trainings costs ${\sim}2.4\times$ more GPU-hours for~\ourmethod; our reference run takes ${\sim}25$\,h on 16 H100 GPUs, i.e.\ ${\sim}400$ H100-hours. The cost is confined to pre-training: the extra channel is one output dimension and the MAE branch is discarded afterwards, so fine-tuning and inference cost what CroCo costs.

\paragraph{Where the ratio is uninformative.}
$\m{C}(\bp)$ is a ratio, and it is unreliable wherever its denominator is small. A low $\ell_{\text{MAE}}(\bp)$ means the pixel was already predictable from the target alone, in uniform regions but equally in repeated or self-similar structure, so the reference cannot improve on it and $\m{C}(\bp)$ stays near zero regardless of true co-visibility. Eq.~\ref{eq:loss_rel} makes $\ell_{\text{MAE}}$ its own detector of this failure by down-weighting exactly those pixels, which~\Cref{tab:ablations} shows is worth 18.0 points. It is a soft down-weighting, not an exclusion; explicitly masking such regions could help further.

\paragraph{Cross-view, frozen.}
Two scope limits are worth stating plainly. First, the relative improvement is a \emph{binocular} signal and improves the tasks that need one: on single-image probes (NYUv2 depth, ADE20K segmentation, ImageNet classification), \ourmethod and CroCo are within about a point of each other (\Cref{sec:mono}), so we claim no general-purpose representation improvement. Second, every table here probes a \emph{frozen} backbone, which measures what pre-training makes directly accessible rather than its value as an initialisation. Under full fine-tuning the advantage narrows unevenly: on 7-Scenes the position advantage survives and the rotation advantage does not, and on MPI-Sintel it does not survive at all against the released CroCo~v2, pre-trained on 7.3M curated stereo and flow pairs (\Cref{sec:fullft}). Our claim is accordingly the narrower one: \ourmethod gives better frozen cross-view features, not necessarily a better initialisation for full fine-tuning.

\paragraph{Data composition, dynamic scenes, and architecture.}
More data is not automatically better: swapping an indoor-only mix for the full 12-source mix \emph{costs}~\ourmethod five points on indoor ScanNet-1500 while leaving CroCo unchanged (\Cref{sec:data_scaling}). We evaluate only static scenes; nothing in the objective assumes rigidity, but we have not tested it. Finally,~\ourmethod inherits CroCo's restriction to image pairs and a fixed patch size, and a multi-view extension~\cite{li2025muskie,nordstrom2025mum} would have to define the relative improvement against a set of references rather than a single one, which we see as an interesting and non-trivial research direction.

\section{Conclusion}\label{sec:conclusion}

Cross-view completion learns strong features for 3D vision in co-visible regions, but where the reference adds little the training signal is implicitly monocular. We showed that the relative improvement of the cross-view reconstruction error over an MAE error is a self-supervised proxy for co-visibility needing no ground-truth depth or poses, and that a network trained to predict it recovers co-visibility better than the pseudo-label it was trained on. Building on this, \ourmethod jointly performs cross-view completion, masked autoencoding and relative improvement prediction, supplying a binocular signal for \emph{all} masked regions at the cost of one extra output channel. At matched architecture and data it consistently outperforms CroCo on the cross-view tasks we evaluate, and it trains directly from raw video with a simple stride curriculum, removing the 3D preprocessing prior work requires~\cite{weinzaepfel2022croco,loiseau2025alligat0r}.

\begin{ack}
This work was supported by the Bosch Research Foundation (Bosch Forschungsstiftung) and
by the European Union (ERC Advanced Grant Explorer, Funding ID \#101097259). It was
granted access to the HPC resources of IDRIS under the allocation
2026-AD010617525R1 made by GENCI.
\end{ack}

\newpage
\bibliography{main}
\bibliographystyle{plainnat}

\appendix
\clearpage
\section{Broader impact statement}\label{sec:broader_impact}
This work contributes a self-supervised pre-training method for 3D vision. The primary applications are in 3D reconstruction, visual localization, and camera pose estimation, which benefit fields such as robotics, autonomous navigation, and augmented reality. As with any method that improves visual understanding, potential dual-use risks include surveillance applications. However, the method operates on pre-training and does not introduce capabilities beyond those already available in existing systems.

\section{CroCo and MAE formalism}\label{sec:croco_mae_formalism}
\paragraph{CroCo.}
Given a target image $\m{I}_{\text{T}}$ and a reference image $\m{I}_{\text{R}}$ of the same scene from different viewpoints, a Cross-view Completion network $F_{\text{CroCo}}$ reconstructs the target image $\m{I}_{\text{T}}$ from a masked version of itself and the reference image $\m{I}_{\text{R}}$:
\begin{equation}
    \hat{\m{I}}_{\text{T}|\text{R}} = F_{\text{CroCo}}\left( \m{M} \odot  \m{I}_{\text{T}}, \m{I}_{\text{R}}  \right),
\end{equation}
where $\m{M}$ masks out a random subset (90\% in practice) of the target input patches.
A ViT encoder first encodes the reference patches and the unmasked target patches separately. A cross-attention based decoder then processes reference and target tokens to warp information from the reference into the target representation. A lightweight pixel decoder reconstructs the target image $\hat{\m{I}}_{\text{T}|\text{R}}$.
CroCo is trained to minimize the total reconstruction error on masked target pixels: $\calL_{\text{CroCo}} = \sum_{\bp\in \Omega_{\m{M}}} \ell_{\text{CroCo}}\left( \bp \right)$, where $\Omega_{\m{M}}$ is the set of masked target pixel locations and $\ell_{\text{CroCo}}$ is defined in Eq.~\ref{eq:lossCroco}.

\paragraph{MAE.}
A Masked Autoencoder $F_{\text{MAE}}$ reconstructs $\m{I}_{\text{T}}$ from a masked version of itself \emph{only}:
\begin{equation}
    \hat{\m{I}}_{\text{T}} = F_{\text{MAE}}\left( \m{M} \odot  \m{I}_{\text{T}}  \right),
\end{equation}
where $\m{M}$ masks out a random subset (75\% in monocular MAE, we perform 90\% to get the same mask as CroCo and be able to compute the loss and the relative improvement) of the target input patches. MAE is trained to minimize $\calL_{\text{MAE}} = \sum_{\bp\in \Omega_{\m{M}}} \ell_{\text{MAE}}\left( \bp \right)$, where $\ell_{\text{MAE}}$ is defined in Eq.~\ref{eq:lossMAE}.

\section{Training details}\label{sec:training_details}
\paragraph{Stride curriculum.} The two curriculum variants of~\Cref{sec:curriculum} share $N{=}5\text{k}$ steps between increments and differ only in the increment itself: $\Delta k{=}2$ for the \emph{-50} variant and $\Delta k{=}10$ for the \emph{-all} variant. All other hyperparameters are identical to the preprocessed-data runs. The schedule is fixed rather than adaptive so that the comparison against preprocessed baselines is controlled; an adaptive per-sequence rule (for instance increasing the stride on plateau) would make the approach fully automatic.

\paragraph{Pre-training.}
All models are trained from scratch with a batch size of 768 and a learning rate of $1.5 \times 10^{-4}$ for 100k steps. A cosine learning rate decay with a linear warmup of 5000 steps is used. Following CroCo~\cite{weinzaepfel2022croco}, masking ratio is 90\% and patches are normalized to zero mean and unit variance. Images are resized to a maximum dimension of 512 pixels while maintaining aspect ratio.

\paragraph{Fine-tuning for relative metric pose estimation.}
A simple MLP head is added on top of the frozen pre-trained decoder features, trained with a homoscedastic loss~\cite{kendall2018multi} for 20k steps with a batch size of 768 and a constant learning rate of $1 \times 10^{-4}$. Fine-tuning uses the \emph{all} overlap variant to leverage the full range of viewpoint changes.

\paragraph{Fine-tuning for pointmap regression.}
A DPT head~\cite{ranftl2021vision} is fine-tuned on top of frozen pre-trained features using the confidence-aware regression loss from DUSt3R~\cite{wang2024dust3r}. Fine-tuning uses the ScanNet-all dataset with the same batch size and learning rate as pose estimation.

\section{Architecture details}\label{sec:archi_details}
Following CroCo~\cite{weinzaepfel2022croco}, all models use a Vision Transformer (ViT) architecture~\cite{dosovitskiy2020image} in two sizes:
\begin{itemize}
    \item \textbf{Base}: 12-layer ViT encoder and 8-layer transformer decoder with embedding sizes of 768 and 512, pre-trained on both variants of ScanNet.
    \item \textbf{Large}: 24-layer ViT encoder and 12-layer transformer decoder with embedding sizes of 1024 and 768, pre-trained on ScanNet-all only.
\end{itemize}

\section{CroCo qualitative reconstructions}\label{sec:croco_rec_suppl}
\begin{figure}[H]
    \centering
    \begin{tabular}{@{}c@{}}
        \begin{tabular}{@{}c@{}}
            \makebox[0.333\linewidth][c]{Target}\makebox[0.333\linewidth][c]{Reference}\makebox[0.333\linewidth][c]{Reconstruction} \\
        \end{tabular} \\
        \adjustbox{frame}{\includegraphics[width=\linewidth]{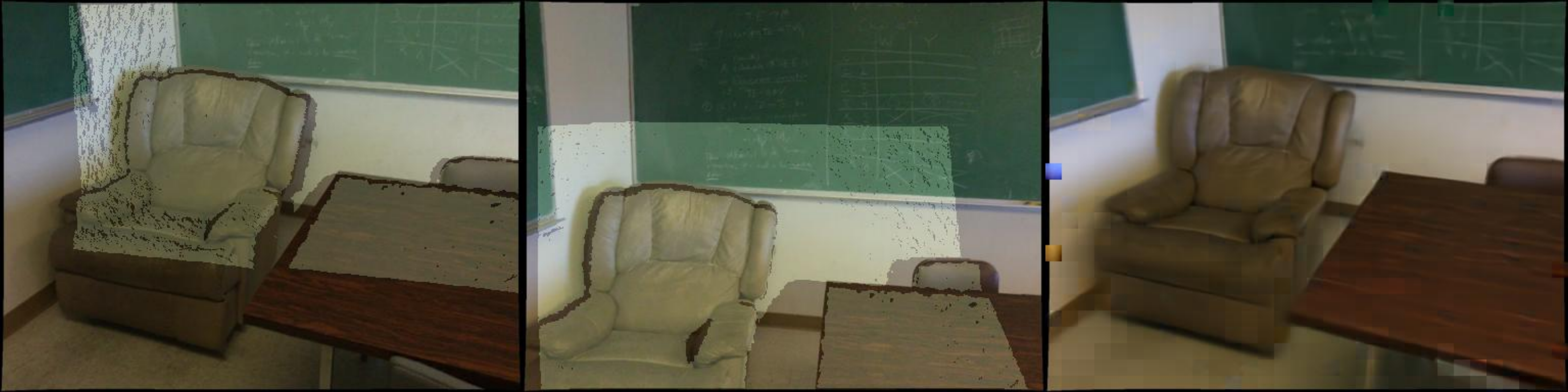}} \\
        \adjustbox{frame}{\includegraphics[width=\linewidth]{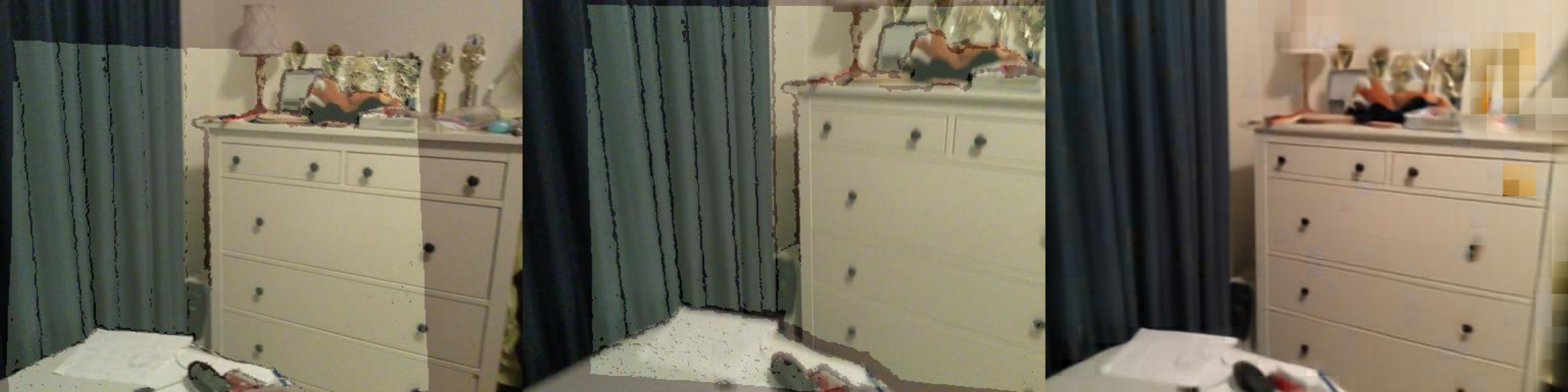}} \\
        \adjustbox{frame}{\includegraphics[width=\linewidth]{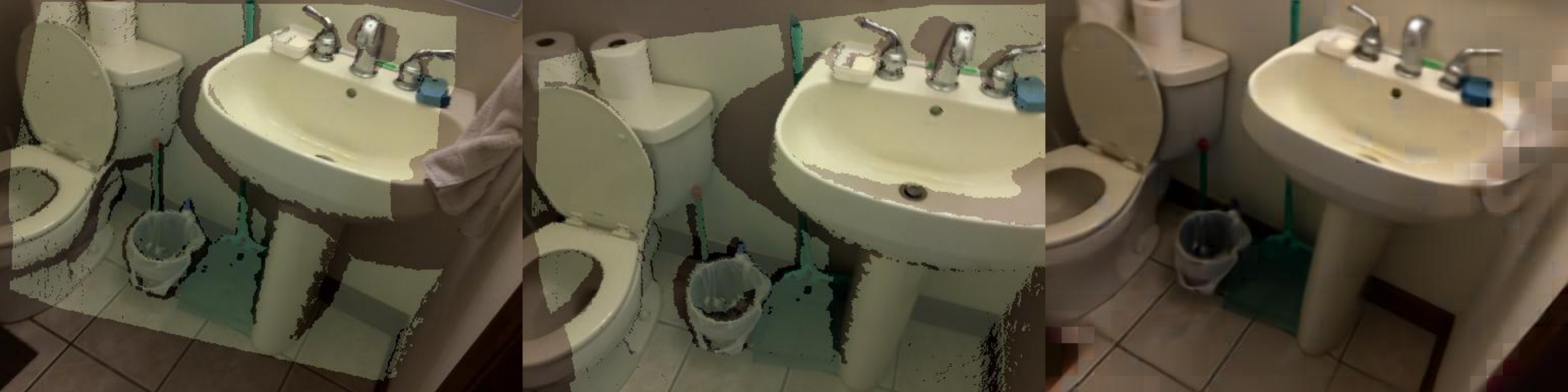}} \\
    \end{tabular}
    \caption{\textbf{Qualitative CroCo reconstructions.} Reconstructions are sharper in \textcolor{mygreen}{co-visible} regions where the network warps information from the reference image. Conversely, reconstructions are blurred in non-co-visible regions where the reference provides little information (best viewed zoomed-in).}
    \label{fig:croco_rec}
\end{figure}

\section{Detailed ablation analysis}\label{sec:detailed_ablation}
This section provides a detailed discussion of each ablation row in~\Cref{tab:ablations}.

\paragraph{Impact of the relative improvement loss.}
Adding the MAE reconstruction branch without the relative improvement loss $\ell_{\text{RI}}$ (row~2) provides only marginal improvement over the CroCo baseline (20.9\% vs.\ 18.5\% at $10^\circ$/1m). The key contribution of~\ourmethod lies in the relative improvement prediction, not merely in jointly training with MAE.

\paragraph{Confidence prediction vs.\ relative improvement.}
The confidence-aware loss performs well on high-overlap pairs (45.8\% at $10^\circ$/1m on ScanNet-50) but fails on low-overlap pairs (11.2\% on ScanNet-all). As illustrated in~\Cref{fig:heatmaps}, the confidence maps lack the sharp co-visibility boundaries that $\ell_{\text{RI}}$ achieves.

\paragraph{Relative vs.\ absolute improvement.}
Regressing the absolute improvement $\ell_{\text{MAE}} - \ell_{\text{CroCo}}$ underperforms the relative formulation, particularly on ScanNet-all (39.4\% vs.\ 57.6\%). The relative improvement produces sharper maps that better capture co-visibility boundaries (\Cref{fig:heatmaps}).

\paragraph{Impact of patch normalization.}
Without normalizing target patch tokens, performance drops significantly on ScanNet-all (25.1\% vs.\ 39.4\%). Removing normalization leads to noisier predictions in uniform regions where MAE predicts a near-constant value, yielding unreliable relative improvement estimates.

\paragraph{Loss formulation details.}
Directly regressing the ratio underperforms the final formulation (Eq.~\ref{eq:loss_rel}) that multiplies the prediction by the denominator, improving from 25.7\% to 43.7\% at $10^\circ$/.25m on ScanNet-all.

\section{Ablation study: technical details}\label{sec:loss_ablation}
The different losses considered in the ablation study (\Cref{tab:ablations}) are defined as follows:
\begin{enumerate}
\item \begin{equation}
    \ell_{\text{\text{RI}}}\left( \bp \right) =
     \left(\textit{sg}\left[ \ell_{\text{MAE}}\left( \bp \right) - \ell_{\text{CroCo}}\left( \bp \right)\right] - \textit{sg}\left[\ell_{\text{MAE}}\left( \bp \right)\right]\hat{\m{C}}\left( \bp \right)\right)^2\label{eq:loss_rel_1}
\end{equation}
is the relative-improvement loss of~\ourmethod (see Eq.~\ref{eq:RelImp}).

\item \begin{equation}
    \ell_{\text{\text{RI ratio}}}\left( \bp \right) =
     \left(\textit{sg}\left[\frac{\ell_{\text{MAE}}\left( \bp \right) - \ell_{\text{CroCo}}\left( \bp \right)}{\ell_{\text{MAE}}\left( \bp \right)}\right] - \hat{\m{C}}\left( \bp \right)\right)^2\label{eq:loss_rel_3}
\end{equation}
is a relative-improvement loss that reweights the previous loss as follows: $\left(\frac{1}{\textit{sg}\left[\ell_{\text{MAE}}\left( \bp \right)\right]}\right)^2\ell_{\text{\text{RI}}}\left( \bp \right)=\ell_{\text{\text{RI ratio}}}\left( \bp \right)$. This reweighting up-weights pixels where the MAE loss is low, which often correspond to uniform regions that yield noisy pseudo-labels. In practice, $\ell_{\text{\text{RI}}}$ significantly outperforms $\ell_{\text{\text{RI ratio}}}$.

\item \begin{equation}
    \ell_{\text{\text{AI}}}\left( \bp \right) =
     \left(\textit{sg}\left[ \ell_{\text{MAE}}\left( \bp \right) - \ell_{\text{CroCo}}\left( \bp \right)\right] - \hat{\m{C}}\left( \bp \right)\right)^2\label{eq:loss_rel_2}
\end{equation}
is an absolute-improvement loss where $\hat{\m{C}}\left( \bp \right)$ learns to predict the difference between $\ell_{\text{CroCo}}$ and $\ell_{\text{MAE}}$.

\item \begin{equation}
    \ell_{\text{\text{conf}}}\left( \bp \right) = \left(1+\exp\left(\hat{\m{C}}\left( \bp \right)\right)\right)\left\Vert \m{I}_{\text{T}}\left( \bp \right) - \hat{\m{I}}_{\text{T}|\text{R}}\left( \bp \right) \right\Vert_2^2 -0.2\ln\left(1+\exp\left(\hat{\m{C}}\left( \bp \right)\right)\right)\label{eq:loss_conf}
\end{equation}
is a confidence-aware loss where $\hat{\m{C}}\left( \bp \right)$ learns to predict the precision (\textit{i.e.} inverse variance) that is expected to be high in co-visible regions and low in non-co-visible regions.
\end{enumerate}
\section{The 12-source raw-video mix}\label{sec:mix}
Two experiments (\Cref{sec:covis_probe} and~\Cref{sec:vitl}) use a mix of twelve video sources rather than the curated Cub3 pairs: ScanNet~\cite{dai2017scannet}, ScanNet++~\cite{yeshwanth2023scannet++}, ARKitScenes~\cite{baruch2021arkitscenes}, RealEstate10K~\cite{zhou2018stereo} and EPIC-KITCHENS~\cite{damen2018scaling} (real indoor); Hypersim~\cite{roberts2021hypersim} (synthetic indoor); CO3Dv2~\cite{reizenstein2021common} and Objectron~\cite{ahmadyan2021objectron} (object-centric); DL3DV~\cite{ling2024dl3dv} and KITTI-360~\cite{liao2022kitti} (real outdoor); and VirtualKITTI2~\cite{cabon2020virtual} and TartanAir~\cite{wang2020tartanair} (synthetic outdoor).

Every source is consumed as \emph{raw video} through the stride curriculum of~\Cref{sec:curriculum}: pairs are formed by sampling two frames of the same clip at a stride drawn from the current curriculum window. No structure-from-motion, depth, camera pose or overlap estimate is used at any point, so adding a source costs only the frames themselves. Sources are sampled with fixed weights, and validation uses a scene-disjoint held-out split of each source. Unless stated otherwise, mix models are trained with the same optimiser, resolution and global batch size as the ScanNet models of~\Cref{sec:training_details}.

\section{Co-visibility probing protocol}\label{sec:covis_protocol}
For~\Cref{tab:covis_probe} we run each model once per pair with \emph{no masking} and read the fourth output channel $\hat{\m{C}}$ directly; there is no post-processing, no test-time augmentation and no fitting of the network. Ground truth is computed on the fly: a target pixel is labelled co-visible if its back-projected 3D point reprojects inside the reference image and the two depths agree within 5\%. Pixels whose reprojection falls on a hole in the reference depth map are \emph{excluded} rather than counted as negatives, which would otherwise reward a model for predicting low co-visibility on missing data.

We evaluate on all 1500 pairs of ScanNet-1500 and on 558 retrieval pairs of 7-Scenes~\cite{shotton2013scene}. Neither benchmark contributes a scene to the mix of~\Cref{sec:mix}. Average precision and ROC-AUC are threshold-free. For balanced accuracy we fit a single one-dimensional logistic regression per dataset and per score over all evaluated pixels and threshold it at $p=0.5$; the question is whether \emph{one} global threshold separates the classes at all, so the threshold is fitted on the same pixels it is reported on. The CroCo baseline has no fourth channel, so it is scored by $-\ell_{\text{CroCo}}(\bp)$ from a masked forward pass.

The tertile analysis of~\Cref{tab:covis_tertile} sorts the 1500 pairs by their true co-visible fraction and splits them into three equal groups; chance average precision is the positive rate within each group and therefore differs per tertile.

Two remarks. First, the pseudo-label figure quoted in~\Cref{sec:covis_probe} ($0.555$ AP) is $\m{C}(\bp)$ recomputed from the \emph{same trained model} under its 90\% training mask, averaged over mask draws. It is not directly comparable to the $0.74$ of~\Cref{fig:hist}, which is measured on a different set of 100 ScanNet-50 pairs using separately pre-trained CroCo and MAE models (\Cref{sec:details_hist}). Second, restricting the evaluation to SIFT keypoints raises every score, and raises the raw signals more than the prediction ($\hat{\m{C}}$ $0.763\rightarrow0.790$, $\ell_{\text{CroCo}}$ $0.576\rightarrow0.587$): the textureless-region limitation of~\Cref{sec:limitations} showing up exactly where it should.

\section{Probing protocol for~\Cref{tab:vitl}}\label{sec:vitl_details}
All rows of~\Cref{tab:vitl} share the head architecture, the fine-tuning data and the step budget of the corresponding main-paper experiment; only the frozen backbone is swapped. Parameter counts differ substantially: ${\sim}390$M for~\ourmethod-L and MuM-L, ${\sim}420$M for CroCo~v2-L, and ${\sim}1$B for VGGT. Pose uses the MLP head of~\Cref{sec:training_details} on frozen decoder features; pointmap uses the DPT head; optical flow uses a DPT flow head with the tiled inference of CroCo-Flow at $0.9$ overlap on the MPI-Sintel~\cite{butler2012naturalistic} \texttt{subval} split.

Two caveats. The self-supervised rows are not data-matched: CroCo~v2 is pre-trained on 7.3M curated stereo and flow pairs and MuM on a corpus that includes ImageNet-1K, whereas~\ourmethod-L$^{\text{mix}}$ sees only the raw video of~\Cref{sec:mix}. And MuM's frozen rows use a learning rate of $10^{-4}$, since the rate used for the other rows does not train its head at all. VGGT is reported for scale only: it is trained with explicit camera and depth supervision, and both ScanNet and DL3DV are in its training corpus, so three of its four columns are in-distribution.

\section{Data scaling on the raw-video mix}\label{sec:data_scaling}
To test whether the relative improvement signal keeps paying off as data grows, we pre-train matched pairs of Base models on nested, seeded, scene-level subsets of the mix of~\Cref{sec:mix}, from 1\% to 100\% of its scenes. Sources, sampling weights, resolution, global batch size (768) and step count (100k) are identical across all rows, as is the fine-tuning budget of every head.

\begin{table}[h]
    \centering
    \caption{\textbf{Data scaling.} ScanNet-1500 pose ($\uparrow$, at $10^\circ$/0.25\,0.5\,1m) and ScanNet pointmap Chamfer Overall ($\downarrow$) as a function of the fraction of mix scenes used for pre-training. Best per column in \textbf{bold}.}
    \label{tab:data_scaling}
    \resizebox{0.9\linewidth}{!}{%
    \begin{tabular}{lcccc}
        \toprule
        \multirow{2}{*}{Pre-training data} & \multicolumn{2}{c}{Pose ($\uparrow$)} & \multicolumn{2}{c}{Pointmap ($\downarrow$)} \\
        \cmidrule(lr){2-3} \cmidrule(lr){4-5}
        & CroCo-B & \ourmethod-B & CroCo-B & \ourmethod-B \\
        \midrule
        1\%   & 7.3 / 15.7 / 20.7  & \textbf{15.1 / 24.7 / 29.0} & 0.144 & \textbf{0.140} \\
        5\%   & 17.7 / 32.9 / 39.3 & \textbf{25.1 / 40.8 / 46.3} & 0.126 & \textbf{0.121} \\
        10\%  & 15.2 / 26.6 / 33.4 & \textbf{37.0 / 54.1 / 58.7} & 0.118 & \textbf{0.110} \\
        25\%  & 17.4 / 33.7 / 42.0 & \textbf{41.5 / 57.5 / 63.2} & 0.108 & \textbf{0.103} \\
        50\%  & 22.5 / 39.3 / 46.9 & \textbf{37.9 / 56.5 / 61.4} & 0.104 & \textbf{0.101} \\
        100\% & 15.1 / 30.5 / 39.2 & \textbf{34.4 / 51.5 / 56.5} & 0.102 & \textbf{0.096} \\
        \bottomrule
    \end{tabular}%
    }
\end{table}

\ourmethod leads at every scale, by between $+7.4$ and $+24.1$ points at the strictest pose threshold. Two honest caveats. Above 50\% both methods dip on pose at a fixed 100k-step budget, which indicates the backbones are undertrained rather than that data hurts; extending the best mix model to 200k steps recovers $39.4/56.1/61.1$ and a pointmap Overall of $0.092$. And composition matters more than size on an indoor benchmark: an indoor-only mix gives~\ourmethod $39.4/59.2/65.5$ against $34.4/51.5/56.5$ for the full mix, while CroCo is essentially flat ($13.7/29.9/38.6$ against $15.1/30.5/39.2$). Adding outdoor and object-centric video buys generality at a measurable in-domain cost.

\section{Generality across architectures}\label{sec:arch_generality}
The gain could conceivably be an artefact of CroCo's specific encoder--decoder. We therefore pre-trained six Base models from scratch on ScanNet-50 for 100k steps at global batch 768, varying the architecture and keeping everything else fixed.

\begin{table}[h]
    \centering
    \caption{\textbf{The advantage survives substituting either half of the architecture.} ScanNet-1500 pose ($\uparrow$) and ScanNet pointmap Chamfer Overall ($\downarrow$). All models Base, ScanNet-50, 100k steps, matched head budgets.}
    \label{tab:arch_generality}
    \resizebox{\linewidth}{!}{%
    \begin{tabular}{lcccc}
        \toprule
        \multirow{2}{*}{Architecture} & \multicolumn{2}{c}{Pose ($\uparrow$)} & \multicolumn{2}{c}{Pointmap ($\downarrow$)} \\
        \cmidrule(lr){2-3} \cmidrule(lr){4-5}
        & CroCo & \ourmethod & CroCo & \ourmethod \\
        \midrule
        CroCo encoder + cross-attention decoder (main paper) & 5.5 / 12.9 / 18.5 & \textbf{29.1 / 43.3 / 47.5} & 0.125 & \textbf{0.100} \\
        CroCo encoder + joint self-attention decoder & 5.7 / \phantom{0}9.9 / 13.9 & \textbf{22.9 / 34.2 / 38.3} & 0.207 & \textbf{0.105} \\
        Frozen DINOv3~\cite{simeoni2025dinov3} encoder + cross-attention decoder & 5.0 / 10.3 / 14.7 & \textbf{23.9 / 36.8 / 41.3} & 0.125 & \textbf{0.096} \\
        \bottomrule
    \end{tabular}%
    }
\end{table}

The joint self-attention decoder concatenates the two views' tokens with a learned per-view embedding and uses no cross-attention anywhere; the DINOv3 variant freezes an off-the-shelf monocular encoder and trains only the decoder. The advantage survives both substitutions ($+17.2$ and $+18.9$ points against $+23.6$ for the paper's architecture). It is also worth noting that the CroCo baseline barely moves across the three architectures ($5.0$ to $5.7$), while~\ourmethod changes substantially: the training signal, not the architecture, is what varies. As a control, a frozen DINOv3 encoder with no cross-view pre-training at all reaches $6.9/14.5/21.2$ on pose and $0.142$ on pointmap.

\section{Single-image probes}\label{sec:mono}
The relative improvement is a binocular signal, so we do not expect it to help tasks that see one image. It does not, and it does not hurt either. Under the frozen protocol of the main paper with a DPT head, monocular depth on NYUv2~\cite{silberman2012indoor} (REL $\downarrow$ / $\delta_1$ $\uparrow$) and semantic segmentation on ADE20K~\cite{zhou2017scene} (mIoU $\uparrow$ / pixel accuracy $\uparrow$) give:

\begin{table}[h]
    \centering
    \caption{\textbf{Single-image probes are a tie.} Frozen backbone, DPT head, matched schedules. All models Base.}
    \label{tab:mono}
    \resizebox{0.92\linewidth}{!}{%
    \begin{tabular}{lcccc}
        \toprule
        \multirow{2}{*}{Pre-training} & \multicolumn{2}{c}{NYUv2 depth} & \multicolumn{2}{c}{ADE20K segmentation} \\
        \cmidrule(lr){2-3} \cmidrule(lr){4-5}
        & CroCo-B & \ourmethod-B & CroCo-B & \ourmethod-B \\
        \midrule
        ScanNet-50  & \textbf{0.181} / \textbf{73.5} & 0.191 / 72.3 & 17.6 / 71.2 & 17.6 / \textbf{71.5} \\
        ScanNet-all & 0.200 / \textbf{70.9} & 0.200 / 70.5 & \textbf{17.6} / \textbf{71.3} & 16.8 / 70.9 \\
        12-source mix, 200k & 0.147 / 80.8 & \textbf{0.143} / \textbf{82.2} & \textbf{20.8} / 73.6 & 20.2 / \textbf{73.8} \\
        \bottomrule
    \end{tabular}%
    }
\end{table}

The same holds for ImageNet-1K~\cite{deng2009imagenet} classification with an attentive probe (50 epochs, batch 2048): $61.3$ top-1 for~\ourmethod-B$^{\text{mix}}$ against $61.0$ for its matched CroCo-B control, a tie. On the eight dense Taskonomy~\cite{zamir2018taskonomy} tasks under the standard 1k-image transfer protocol,~\ourmethod-B$^{\text{mix}}$ wins three columns (2D texture edges $0.0029$, occlusion edges $0.0006$, principal curvature $0.0441$) and loses the other five to the much larger released CroCo~v2-L and MuM-L. We conclude that the relative improvement objective buys cross-view geometry and nothing else, and we make no claim beyond that.

\section{Full-network fine-tuning}\label{sec:fullft}
Every table in the main paper probes a frozen backbone. Fine-tuning the whole network measures a different property, its value as an initialisation, and it ranks models differently. We report it here because the distinction matters for how our claim should be read.

\paragraph{Optical flow.} On MPI-Sintel, training the whole network at CroCo-Flow's own sample budget (110k steps at effective batch 64, i.e.\ 7.0M of their 7.2M sample-passes) gives AEPE clean/final of $2.02/2.72$ for~\ourmethod-B$^{\text{mix}}$, $2.21/2.59$ for its matched CroCo-B control, $2.59/3.20$ for the released MuM-L, and $1.48/2.15$ for the released CroCo~v2-L. The matched Base pair is a tie, and CroCo~v2-L is clearly ahead of everything. This is consistent with the corpus rather than the objective driving this benchmark: CroCo~v2 is pre-trained on 7.3M curated stereo and flow pairs, and the architecture-matched figure it publishes is $1.76/2.30$. It is also a validation of our pipeline: the released CroCo~v2 checkpoint reaches $1.48$ in our hands against $1.43$ published, a $3\%$ gap explained by four documented deviations (110k of 112.5k steps, effective batch 64 rather than 8 with the learning rate scaled linearly, bf16 rather than fp32, and the last checkpoint rather than the best on validation).

\paragraph{Relative pose on 7-Scenes.} Fine-tuning the whole network at $224\times224$ for 6k steps gives median position/orientation errors of $6.16$\,cm$/1.90^\circ$ for~\ourmethod-B$^{\text{mix}}$ against $7.70$\,cm$/2.08^\circ$ for its matched CroCo-B control, i.e.\ $20\%$ better on position and $9\%$ on orientation. The released ViT-L models are ahead of both Base models ($5.77$\,cm for CroCo~v2-L, $5.29$\,cm for MuM-L) and we do not claim a win over them. Compared to the frozen numbers ($23.2$\,cm for~\ourmethod-B$^{\text{mix}}$ against $31.9$\,cm for CroCo-B), the position advantage survives fine-tuning while the rotation advantage largely does not.

Our pipeline reproduces CroCo~v1's published $5.0$\,cm on this benchmark to within $0.1$\,cm using the released CroCo~v2 checkpoint, despite our own optimiser settings and a DINOv3 global descriptor in place of the original retriever. The retrieved pairs are computed once and are byte-identical across all rows.

\section{Pointmap regression: full metrics}\label{sec:pointmap_full}
\begin{table}[ht]
    \centering
    \caption{\textbf{Pointmap regression, full metrics.} Accuracy, Completeness, and Overall metrics ($\downarrow$) for models pre-trained on DL3DV with curriculum and fine-tuned on ScanNet-all. \ourmethod consistently outperforms CroCo across all benchmarks, including out-of-domain evaluation on DL3DV and ETH3D. Best result per group in \textbf{bold}.}
    \label{tab:pointmap_full}
    \resizebox{\linewidth}{!}{%
    \begin{tabular}{llccccccccc}
        \toprule
        \multirow{2}{*}{Pre-training} & \multirow{2}{*}{Model} & \multicolumn{3}{c}{ScanNet} & \multicolumn{3}{c}{DL3DV} & \multicolumn{3}{c}{ETH3D} \\
        \cmidrule(lr){3-5} \cmidrule(lr){6-8} \cmidrule(lr){9-11}
        & & Acc. & Comp. & Over. & Acc. & Comp. & Over. & Acc. & Comp. & Over. \\
        \midrule
        \multirow{2}{*}{DL3DV-50} & CroCo-B & 0.128 & 0.125 & 0.126 & 1.278 & 1.583 & 1.430 & 0.400 & 0.399 & 0.399 \\
        & \ourmethod-B & \textbf{0.118} & \textbf{0.108} & \textbf{0.113} & \textbf{1.240} & \textbf{1.477} & \textbf{1.359} & \textbf{0.379} & \textbf{0.367} & \textbf{0.373} \\
        \midrule
        \multirow{2}{*}{DL3DV-all} & CroCo-B & 0.123 & 0.121 & 0.122 & 1.268 & 1.549 & 1.409 & 0.399 & 0.396 & 0.398 \\
        & \ourmethod-B & \textbf{0.119} & \textbf{0.110} & \textbf{0.115} & \textbf{1.230} & \textbf{1.503} & \textbf{1.366} & \textbf{0.376} & \textbf{0.369} & \textbf{0.373} \\
        \bottomrule
    \end{tabular}%
    }
\end{table}

\section{Details on the computation of the histograms in~\Cref{fig:hist}}\label{sec:details_hist}
The histograms in~\Cref{fig:hist} are computed on 100 image pairs from the ScanNet-50 set. For each pair, ground-truth co-visibility labels from Cub3~\cite{loiseau2025alligat0r} are used: a pixel is considered co-visible if its 3D point projects within the image boundaries of the other view and the depth is consistent (within 5\% tolerance). Both a pre-trained CroCo model and a pre-trained MAE model are run on each pair to obtain per-pixel reconstruction errors $\ell_{\text{CroCo}}(\mathbf{p})$ and $\ell_{\text{MAE}}(\mathbf{p})$, and the relative improvement $\mathbf{C}(\mathbf{p})$ is computed according to Eq.~\ref{eq:RelImp}.

\section{PCA visualization of encoder features}\label{sec:pca_features}
\begin{figure*}[t]
    \centering
    \setlength{\tabcolsep}{0pt}
    \begin{tabular}{@{}cccccc@{}}
        \makebox[0.166\linewidth][c]{\small Image 1} &
        \makebox[0.166\linewidth][c]{\small Image 2} &
        \makebox[0.166\linewidth][c]{\small CroCo PCA 1} &
        \makebox[0.166\linewidth][c]{\small CroCo PCA 2} &
        \makebox[0.166\linewidth][c]{\small \ourmethod PCA 1} &
        \makebox[0.166\linewidth][c]{\small \ourmethod PCA 2} \\
        \includegraphics[width=0.166\linewidth]{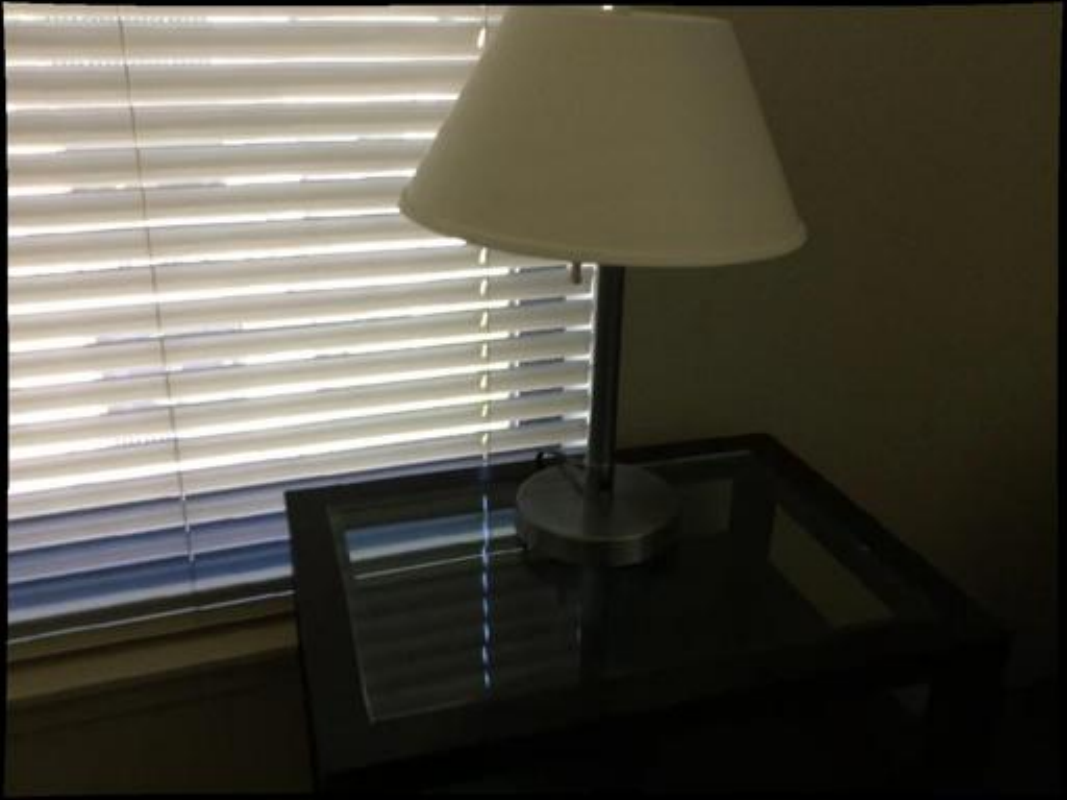} &
        \includegraphics[width=0.166\linewidth]{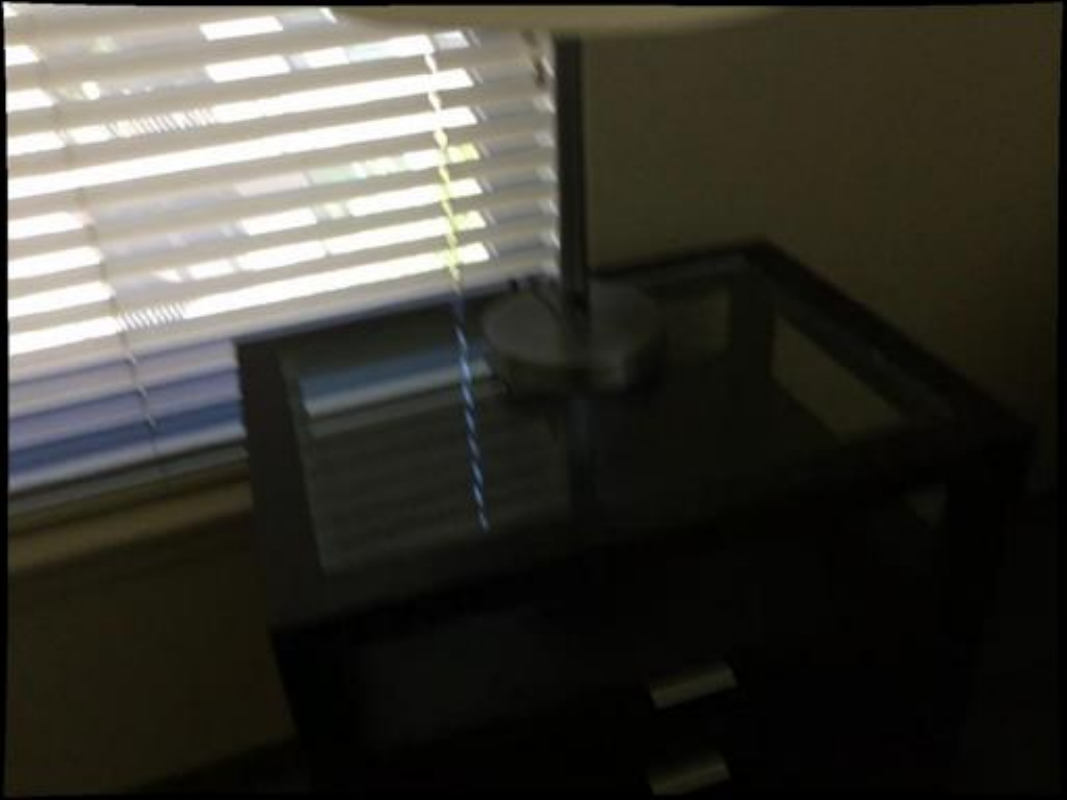} &
        \includegraphics[width=0.166\linewidth]{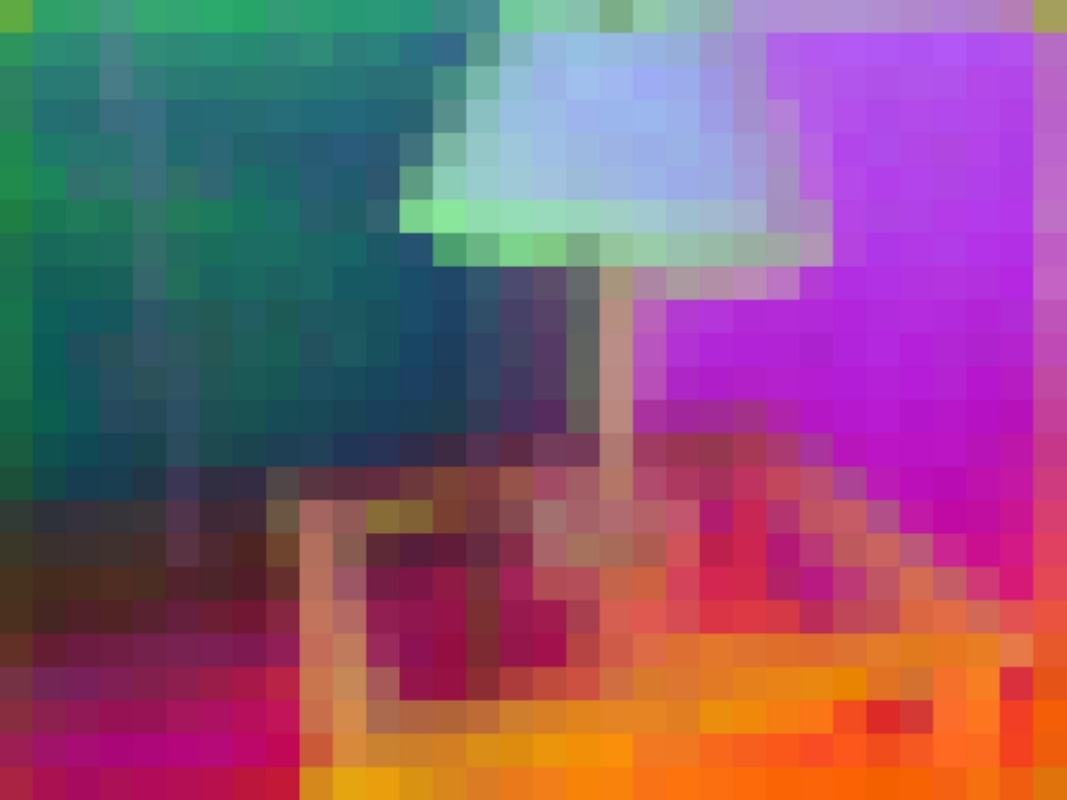} &
        \includegraphics[width=0.166\linewidth]{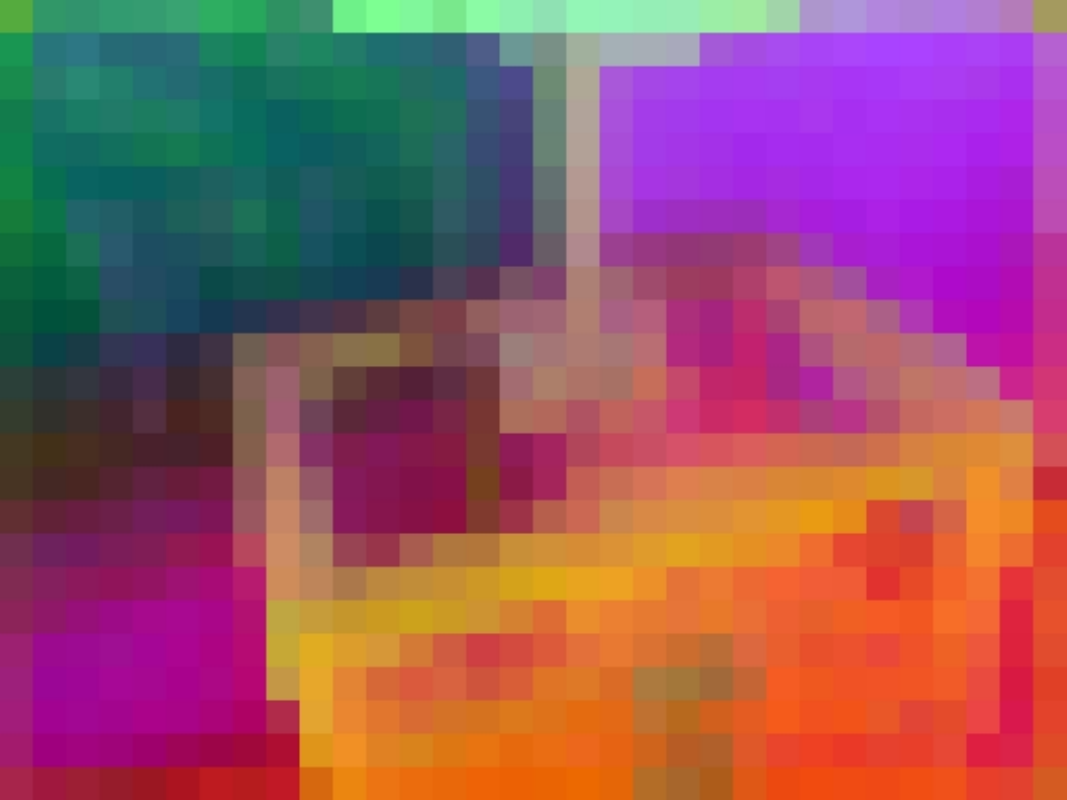} &
        \includegraphics[width=0.166\linewidth]{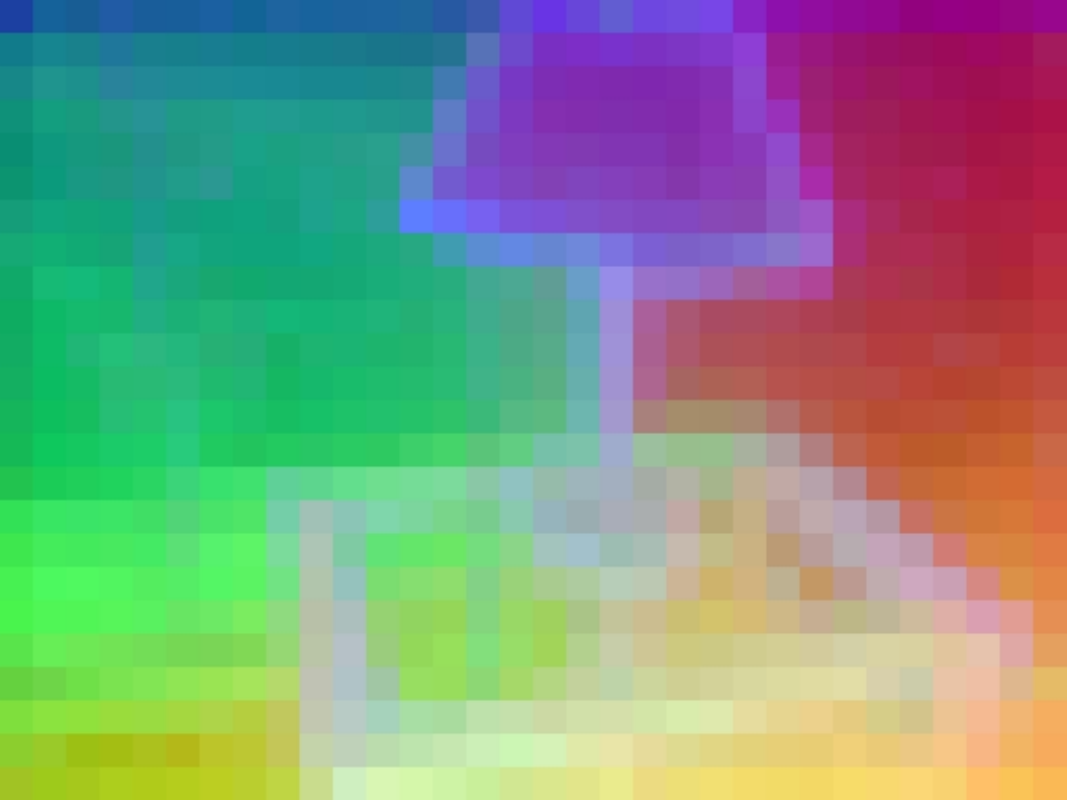} &
        \includegraphics[width=0.166\linewidth]{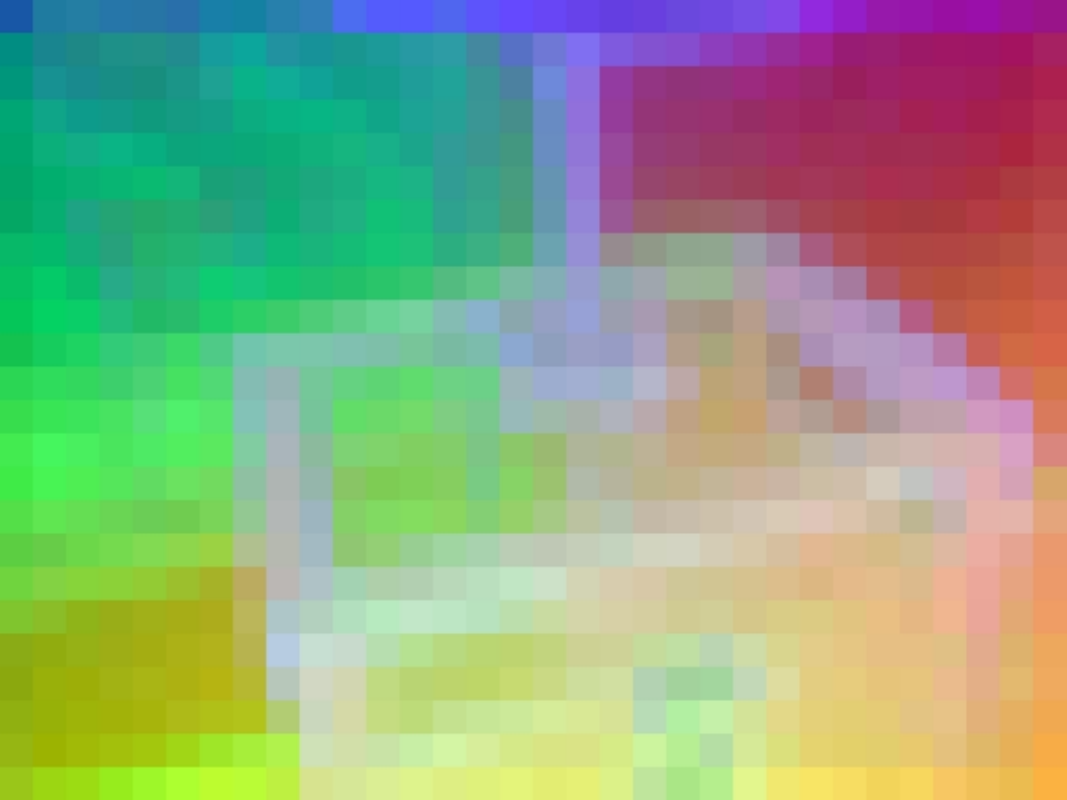} \\
        \includegraphics[width=0.166\linewidth]{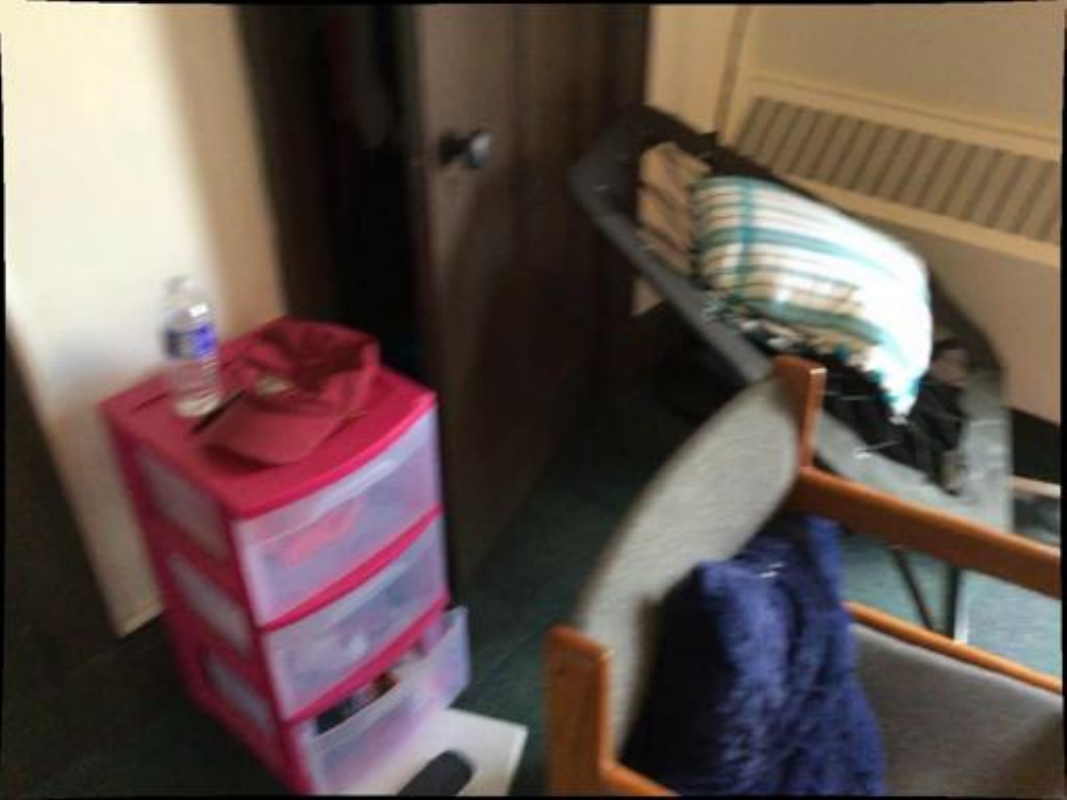} &
        \includegraphics[width=0.166\linewidth]{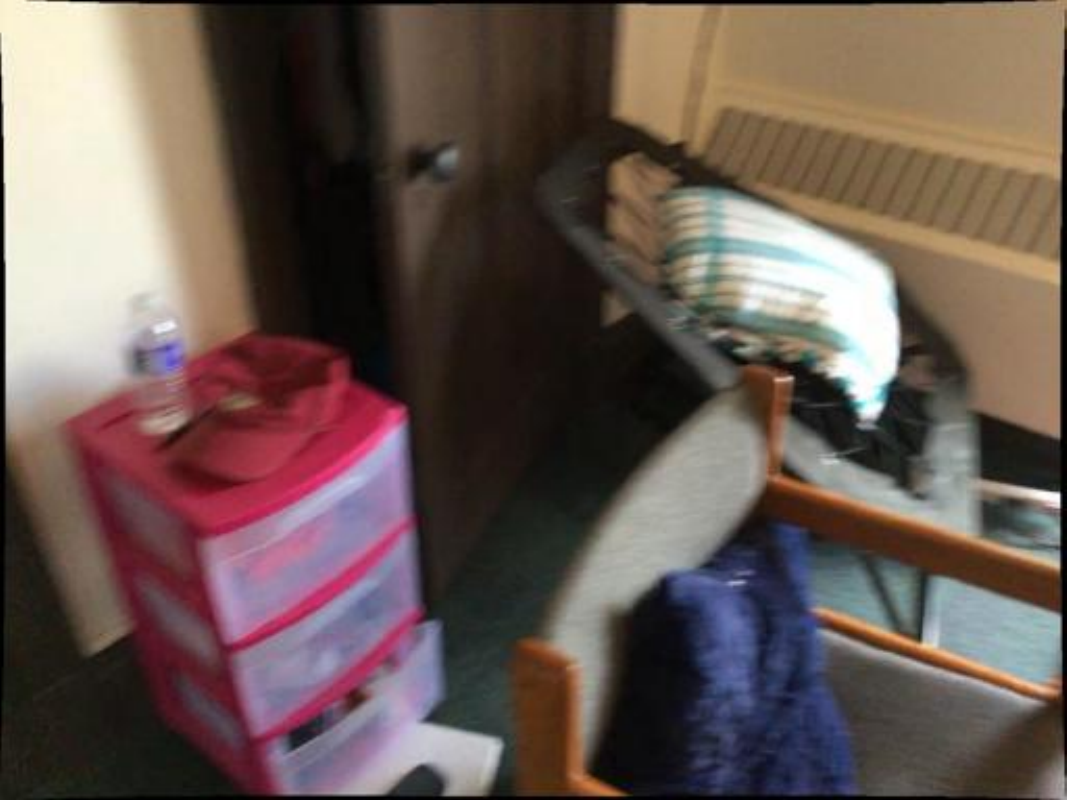} &
        \includegraphics[width=0.166\linewidth]{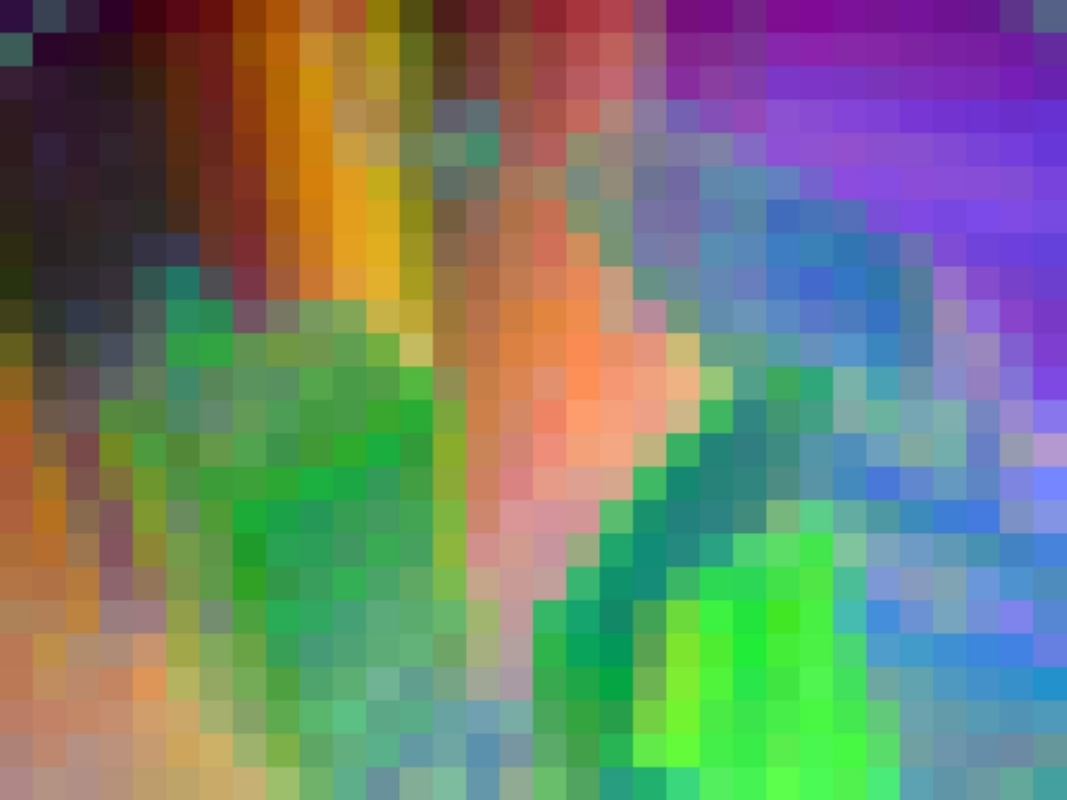} &
        \includegraphics[width=0.166\linewidth]{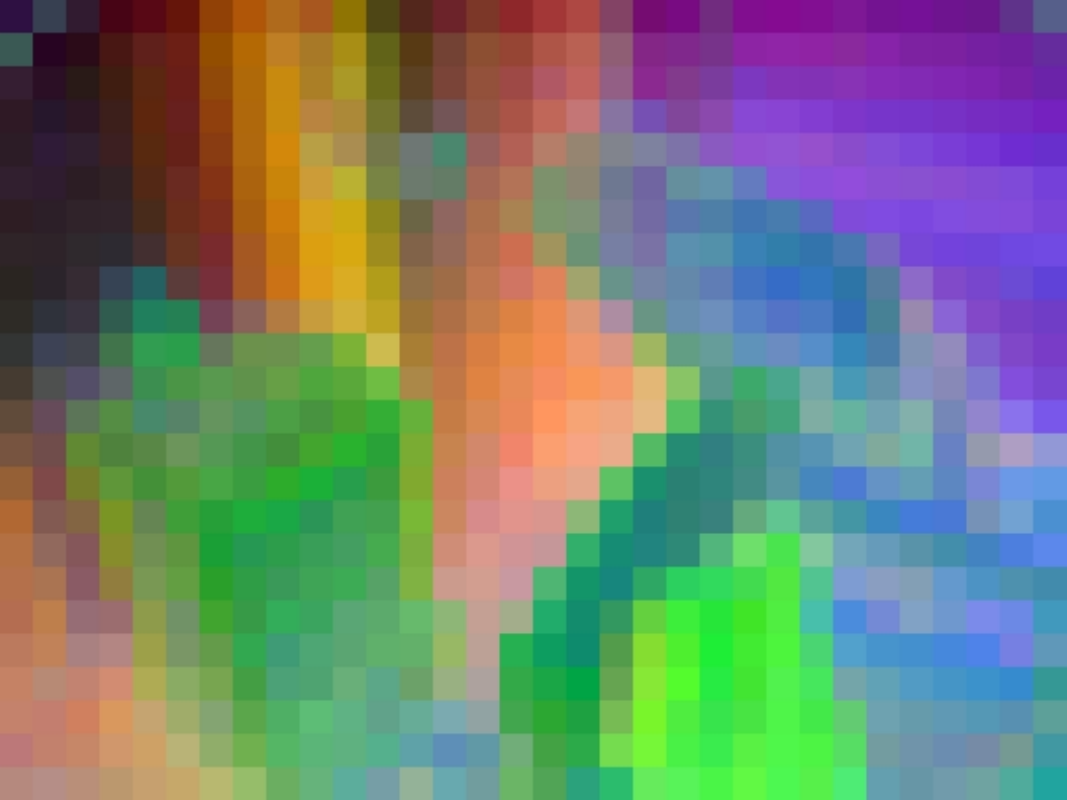} &
        \includegraphics[width=0.166\linewidth]{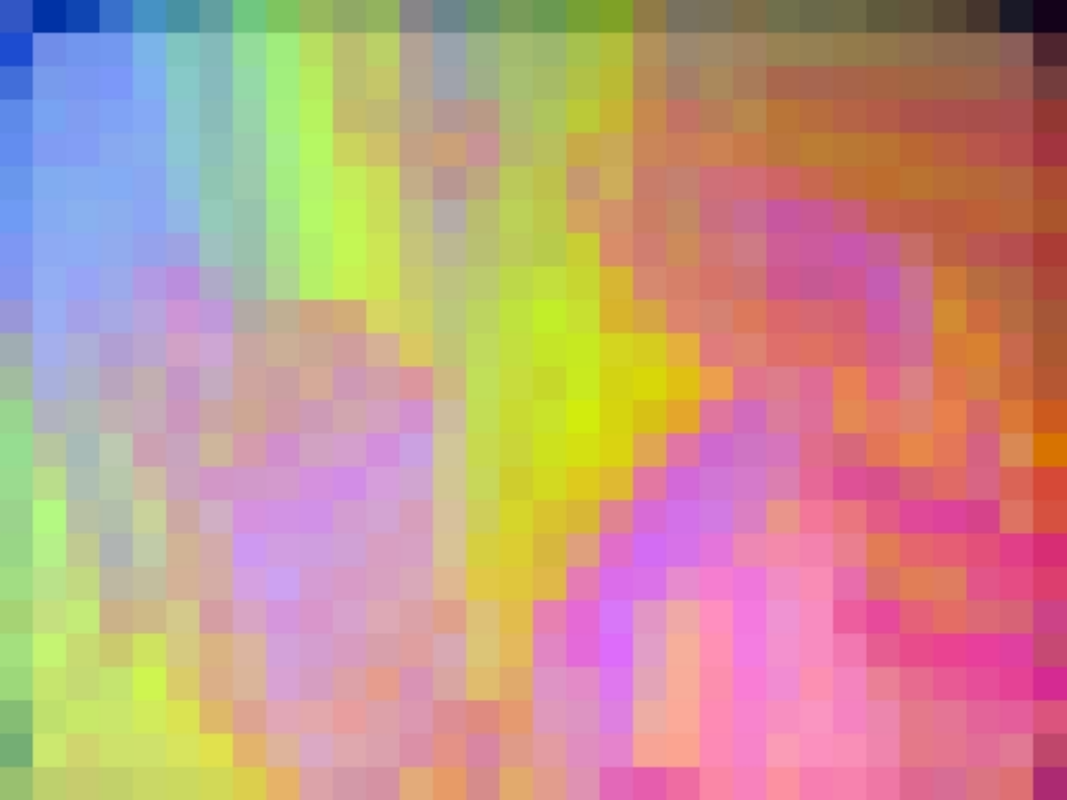} &
        \includegraphics[width=0.166\linewidth]{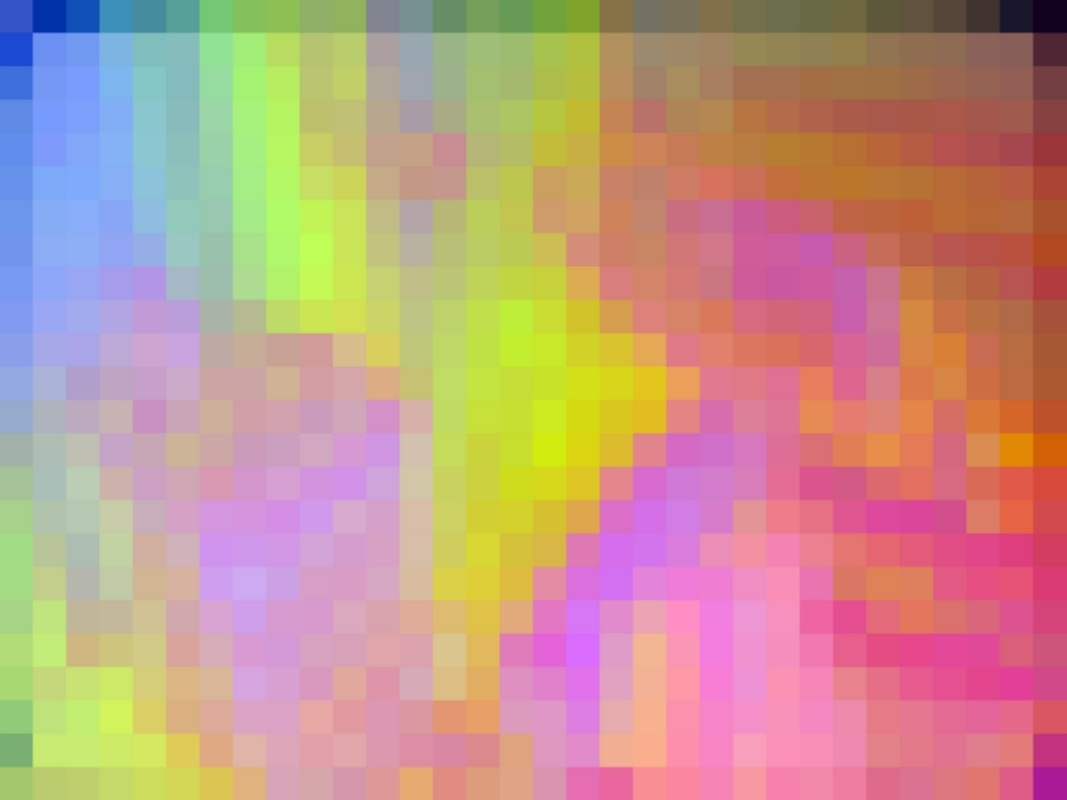} \\
        \includegraphics[width=0.166\linewidth]{figures/pairs_and_pca/203048/croco_img_1.pdf} &
        \includegraphics[width=0.166\linewidth]{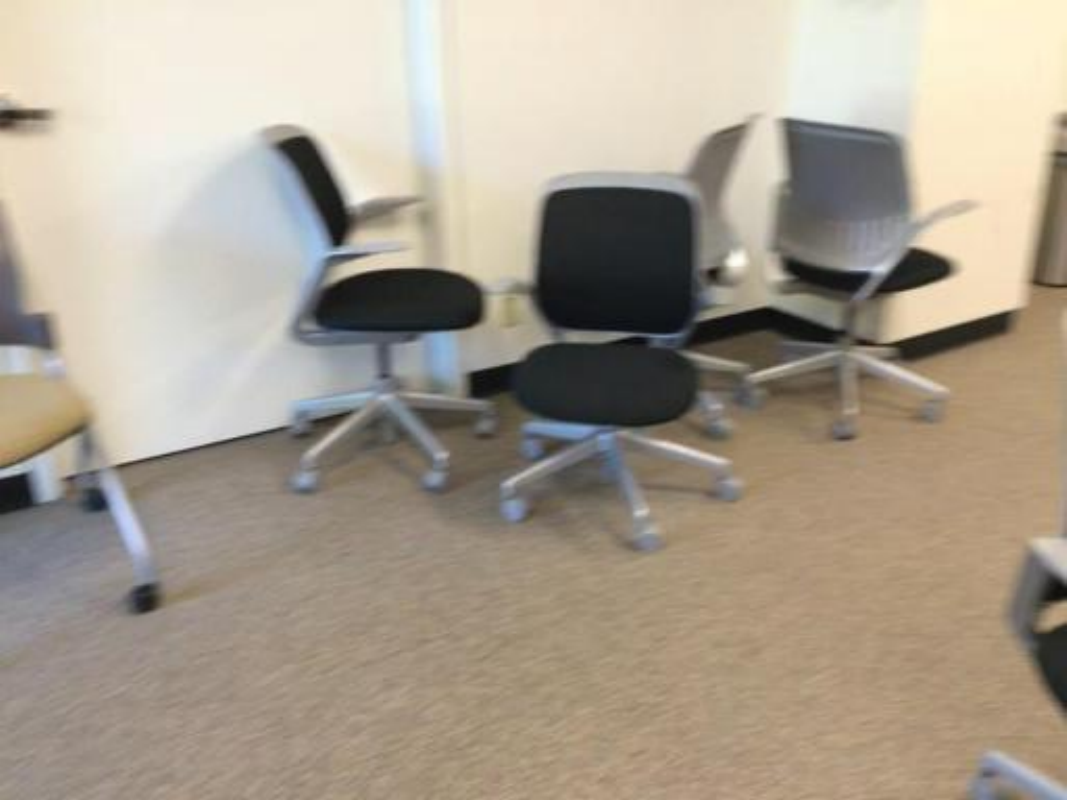} &
        \includegraphics[width=0.166\linewidth]{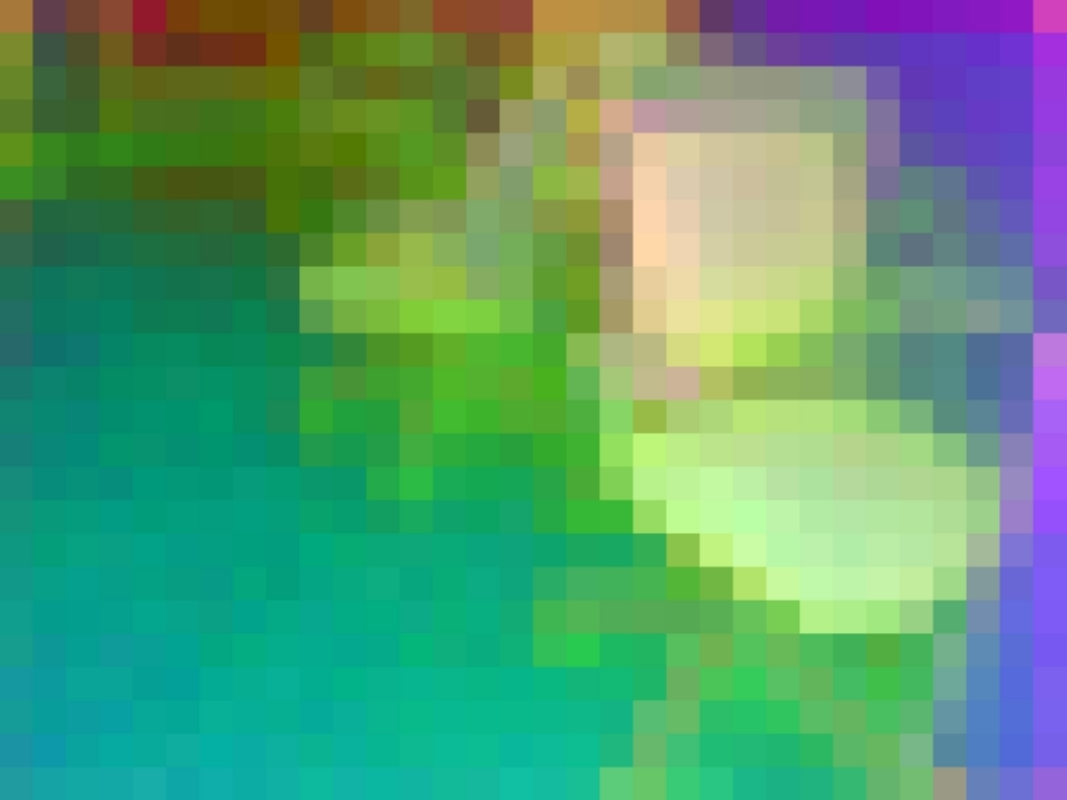} &
        \includegraphics[width=0.166\linewidth]{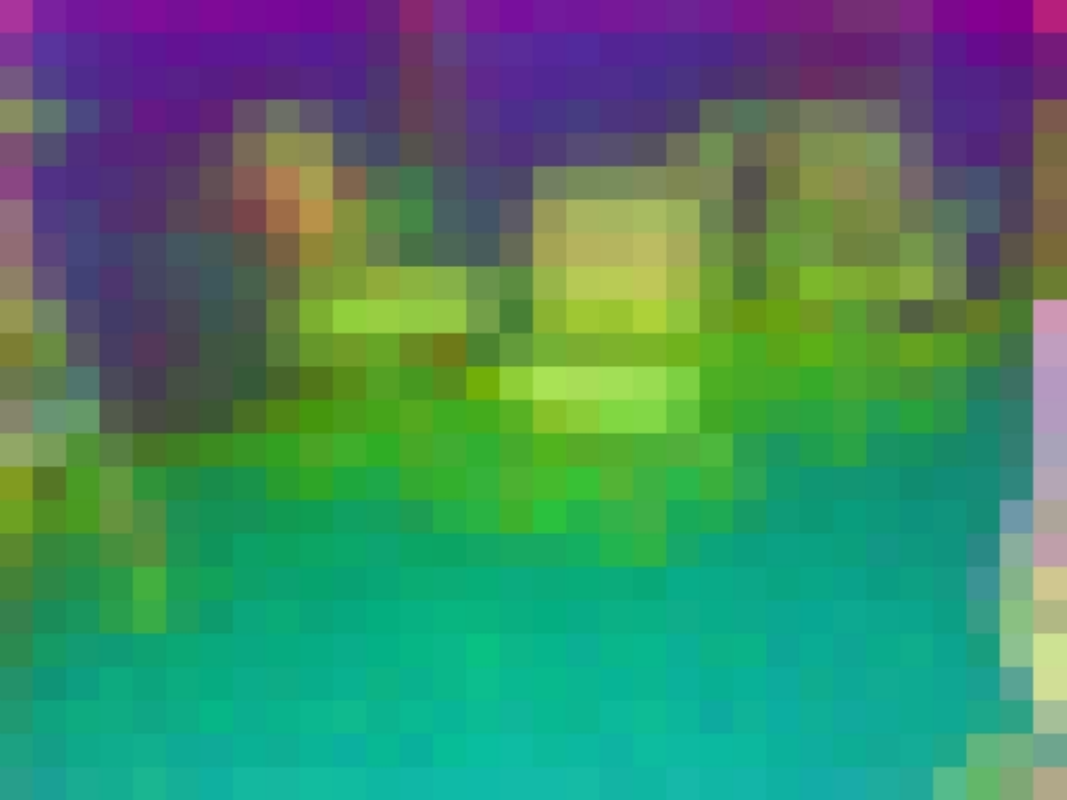} &
        \includegraphics[width=0.166\linewidth]{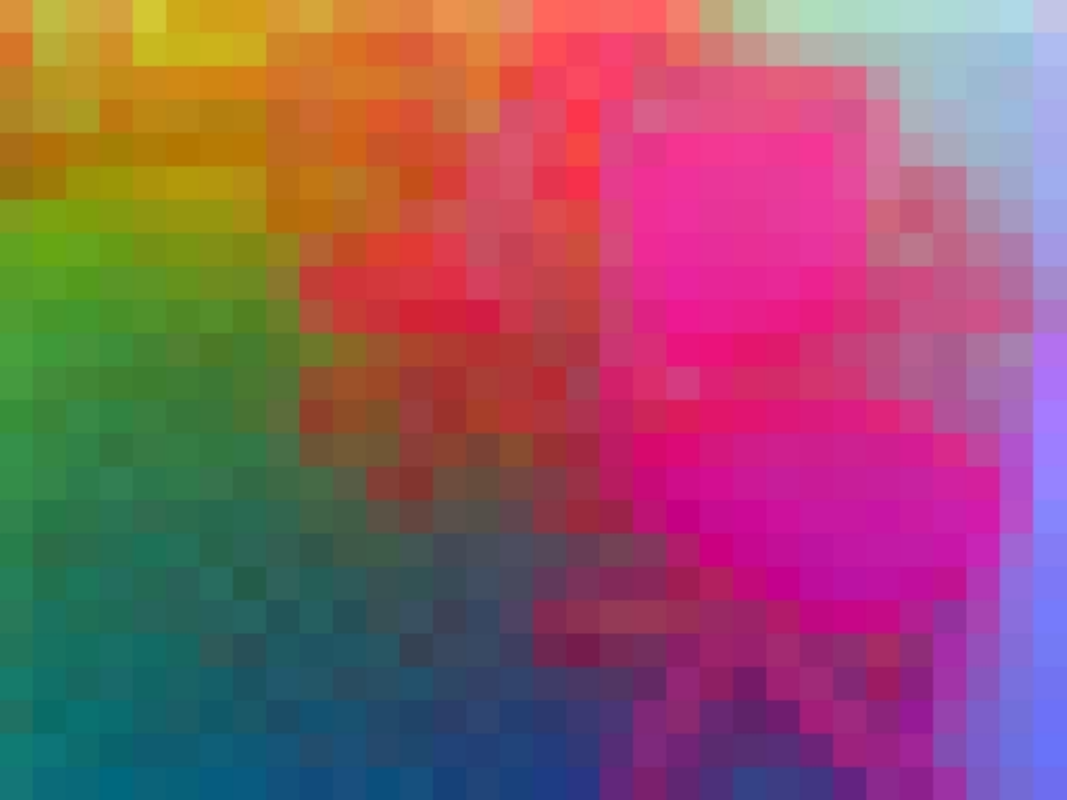} &
        \includegraphics[width=0.166\linewidth]{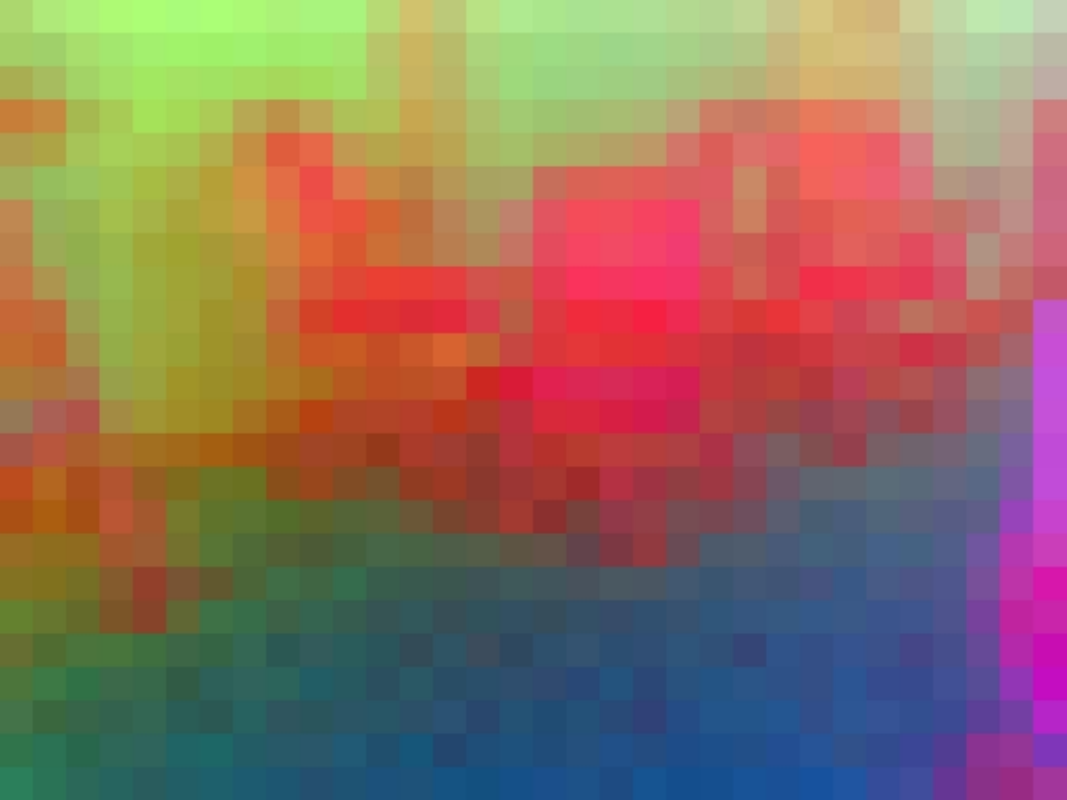} \\
        \includegraphics[width=0.166\linewidth]{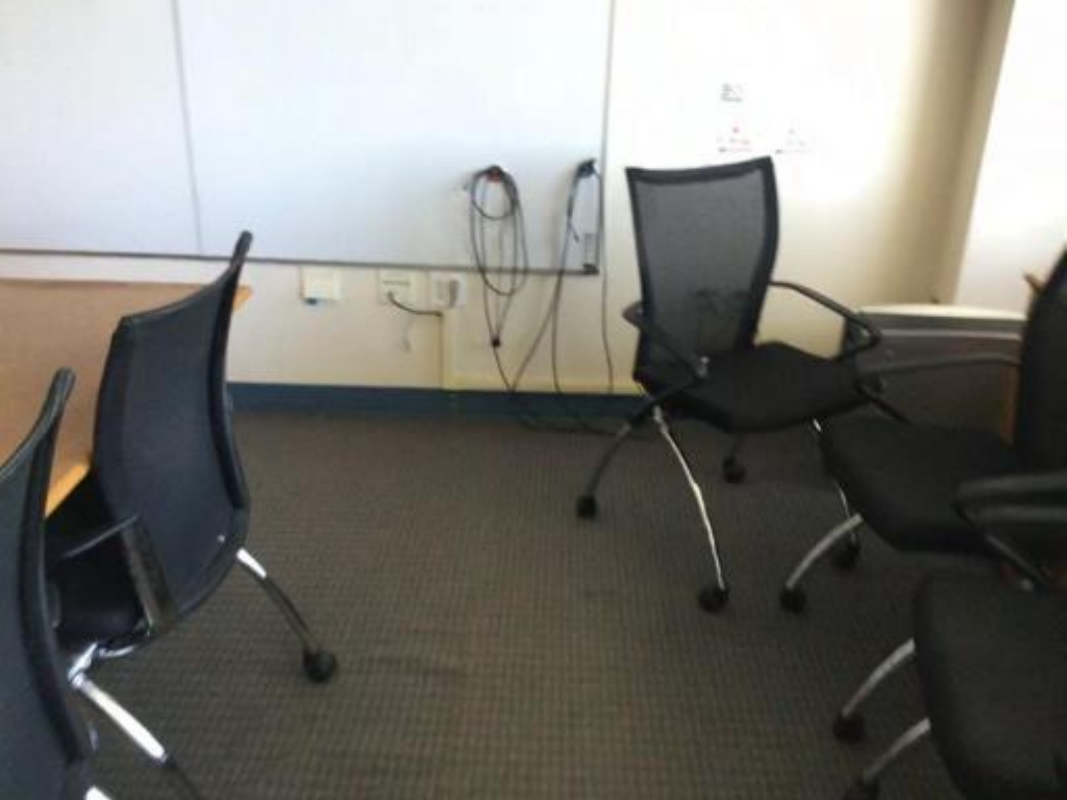} &
        \includegraphics[width=0.166\linewidth]{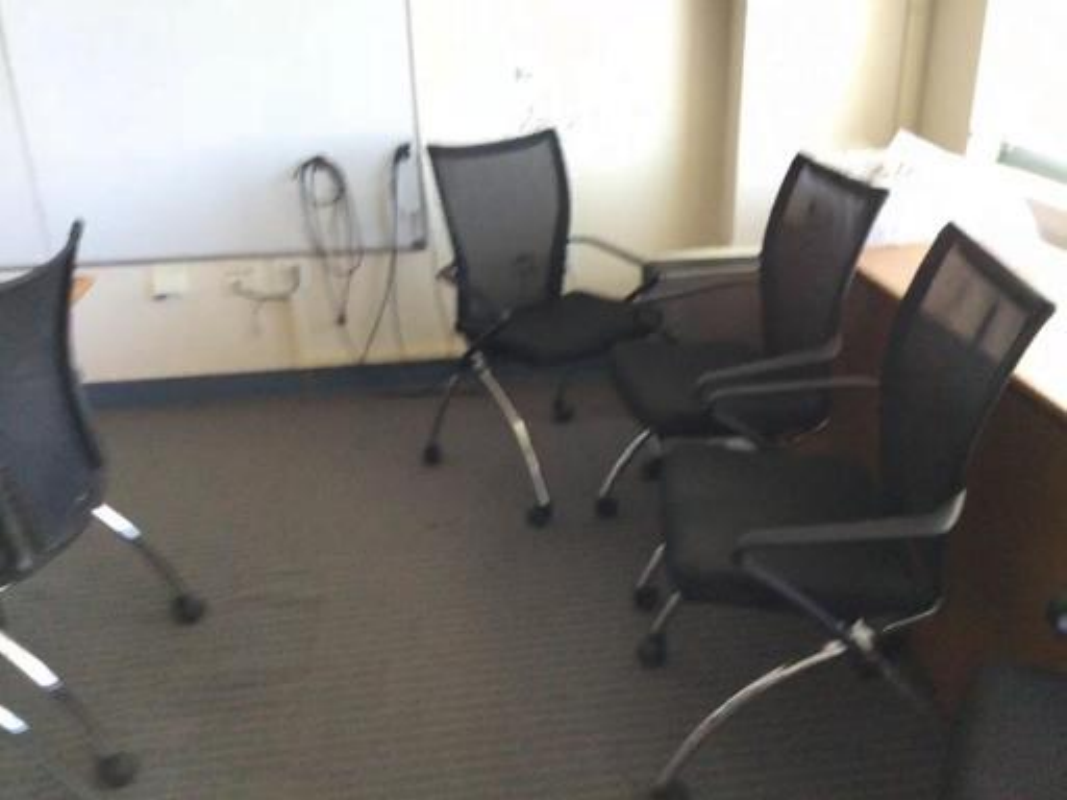} &
        \includegraphics[width=0.166\linewidth]{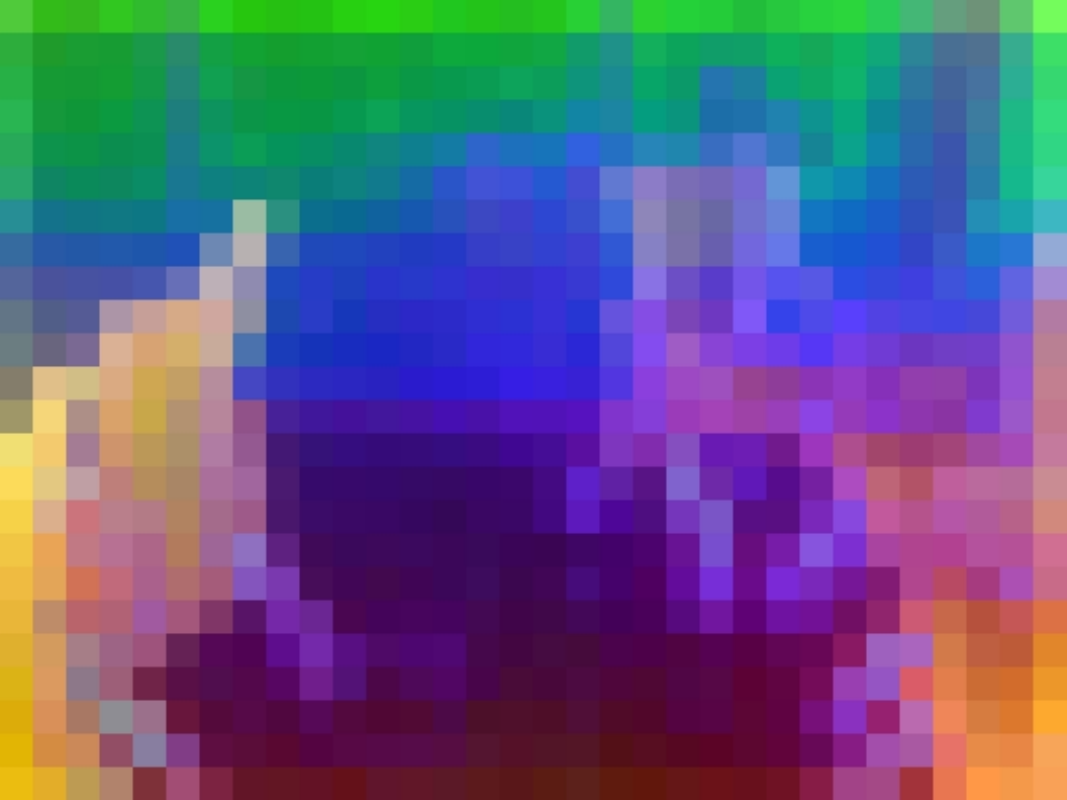} &
        \includegraphics[width=0.166\linewidth]{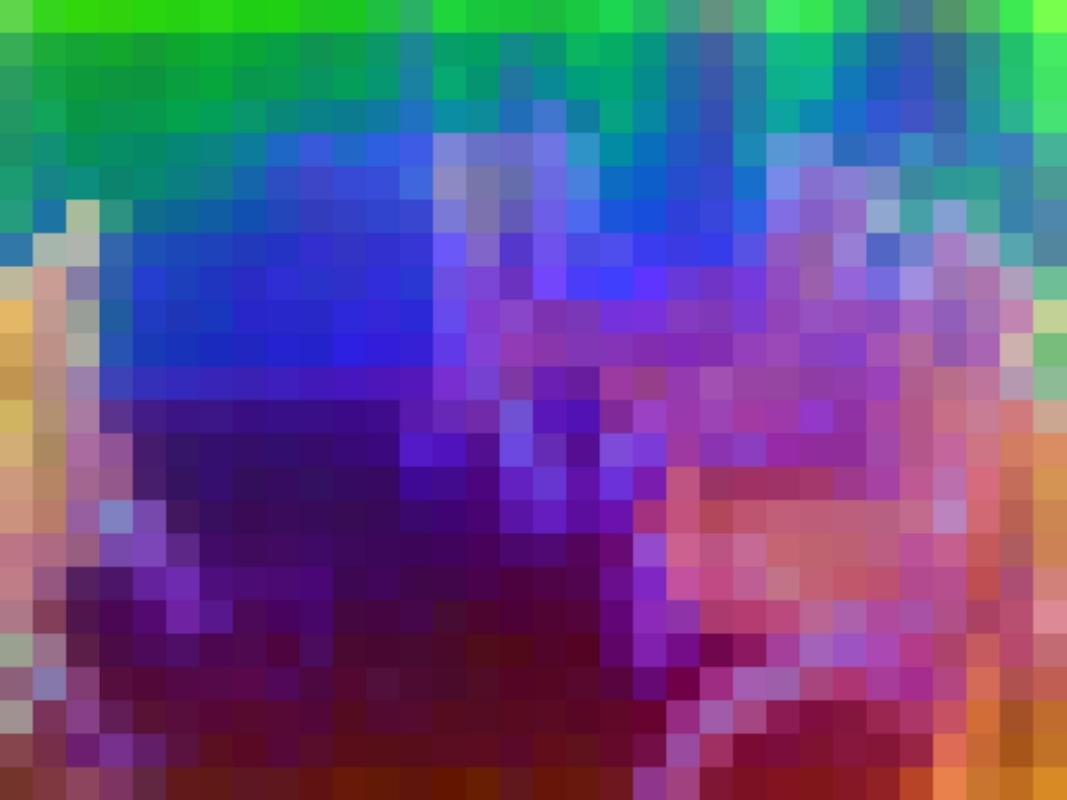} &
        \includegraphics[width=0.166\linewidth]{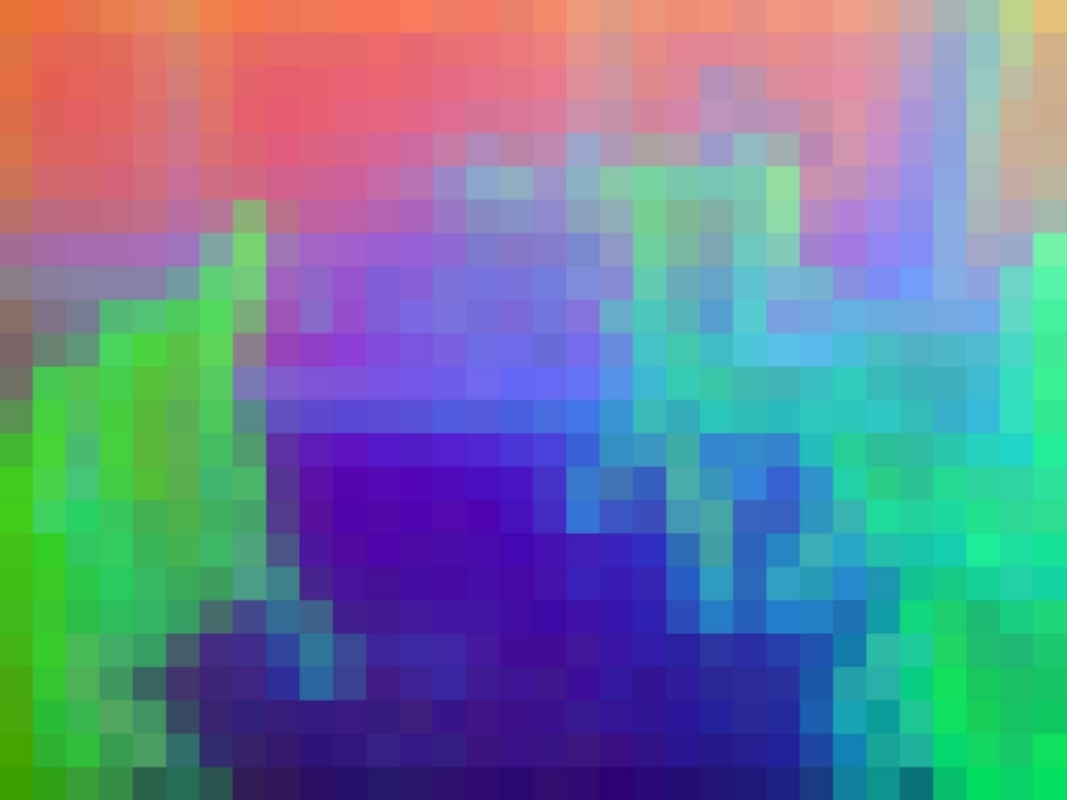} &
        \includegraphics[width=0.166\linewidth]{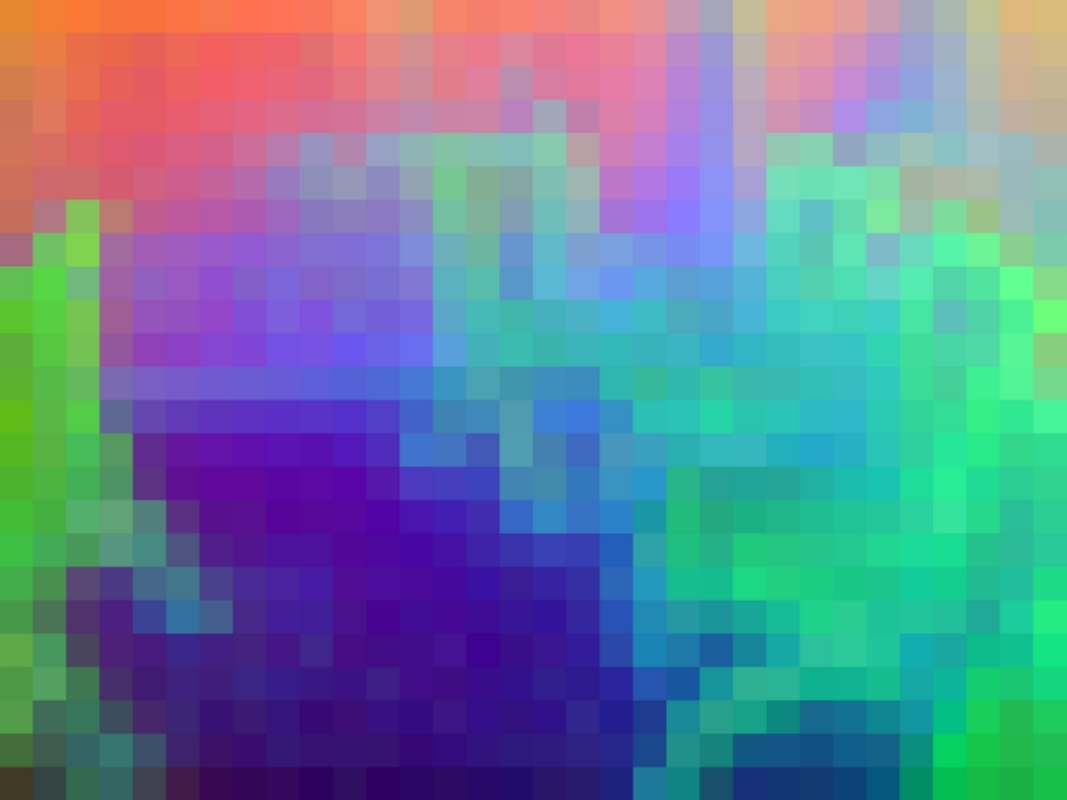} \\
        \includegraphics[width=0.166\linewidth]{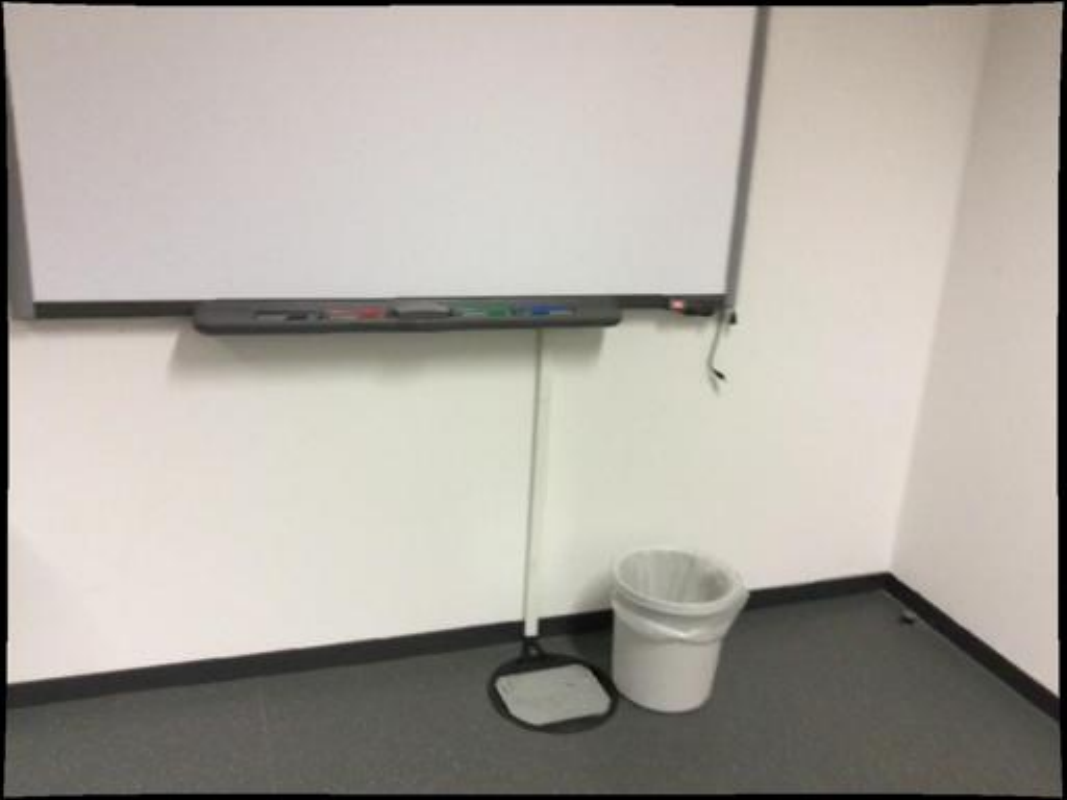} &
        \includegraphics[width=0.166\linewidth]{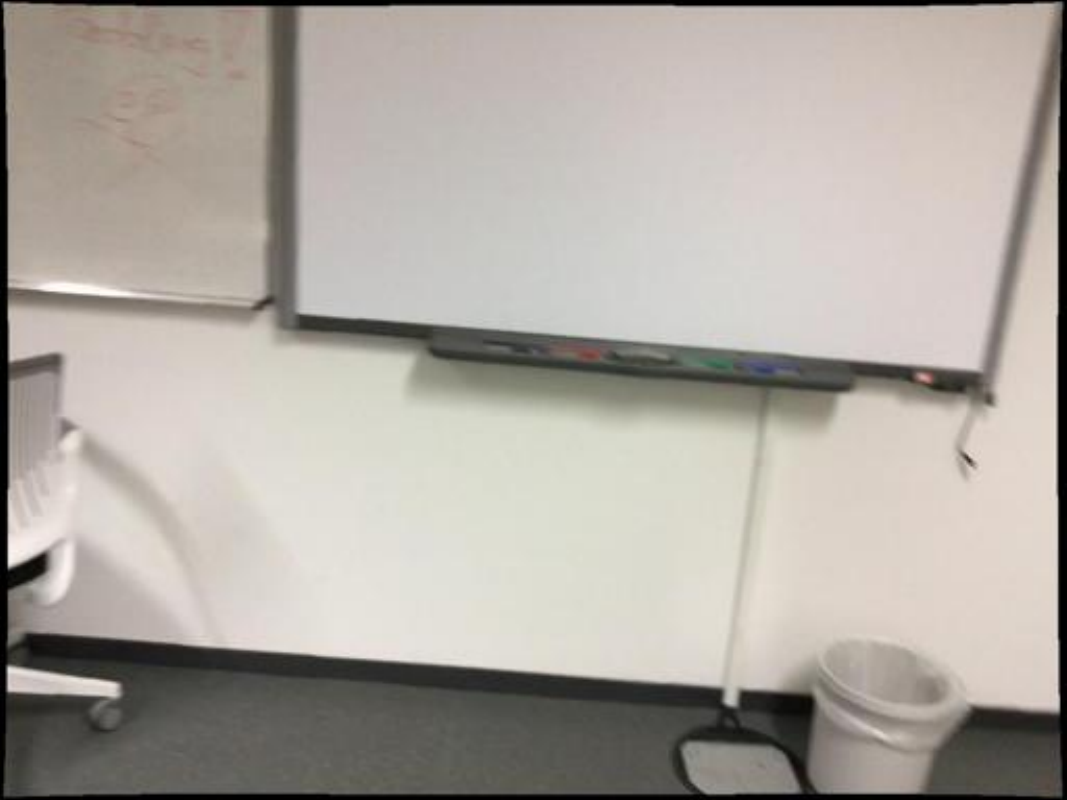} &
        \includegraphics[width=0.166\linewidth]{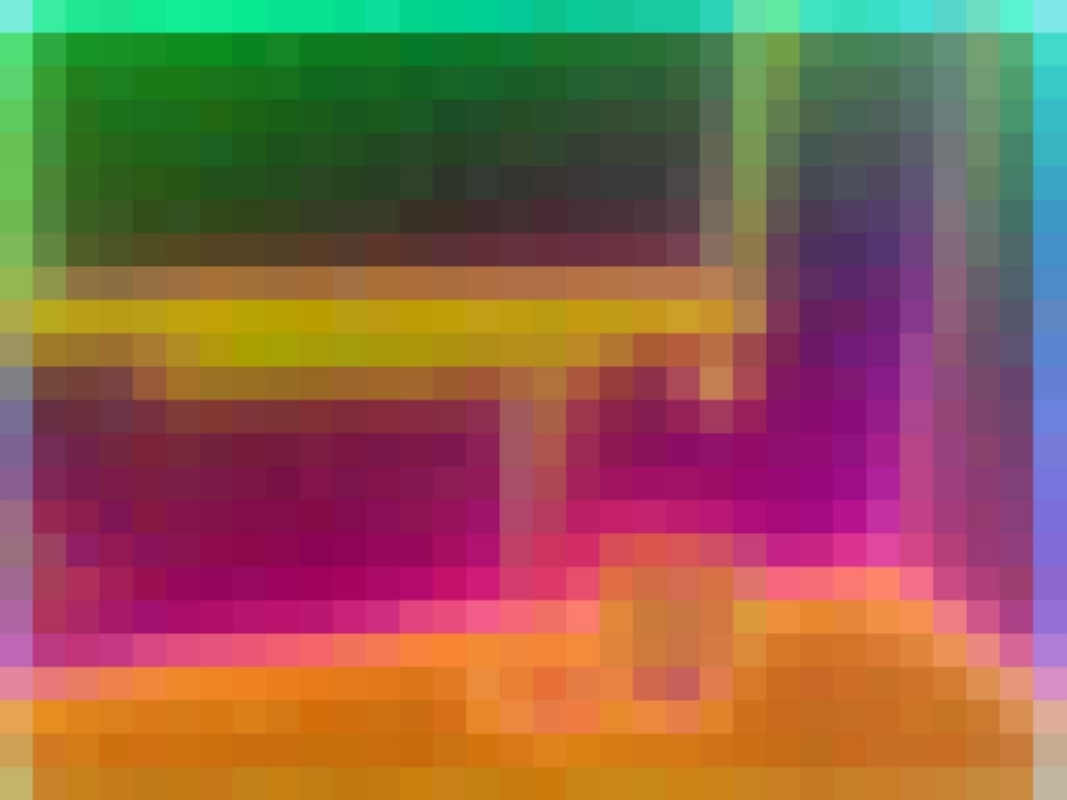} &
        \includegraphics[width=0.166\linewidth]{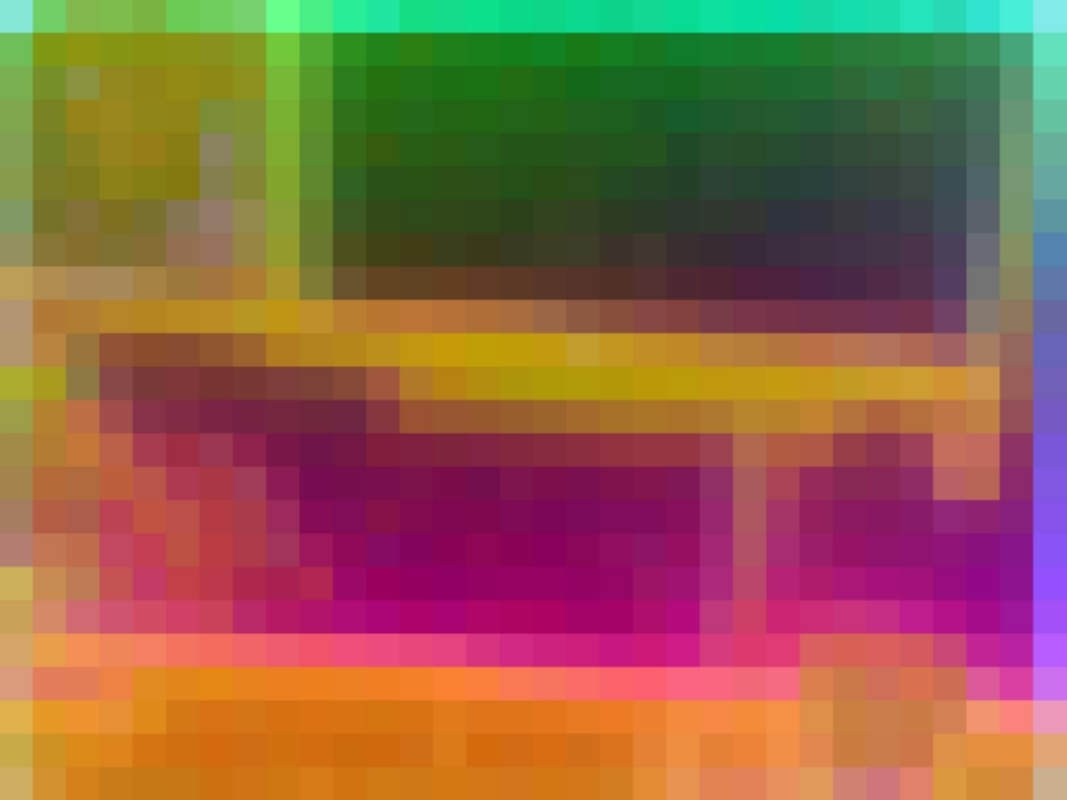} &
        \includegraphics[width=0.166\linewidth]{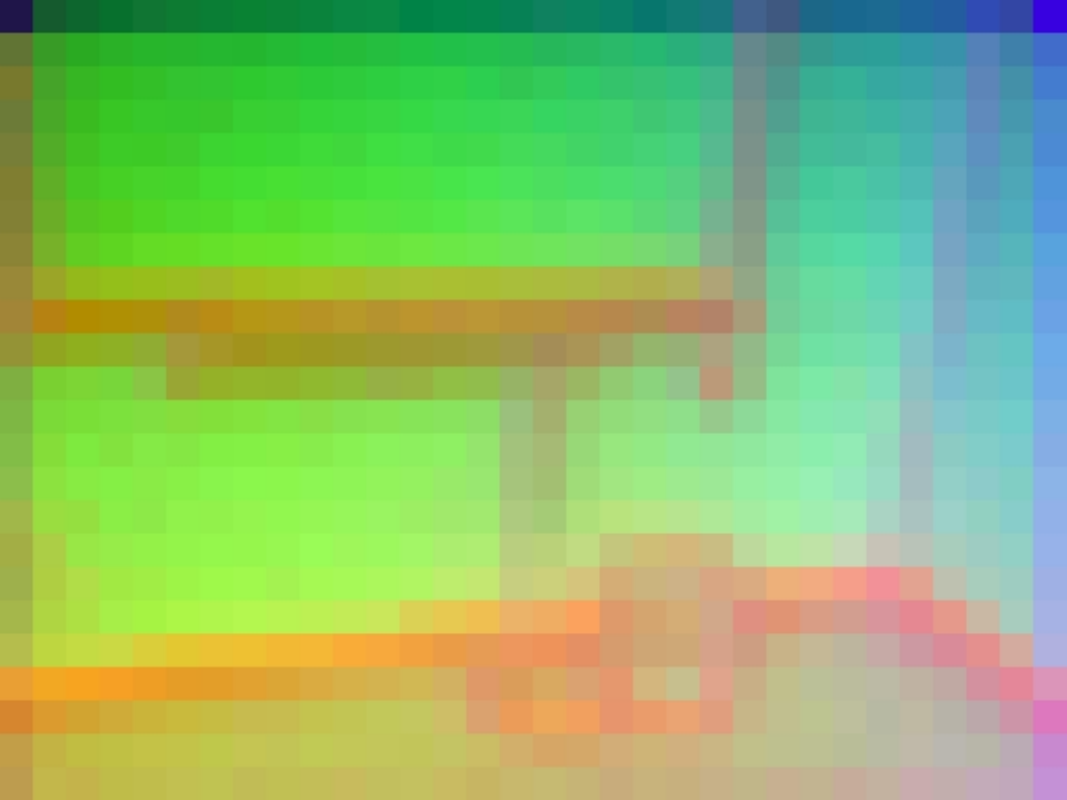} &
        \includegraphics[width=0.166\linewidth]{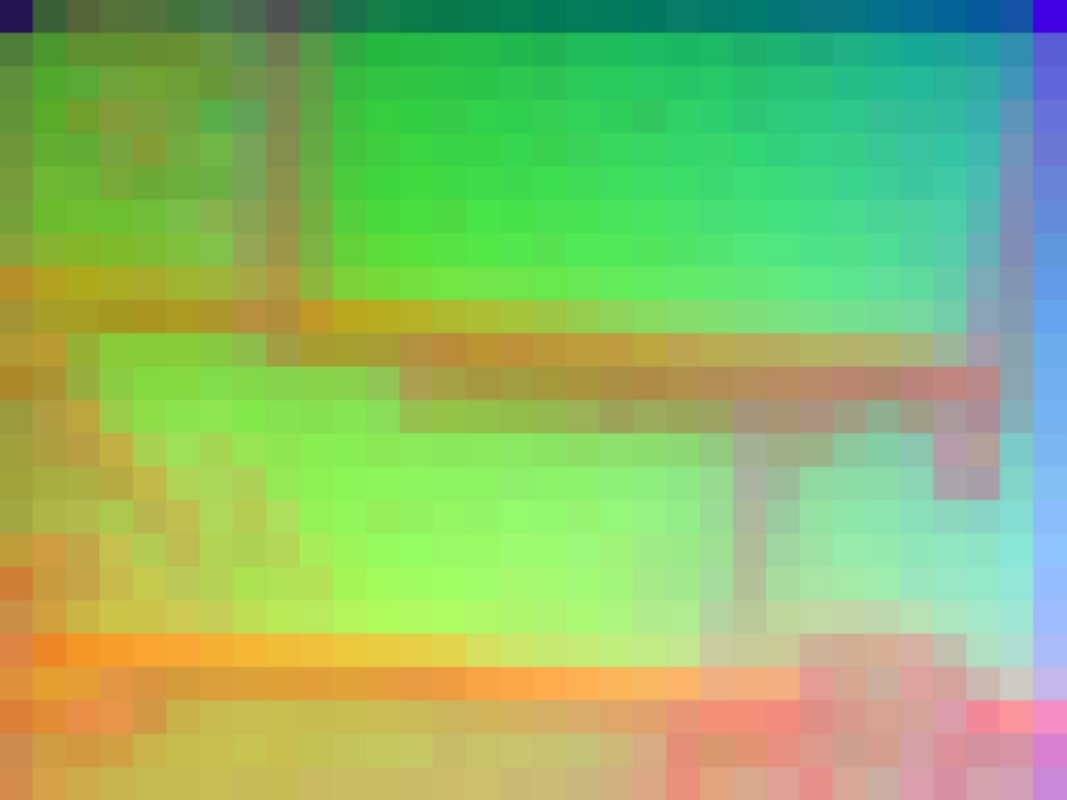} \\
        \includegraphics[width=0.166\linewidth]{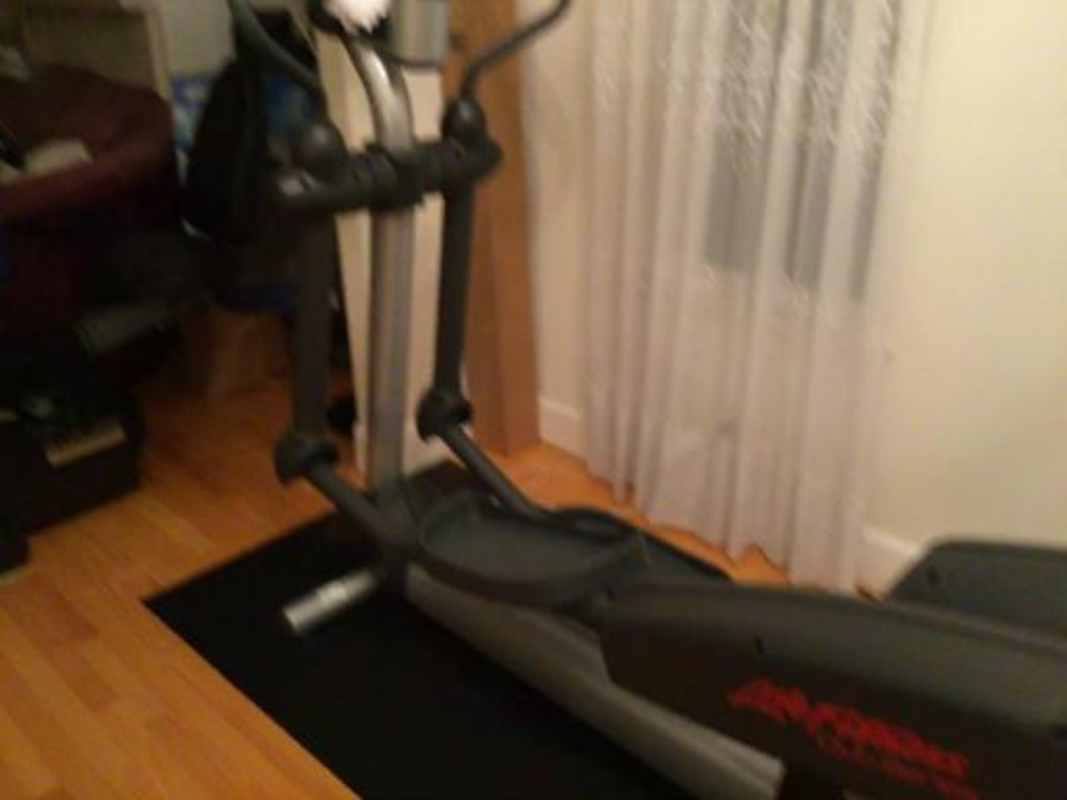} &
        \includegraphics[width=0.166\linewidth]{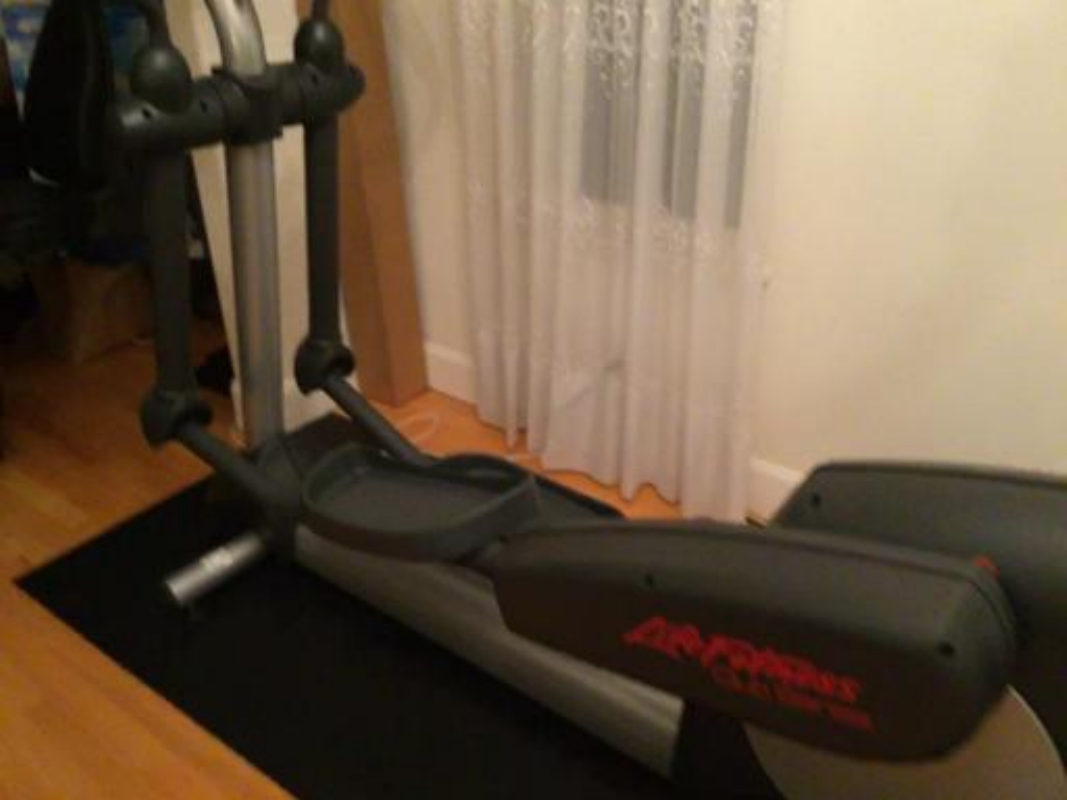} &
        \includegraphics[width=0.166\linewidth]{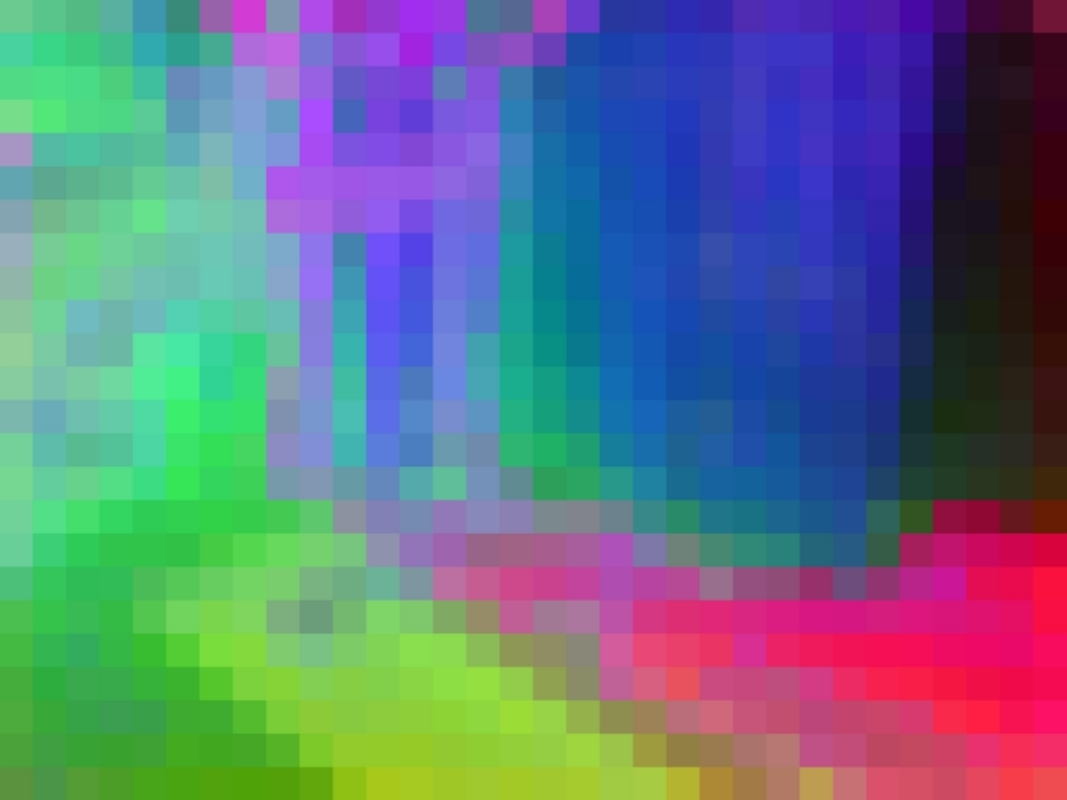} &
        \includegraphics[width=0.166\linewidth]{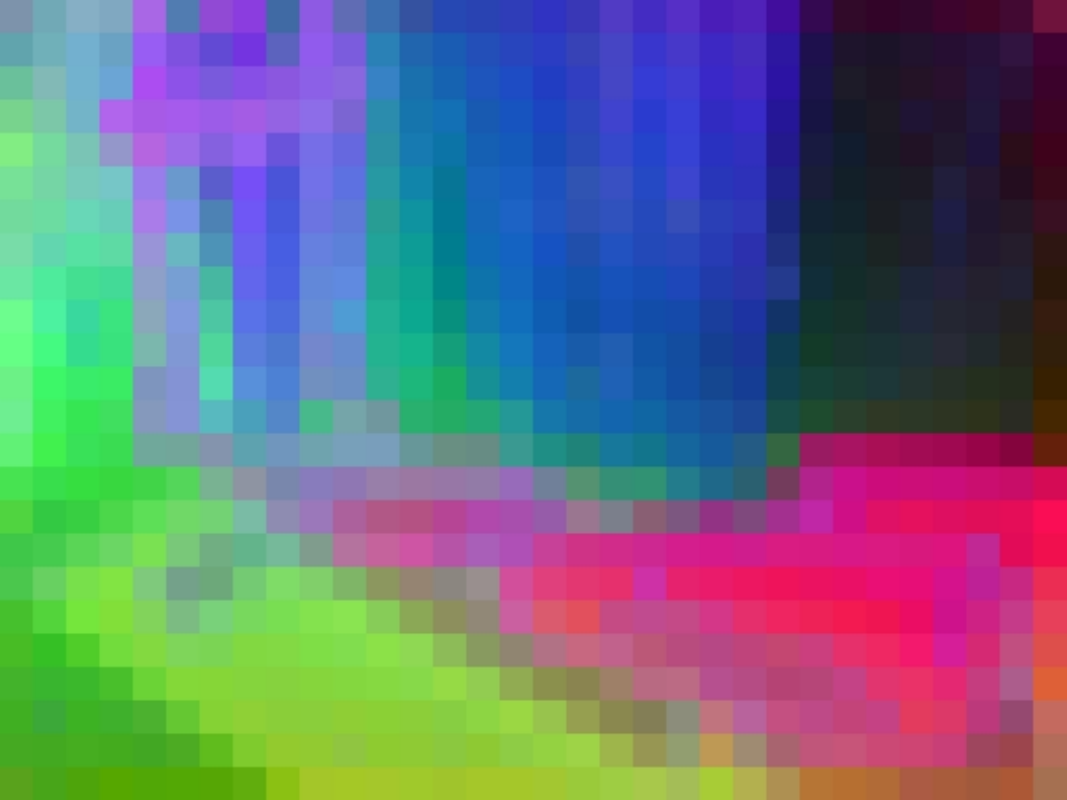} &
        \includegraphics[width=0.166\linewidth]{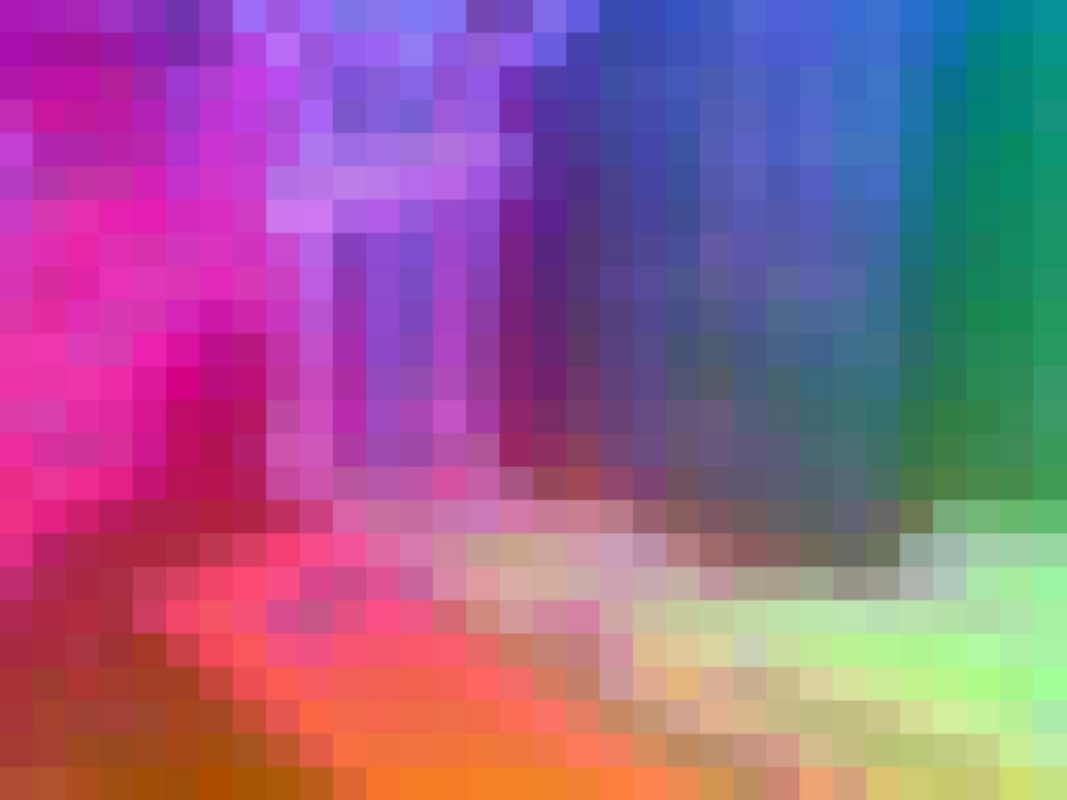} &
        \includegraphics[width=0.166\linewidth]{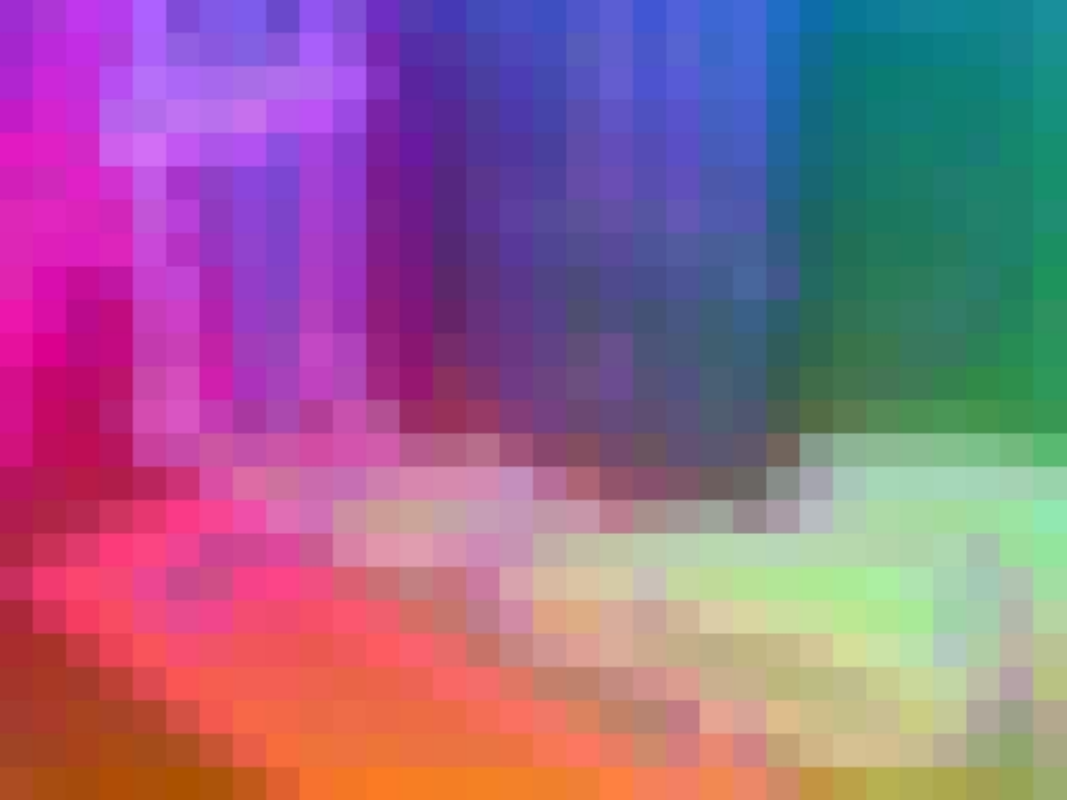} \\
        \includegraphics[width=0.166\linewidth]{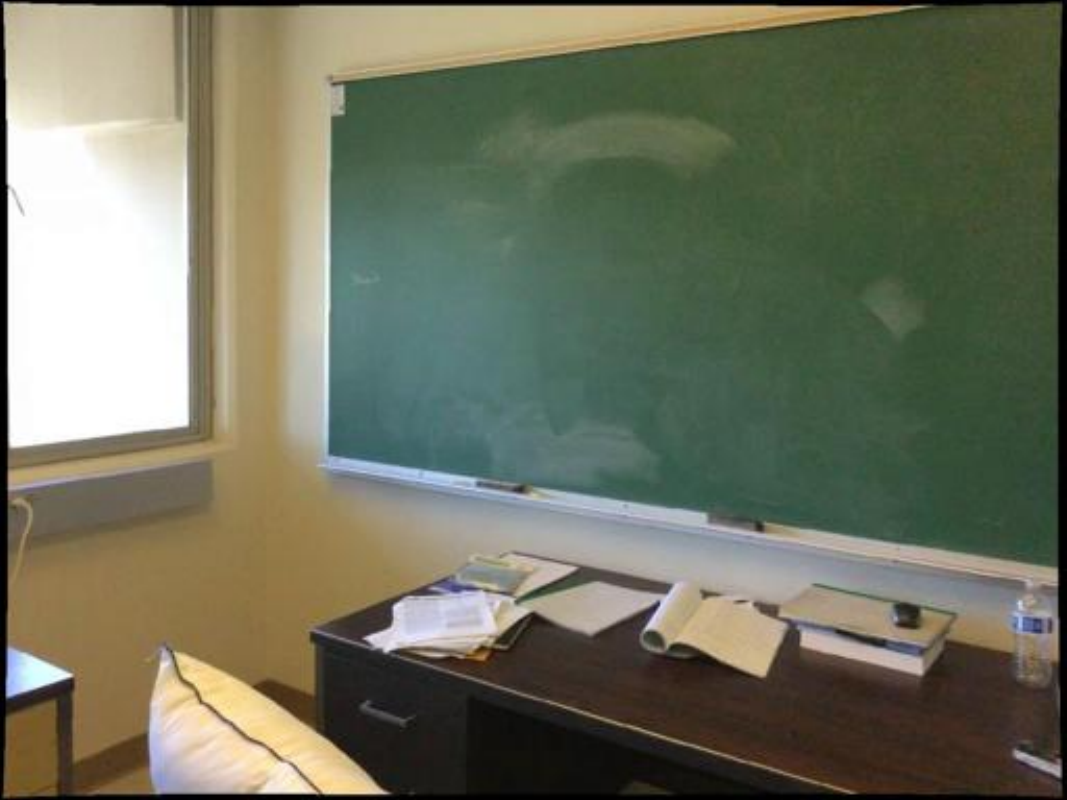} &
        \includegraphics[width=0.166\linewidth]{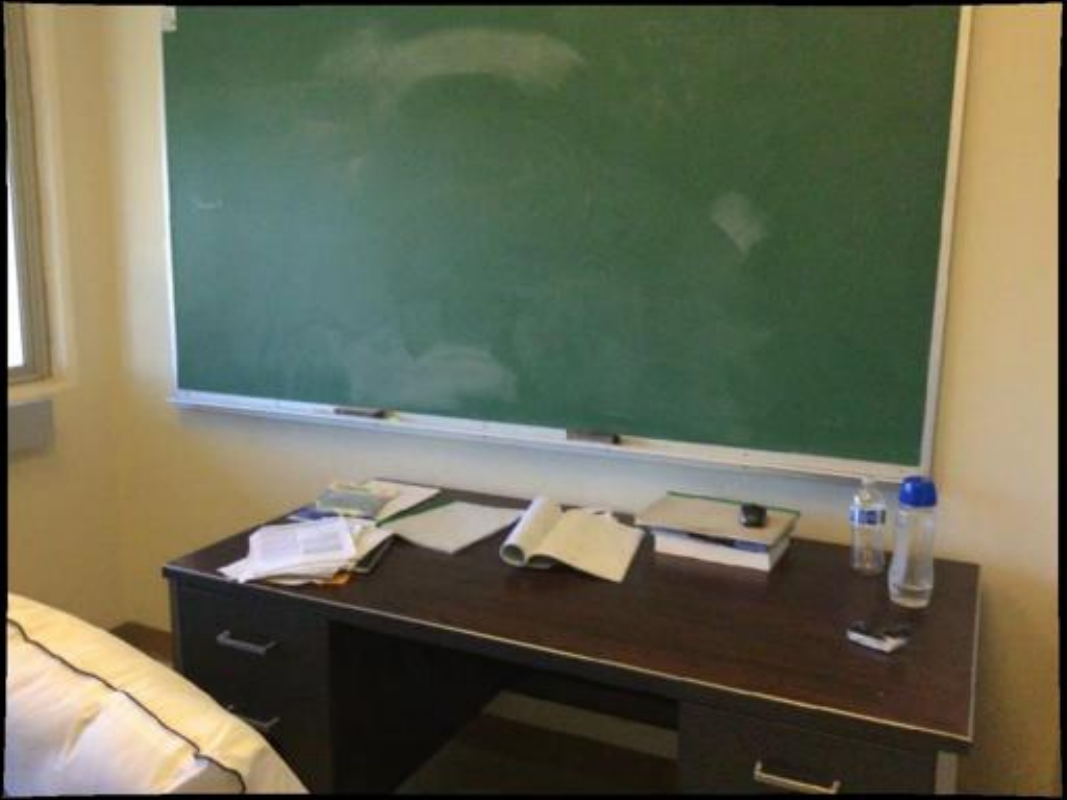} &
        \includegraphics[width=0.166\linewidth]{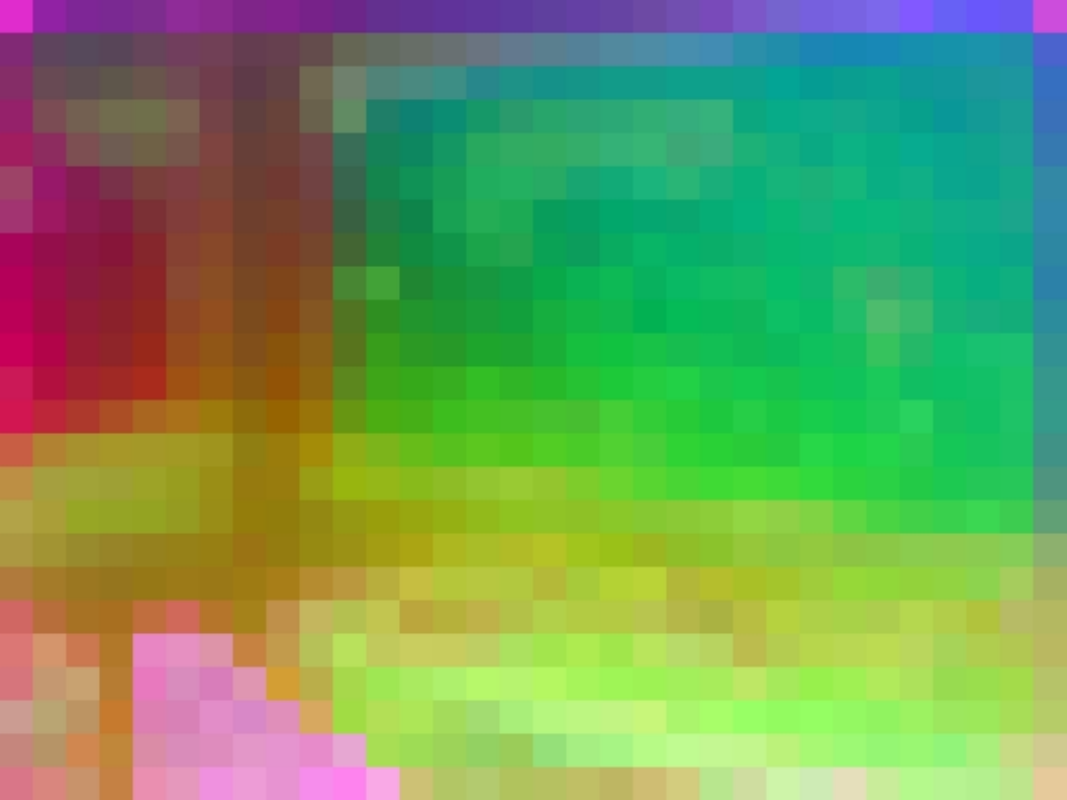} &
        \includegraphics[width=0.166\linewidth]{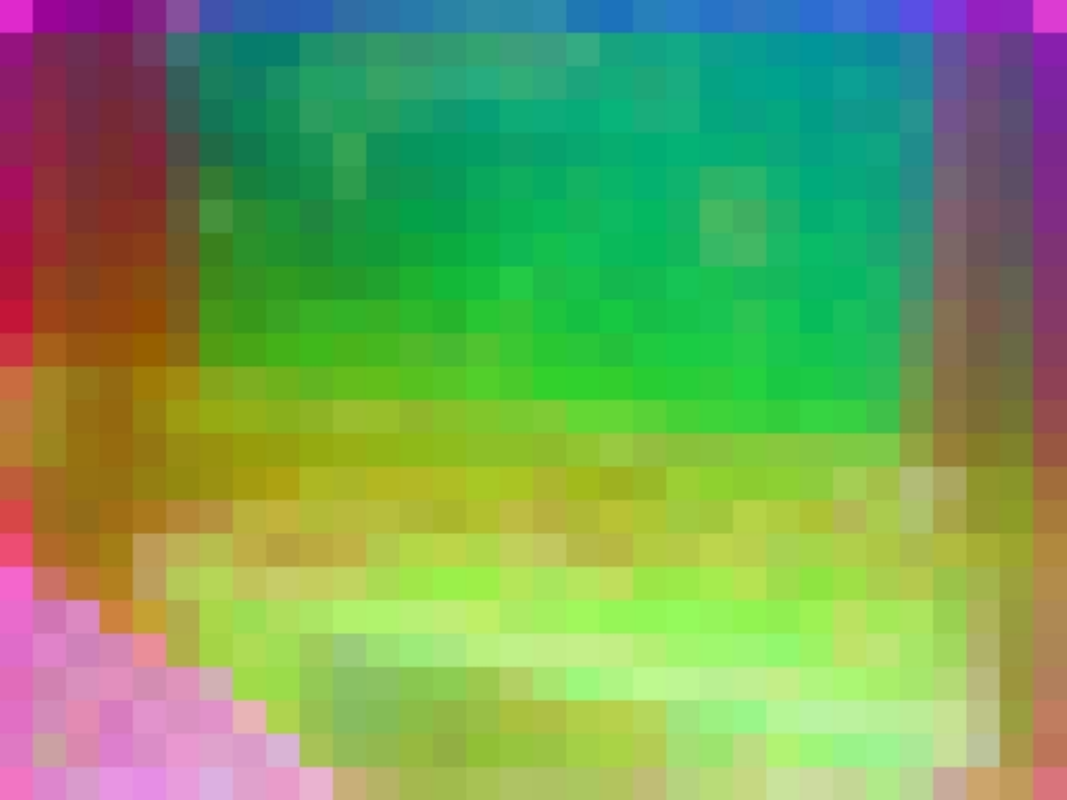} &
        \includegraphics[width=0.166\linewidth]{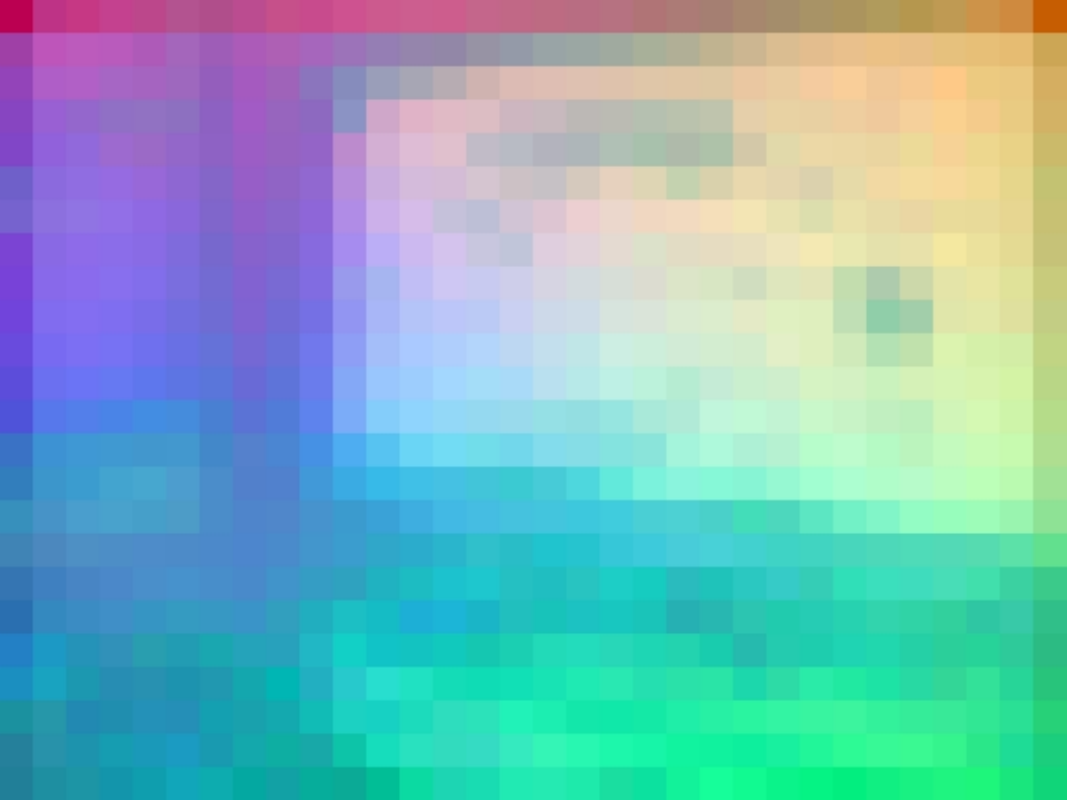} &
        \includegraphics[width=0.166\linewidth]{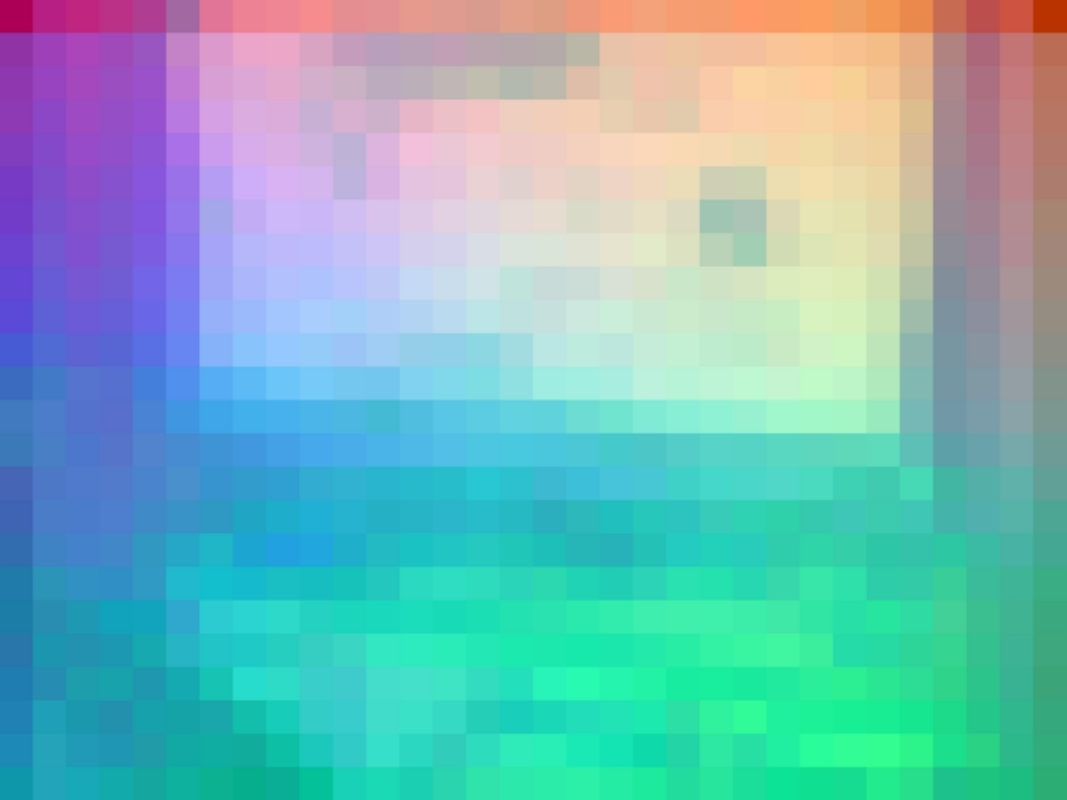} \\
        \includegraphics[width=0.166\linewidth]{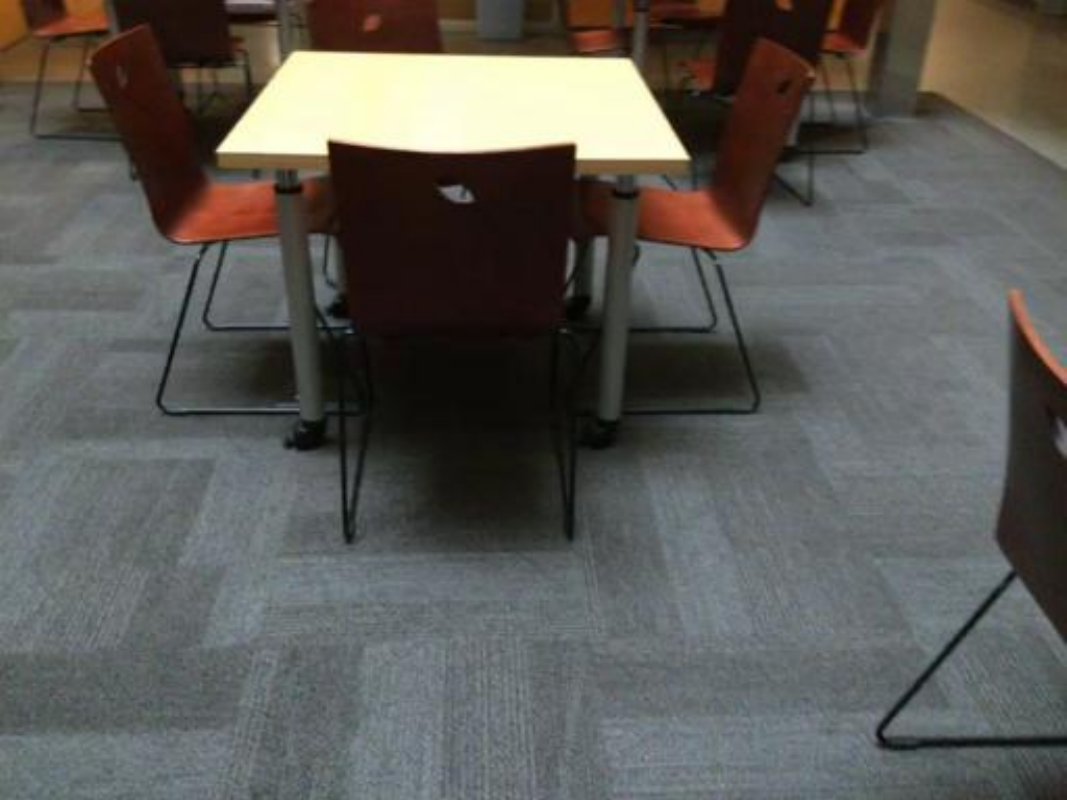} &
        \includegraphics[width=0.166\linewidth]{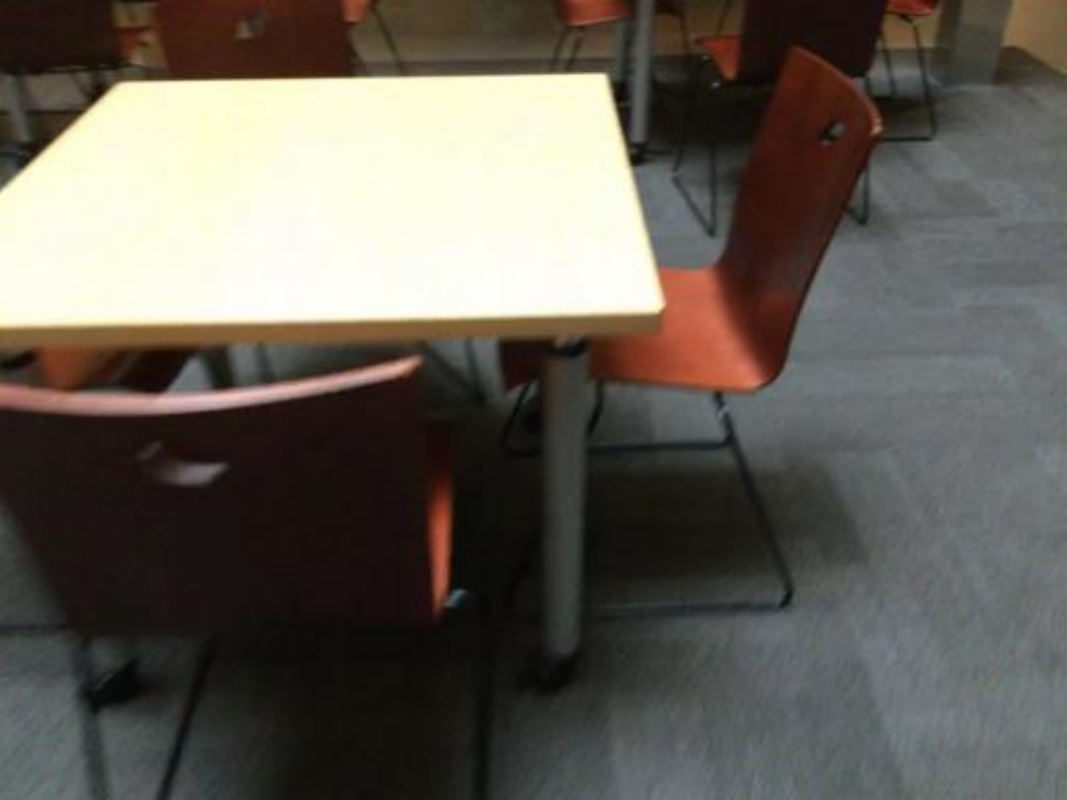} &
        \includegraphics[width=0.166\linewidth]{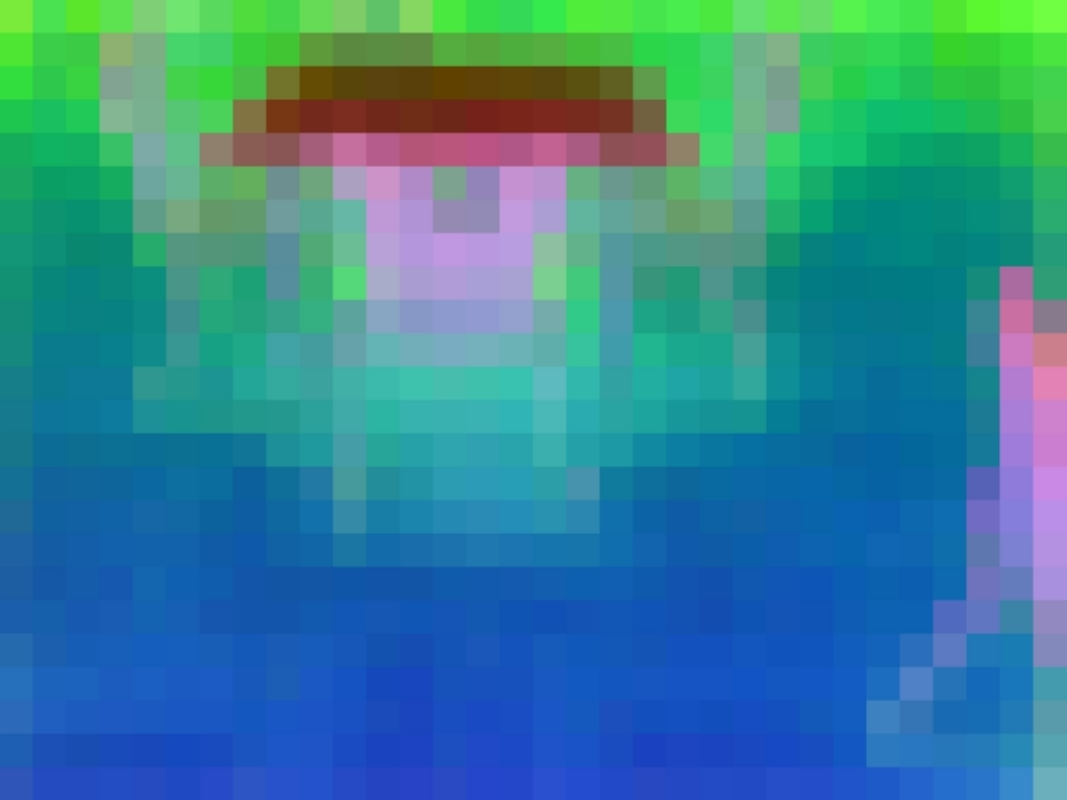} &
        \includegraphics[width=0.166\linewidth]{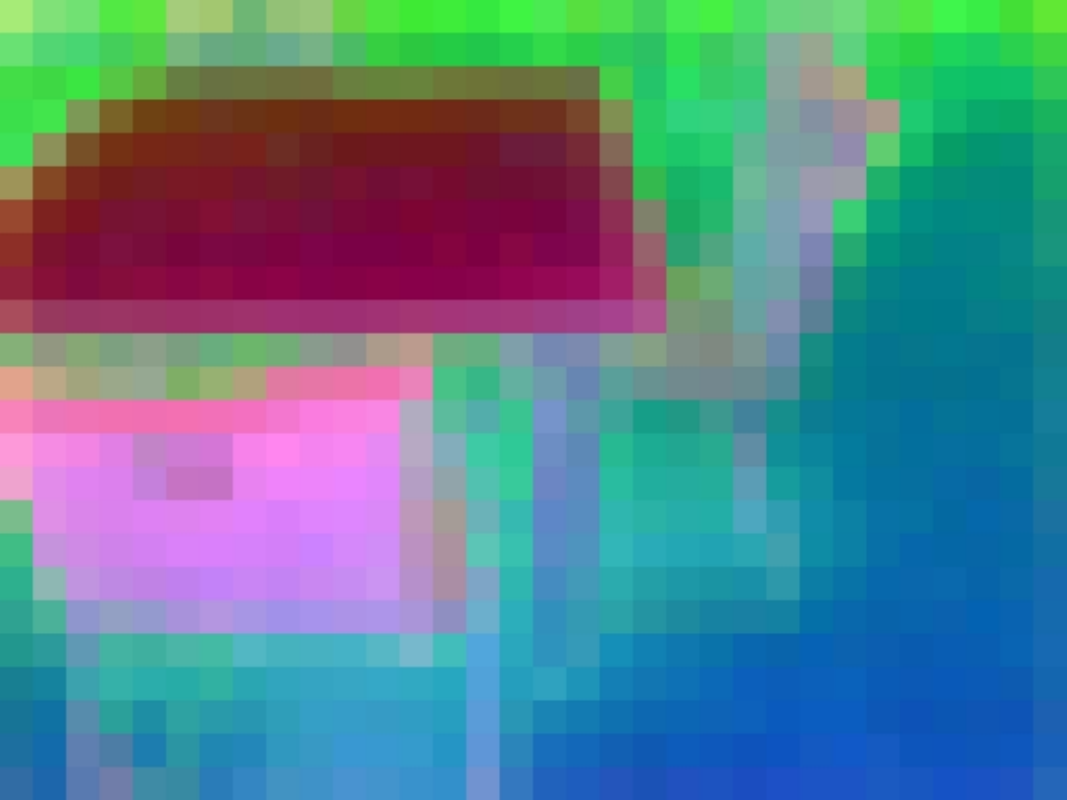} &
        \includegraphics[width=0.166\linewidth]{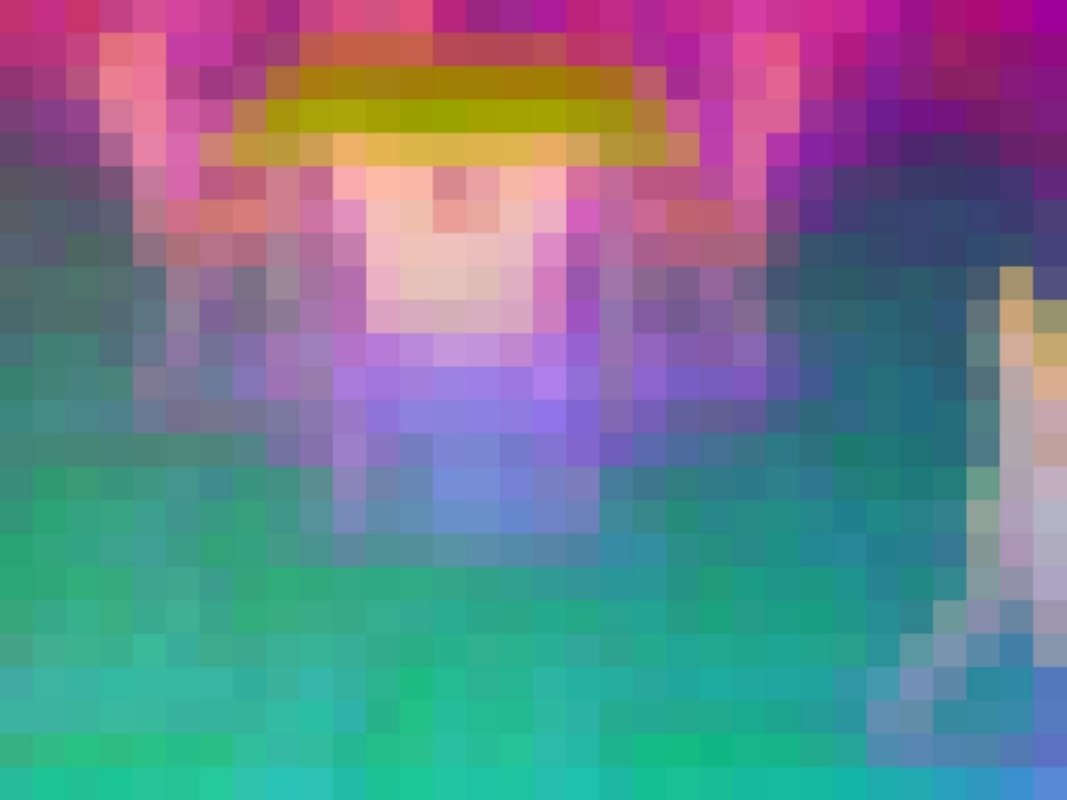} &
        \includegraphics[width=0.166\linewidth]{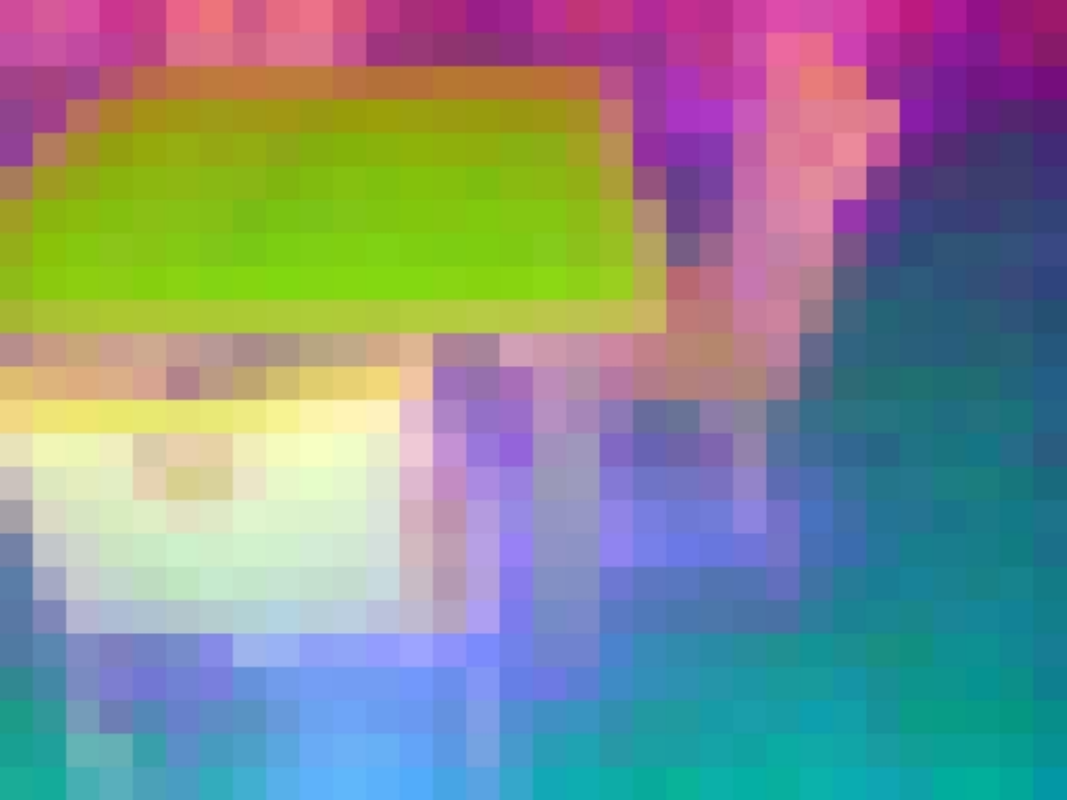} \\
        \includegraphics[width=0.166\linewidth]{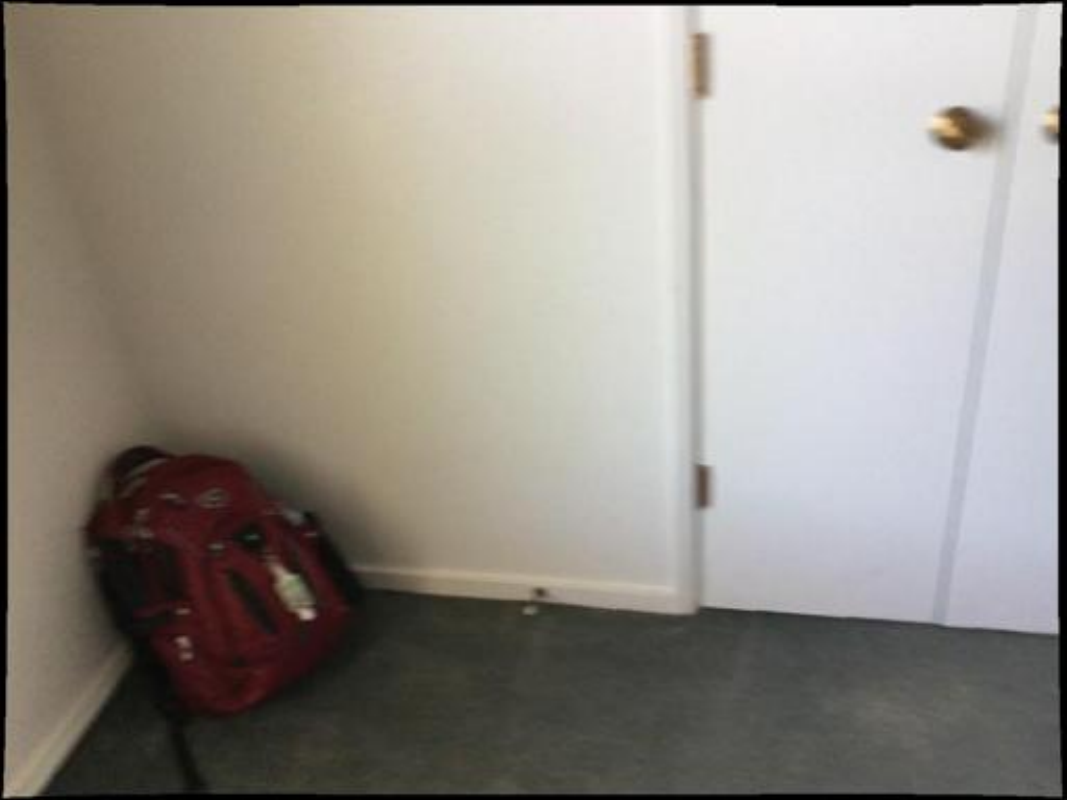} &
        \includegraphics[width=0.166\linewidth]{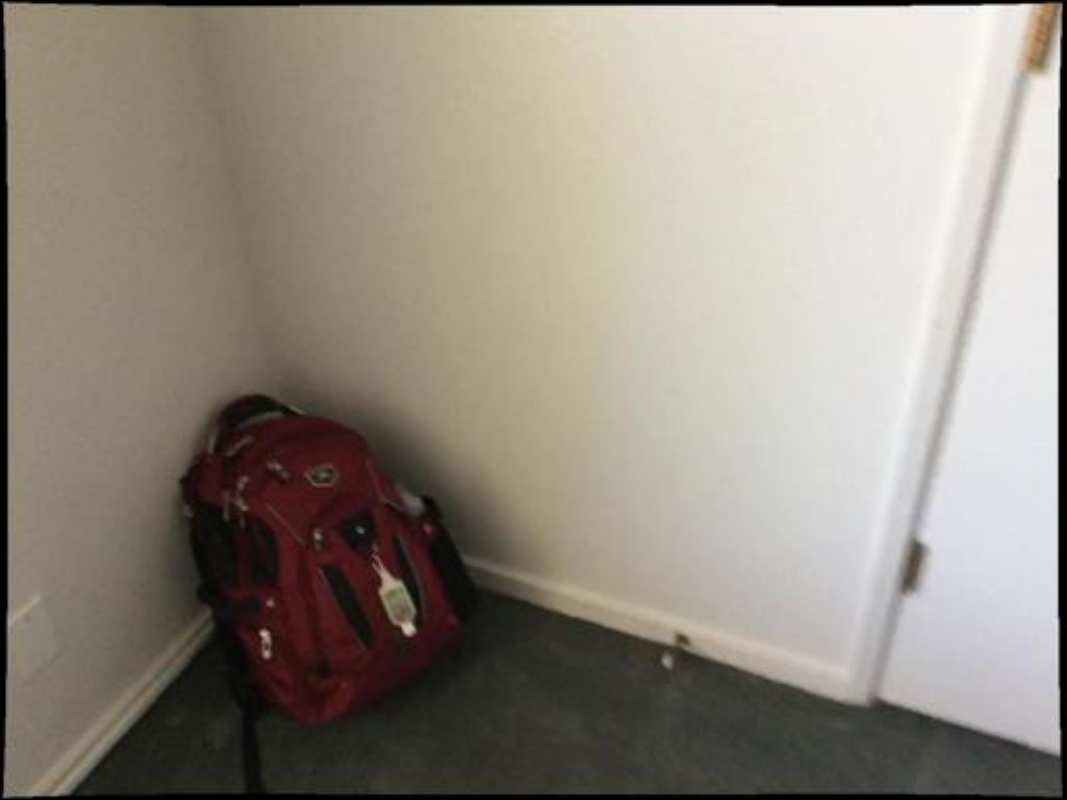} &
        \includegraphics[width=0.166\linewidth]{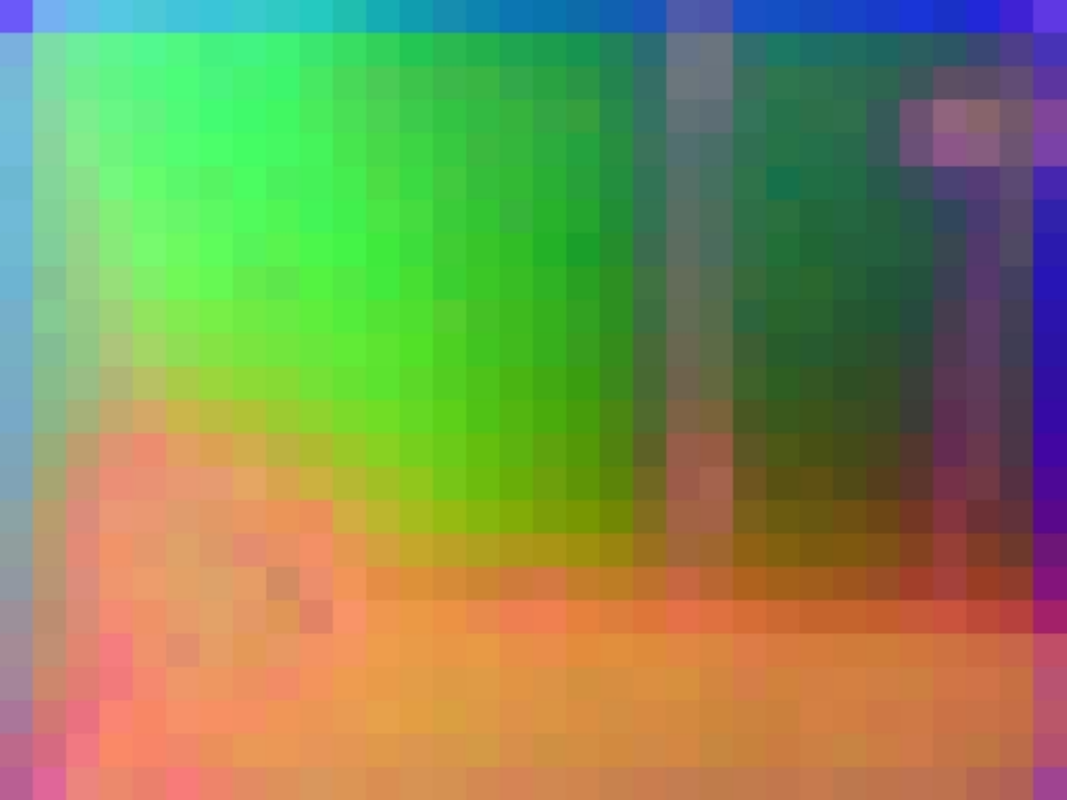} &
        \includegraphics[width=0.166\linewidth]{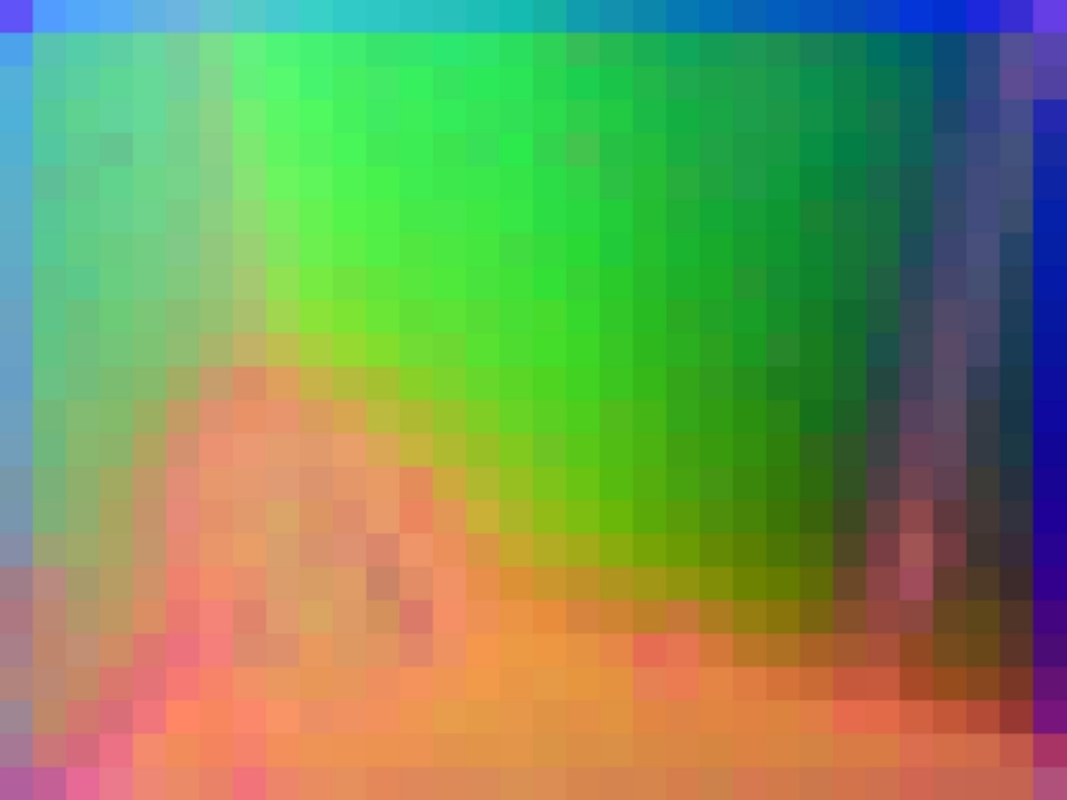} &
        \includegraphics[width=0.166\linewidth]{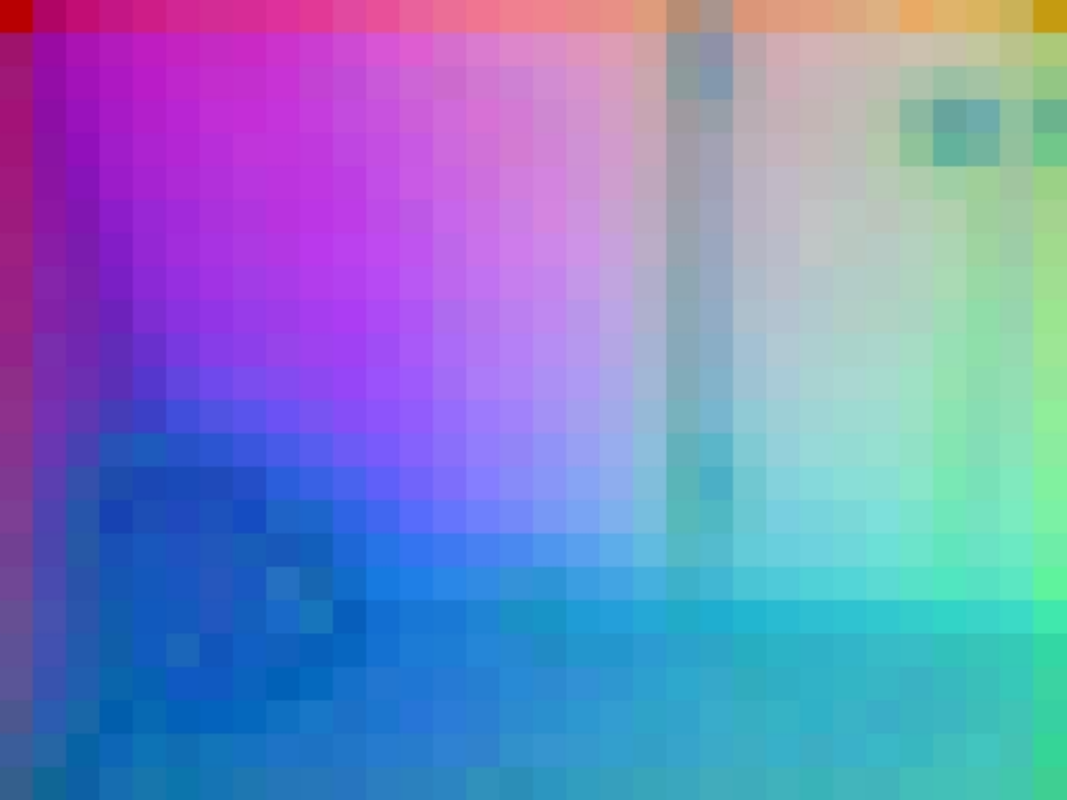} &
        \includegraphics[width=0.166\linewidth]{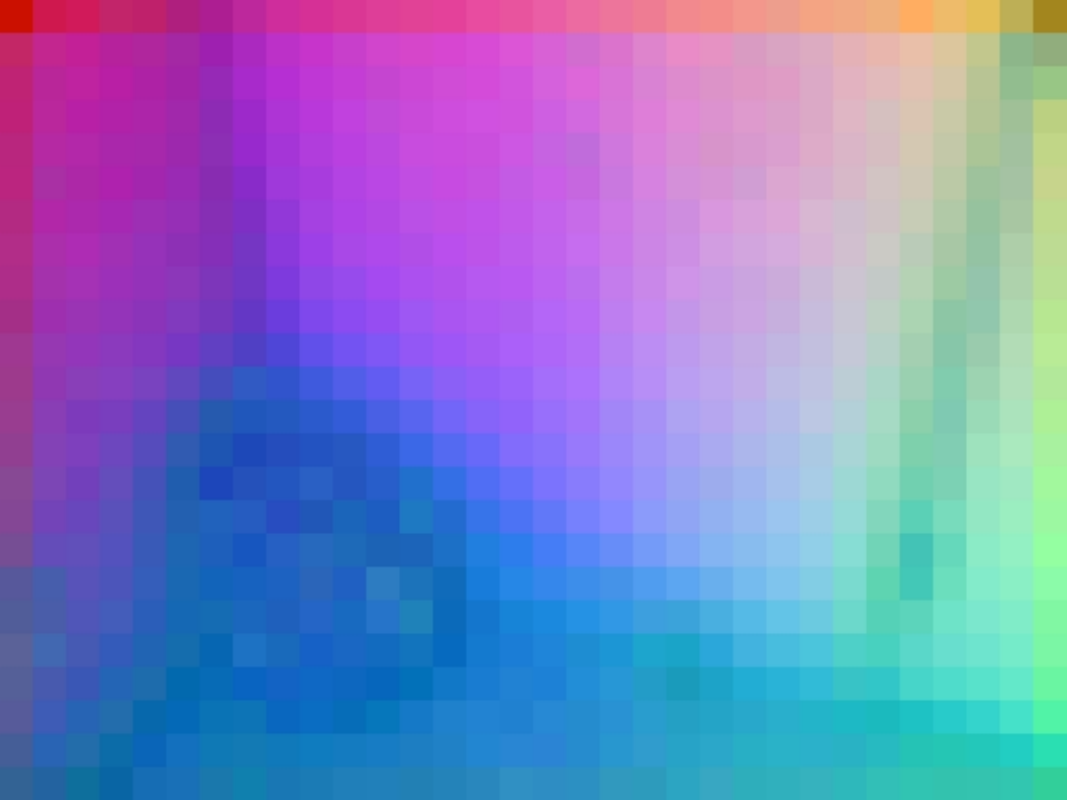} \\
    \end{tabular}
    \caption{\textbf{PCA visualization of encoder features for CroCo and~\ourmethod.} For each image pair, the three principal components of the encoder output tokens are mapped to RGB channels.}
    \label{fig:pca_features_1}
\end{figure*}

\begin{figure*}[t]
    \centering
    \setlength{\tabcolsep}{0pt}
    \begin{tabular}{@{}cccccc@{}}
        \makebox[0.166\linewidth][c]{\small Image 1} &
        \makebox[0.166\linewidth][c]{\small Image 2} &
        \makebox[0.166\linewidth][c]{\small CroCo PCA 1} &
        \makebox[0.166\linewidth][c]{\small CroCo PCA 2} &
        \makebox[0.166\linewidth][c]{\small \ourmethod PCA 1} &
        \makebox[0.166\linewidth][c]{\small \ourmethod PCA 2} \\
        \includegraphics[width=0.166\linewidth]{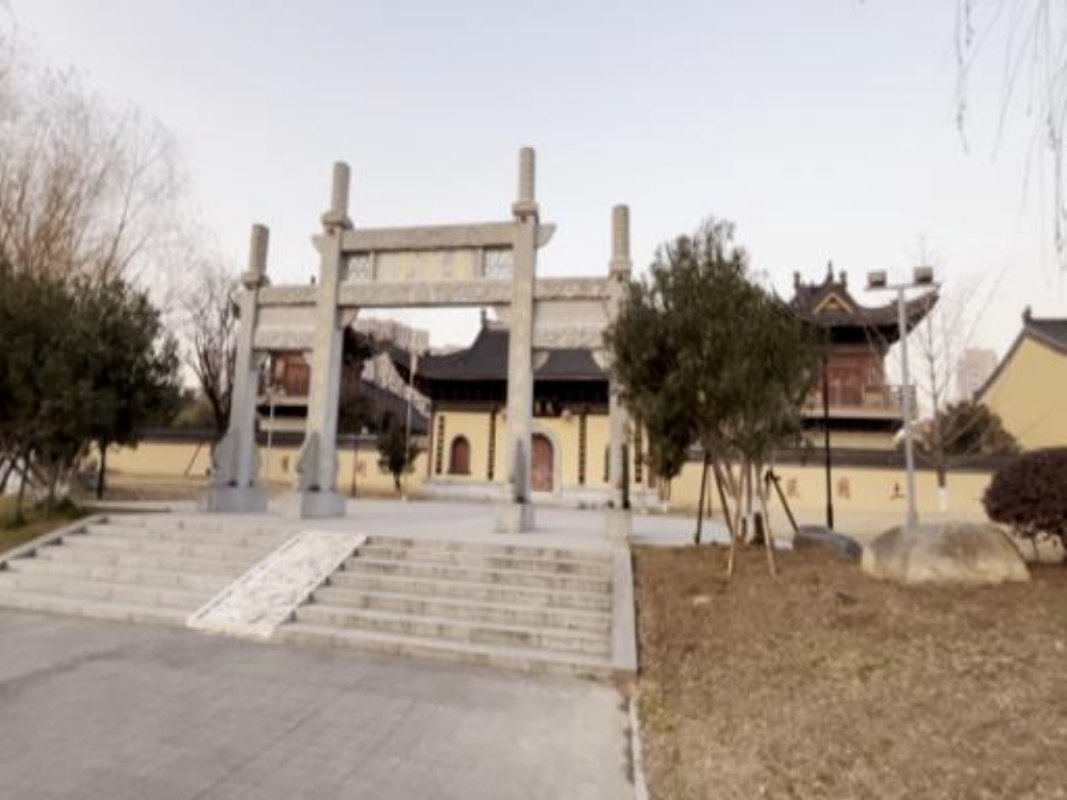} &
        \includegraphics[width=0.166\linewidth]{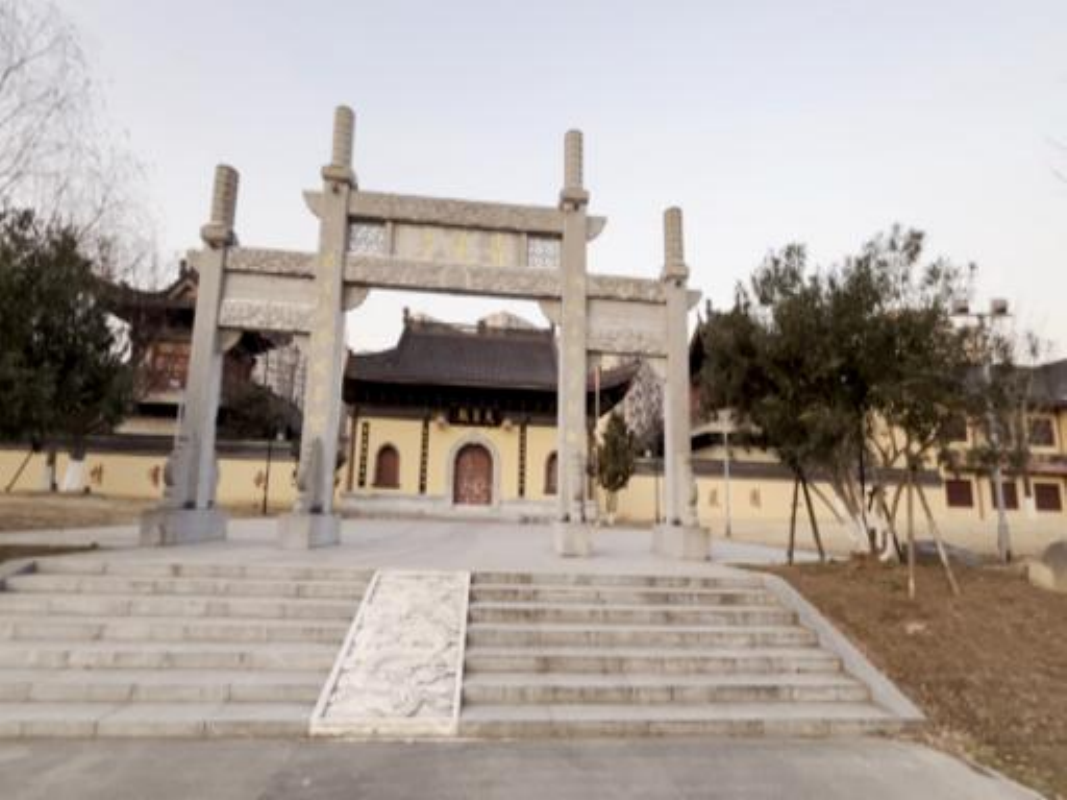} &
        \includegraphics[width=0.166\linewidth]{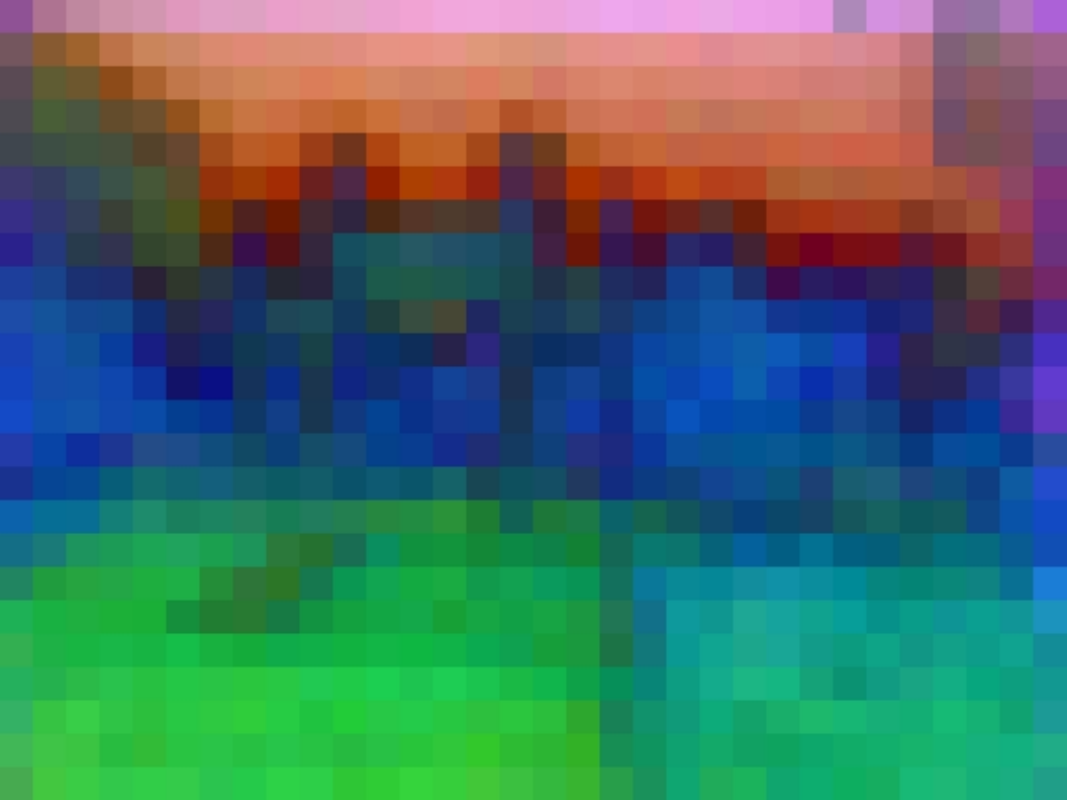} &
        \includegraphics[width=0.166\linewidth]{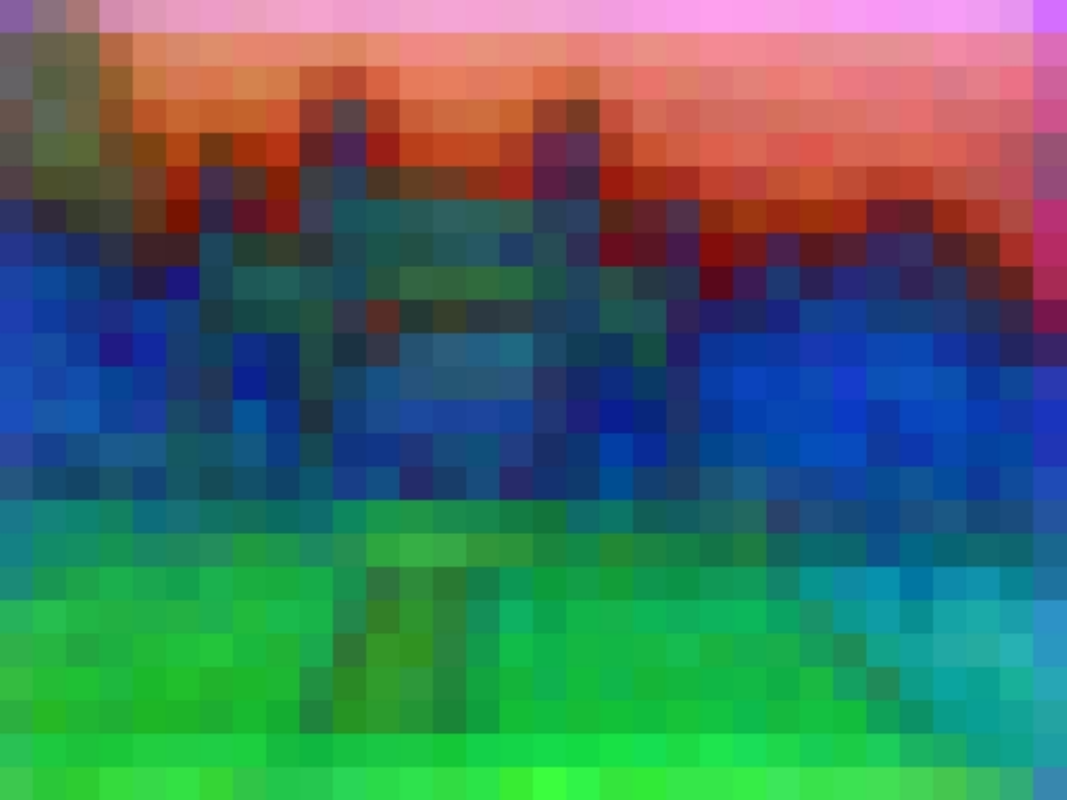} &
        \includegraphics[width=0.166\linewidth]{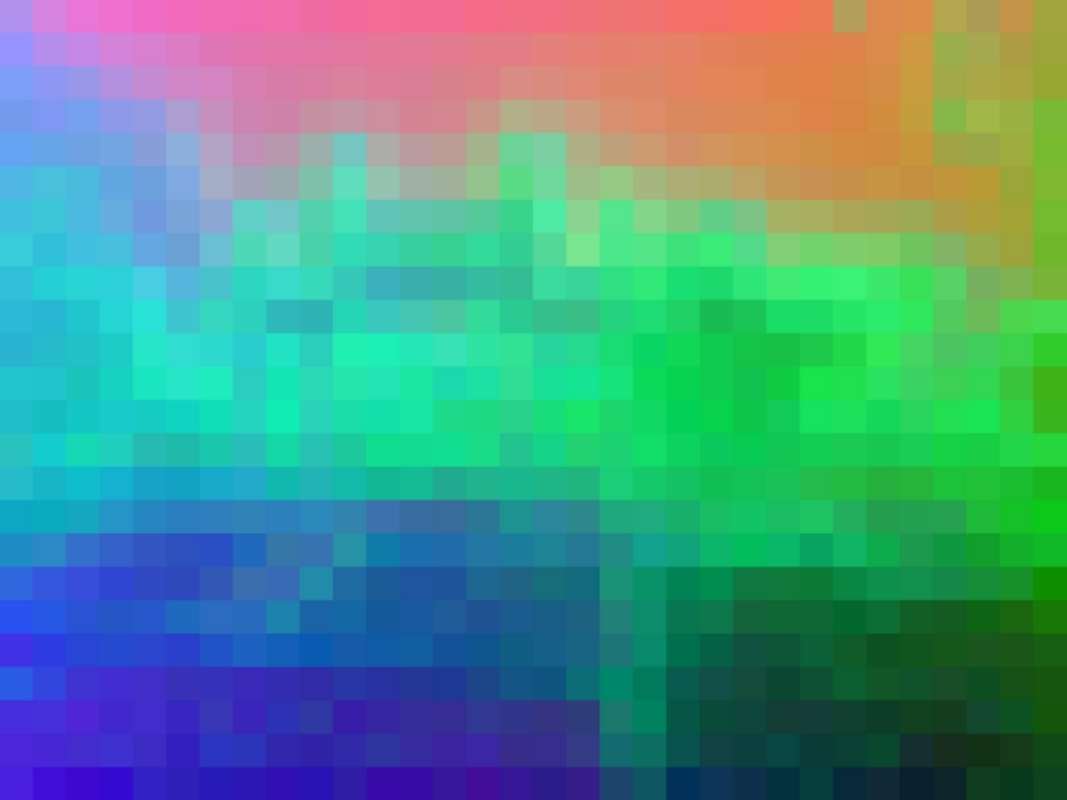} &
        \includegraphics[width=0.166\linewidth]{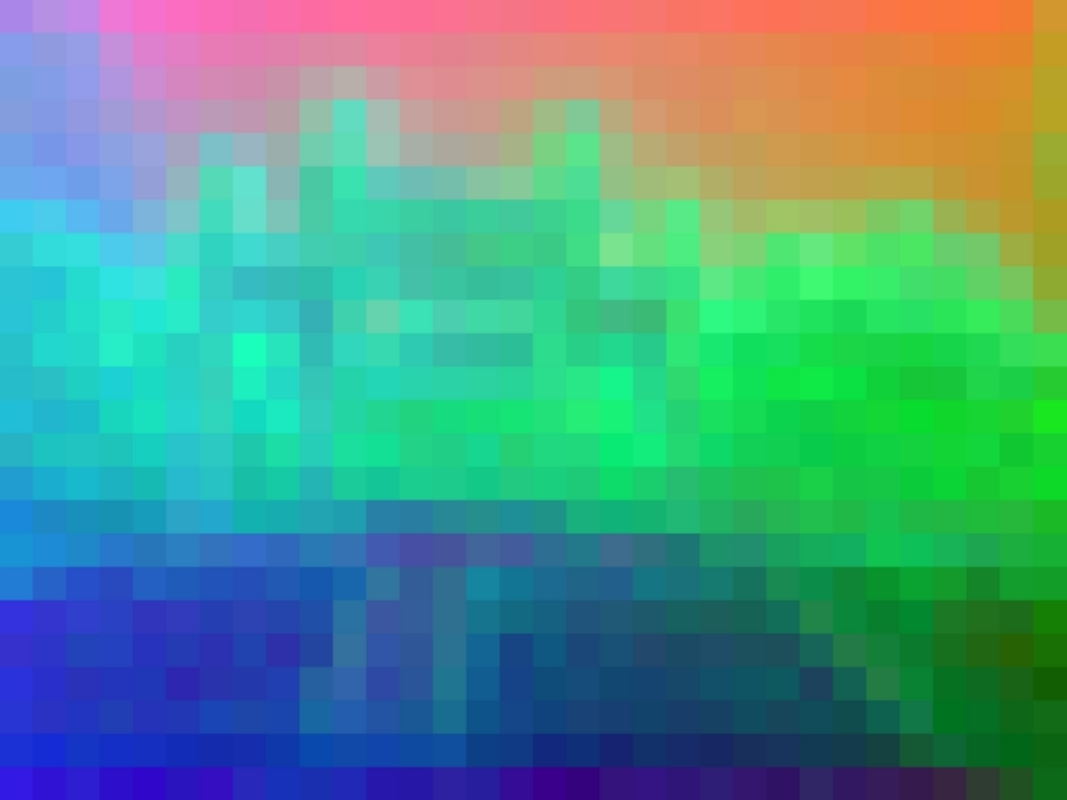} \\
        \includegraphics[width=0.166\linewidth]{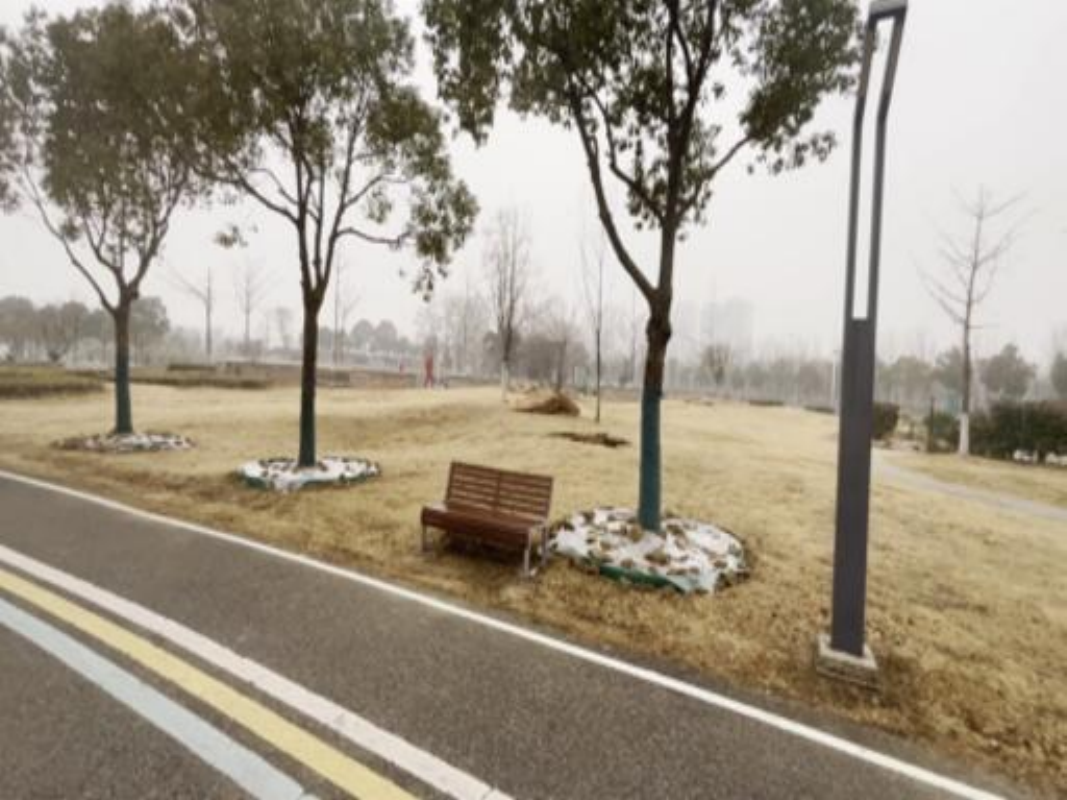} &
        \includegraphics[width=0.166\linewidth]{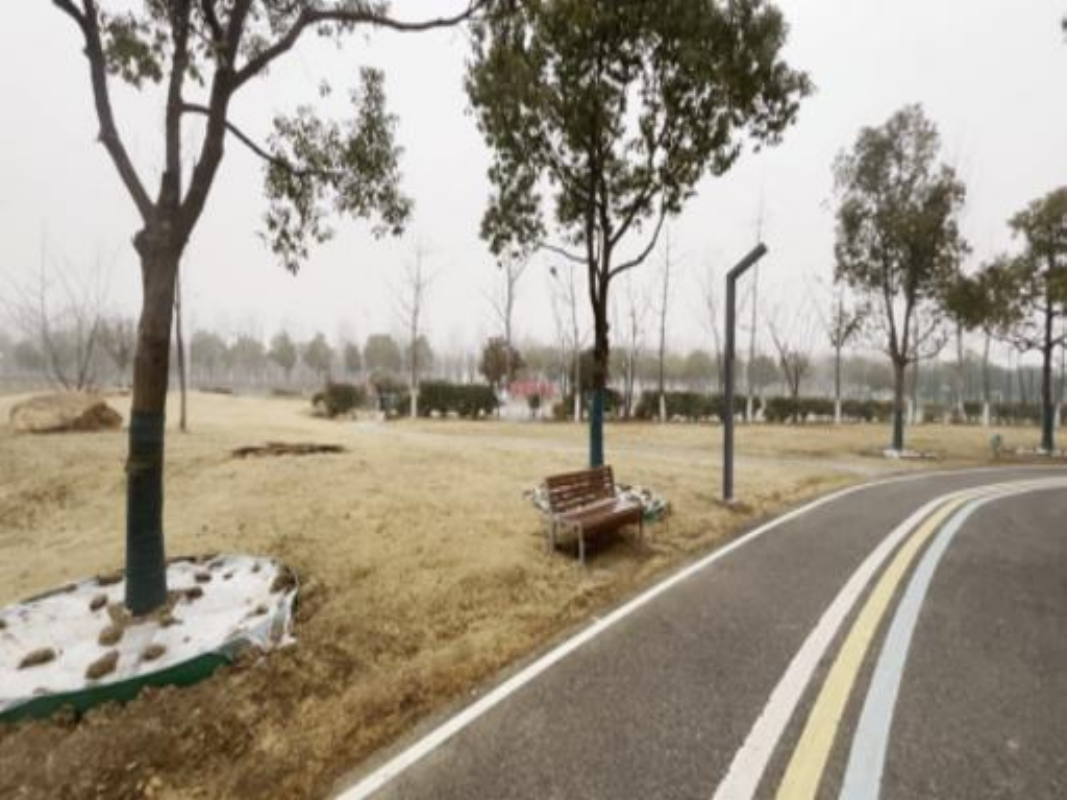} &
        \includegraphics[width=0.166\linewidth]{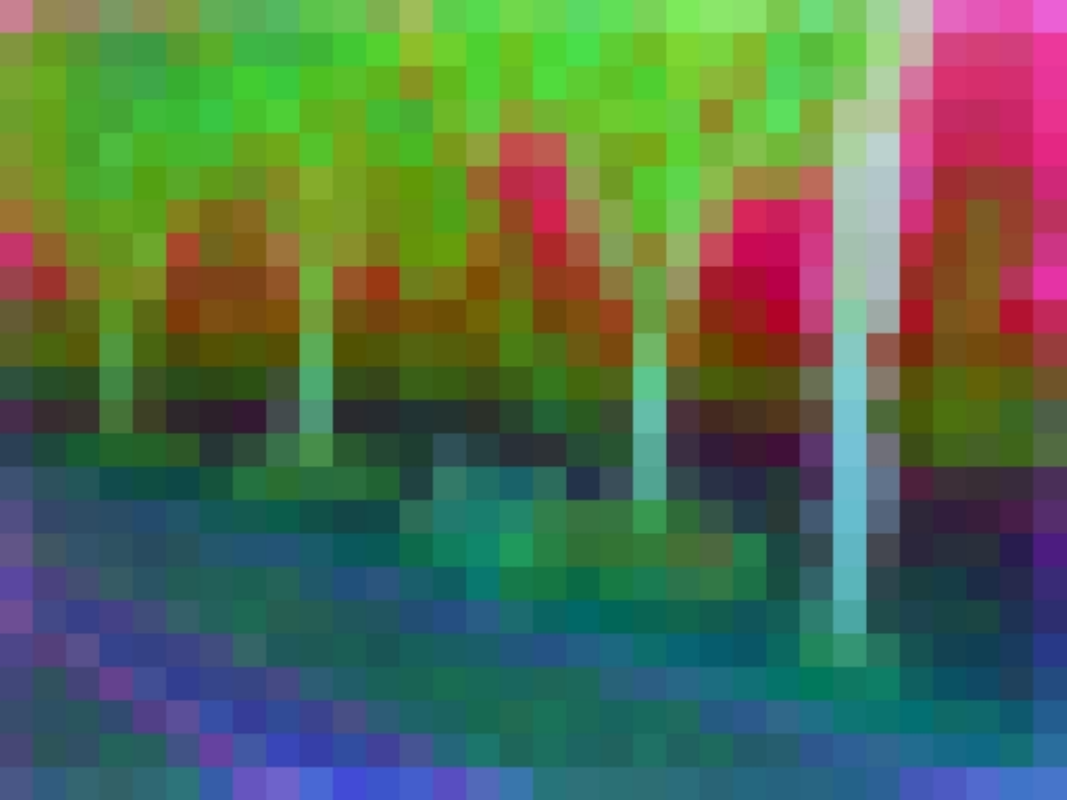} &
        \includegraphics[width=0.166\linewidth]{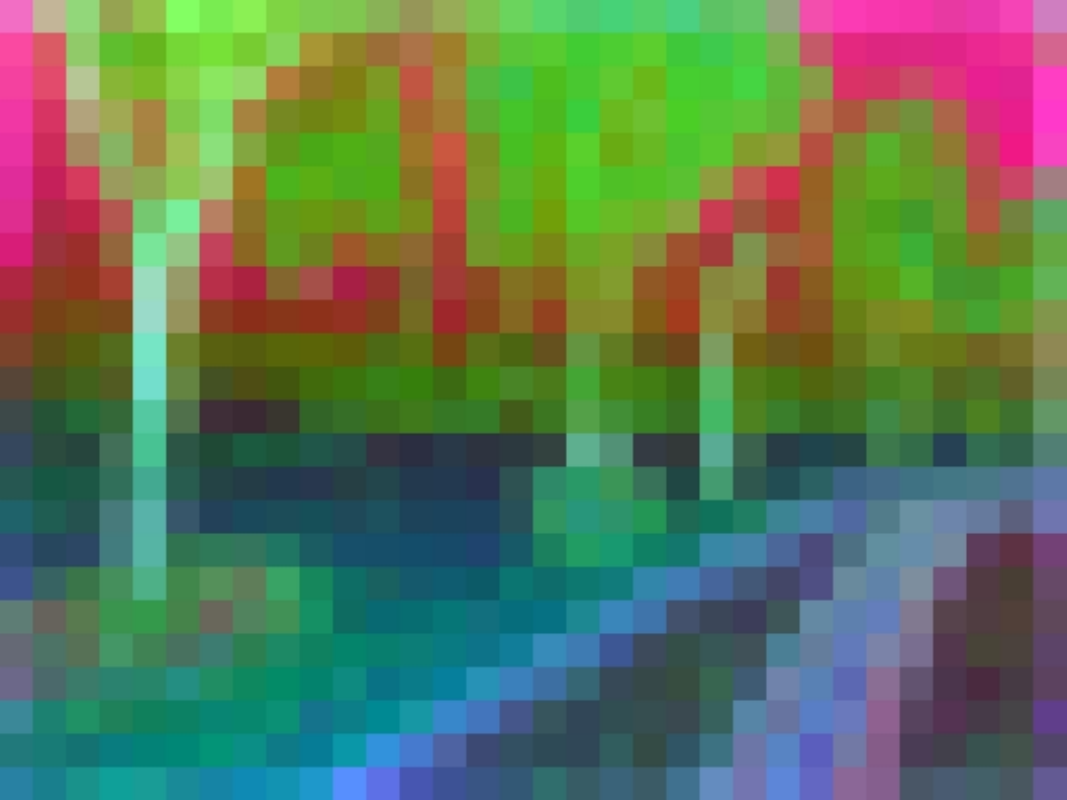} &
        \includegraphics[width=0.166\linewidth]{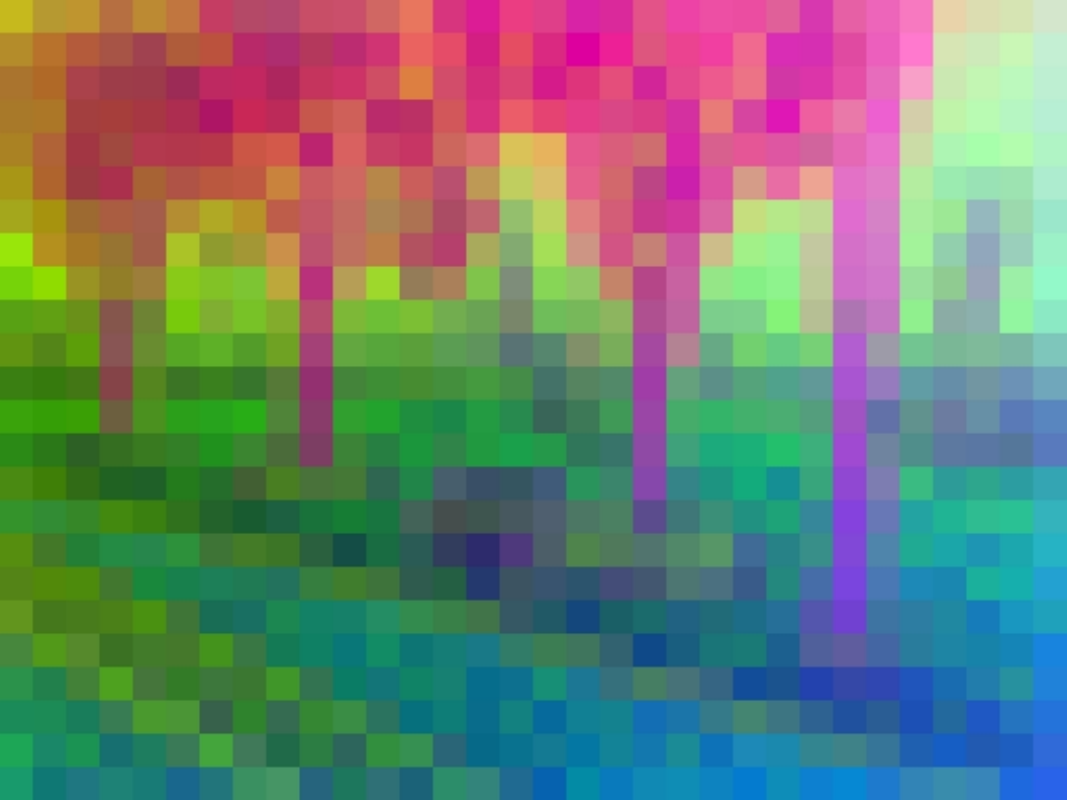} &
        \includegraphics[width=0.166\linewidth]{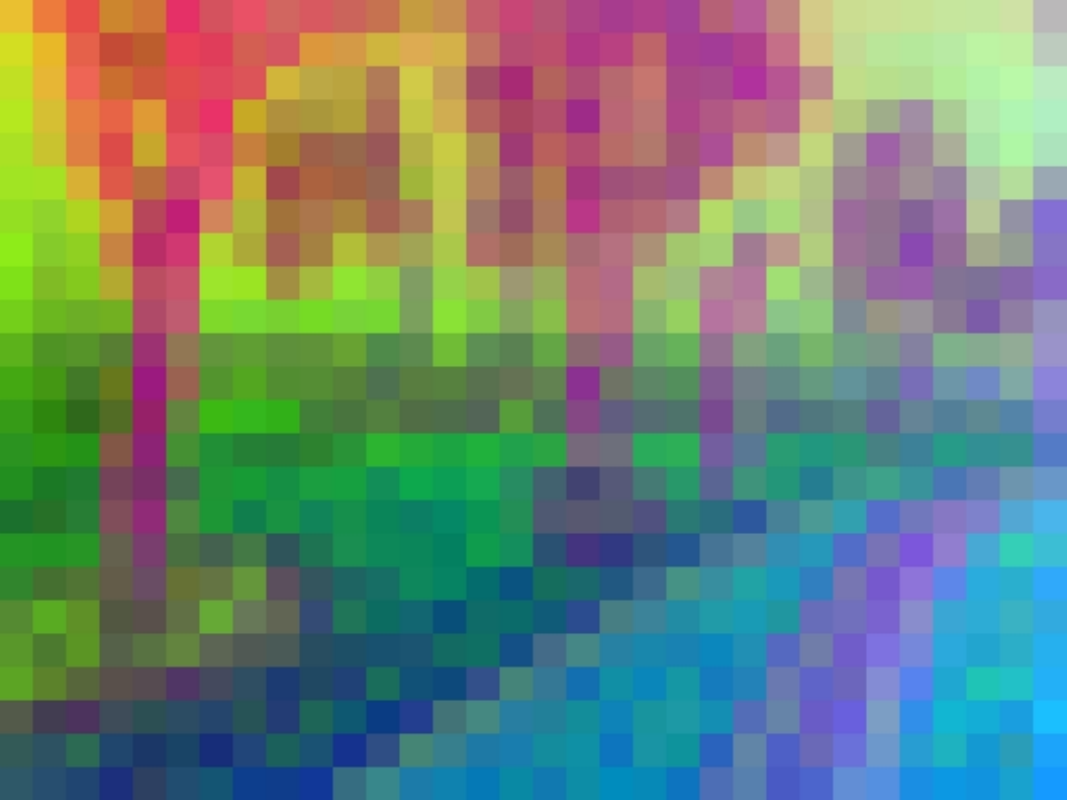} \\
        \includegraphics[width=0.166\linewidth]{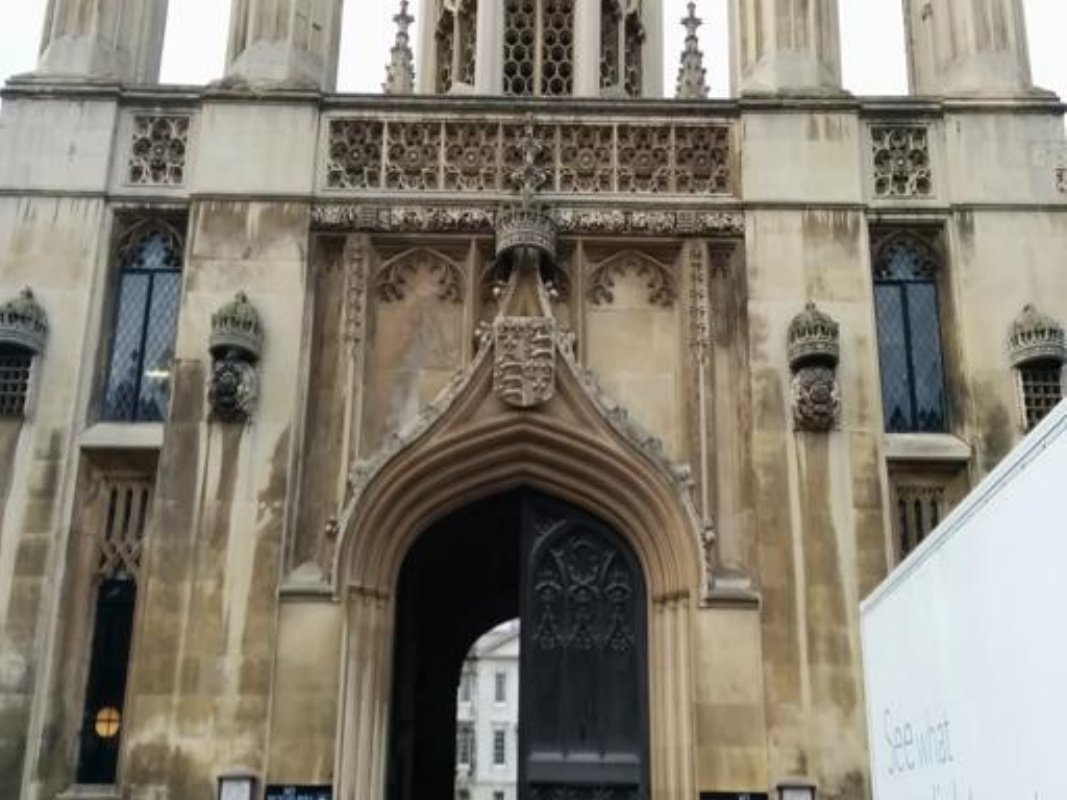} &
        \includegraphics[width=0.166\linewidth]{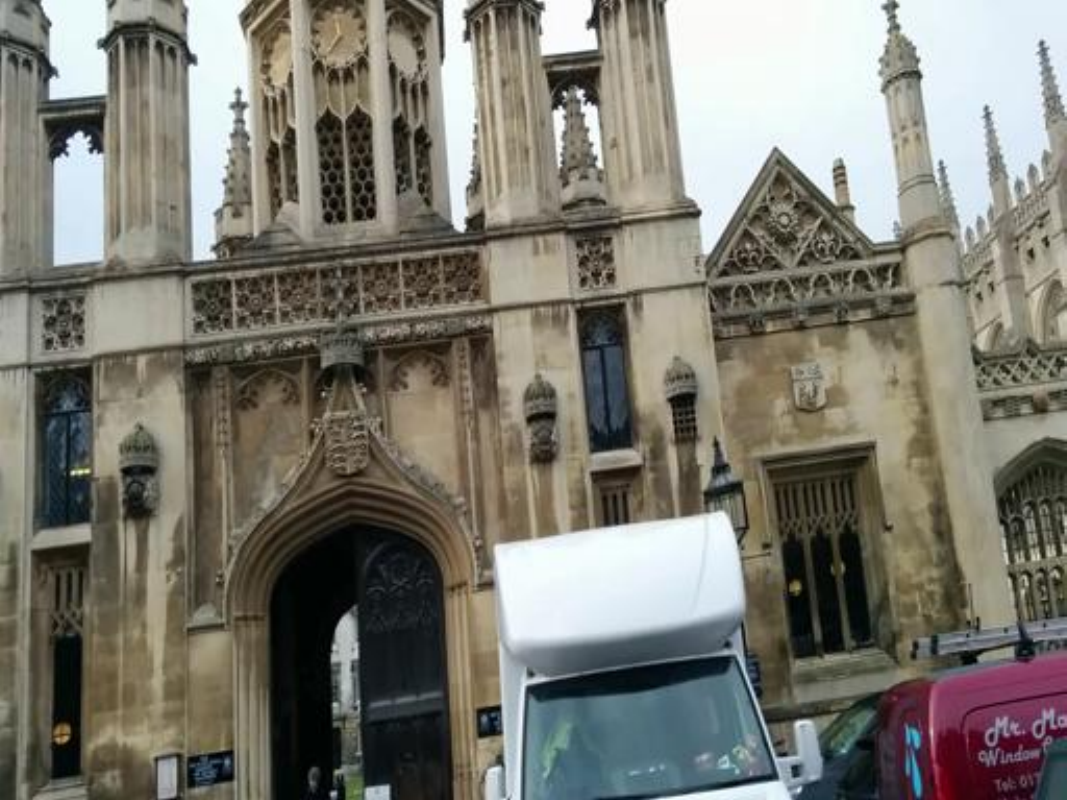} &
        \includegraphics[width=0.166\linewidth]{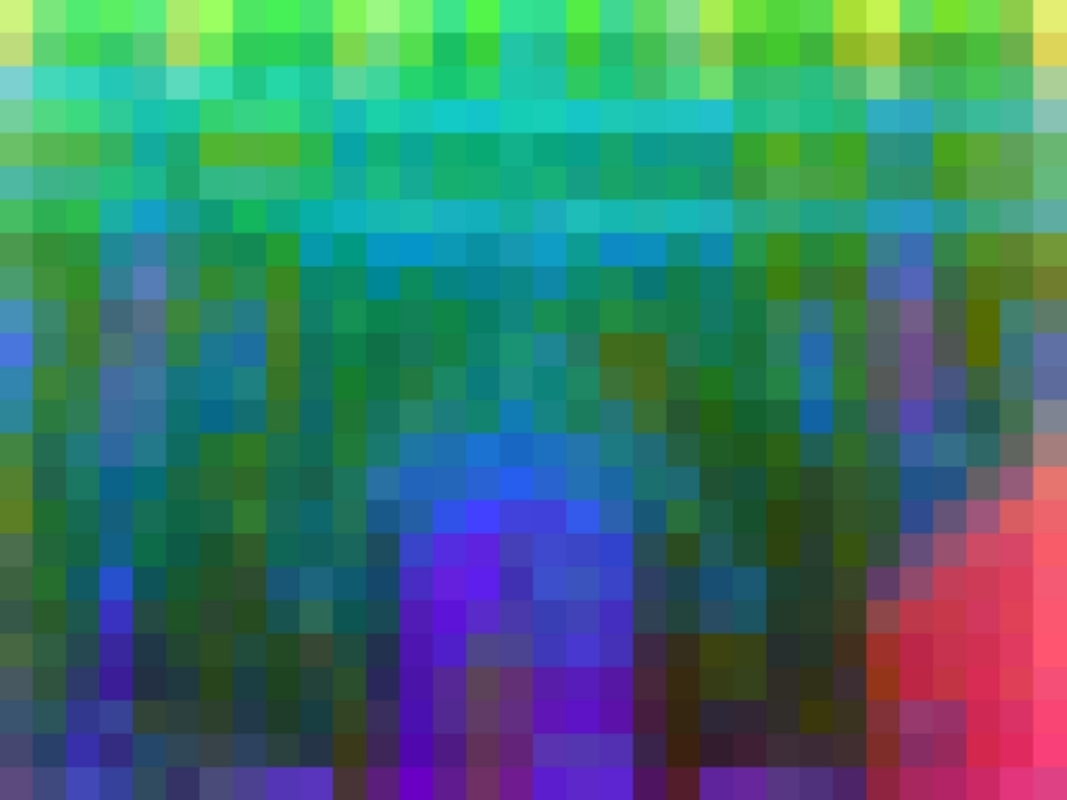} &
        \includegraphics[width=0.166\linewidth]{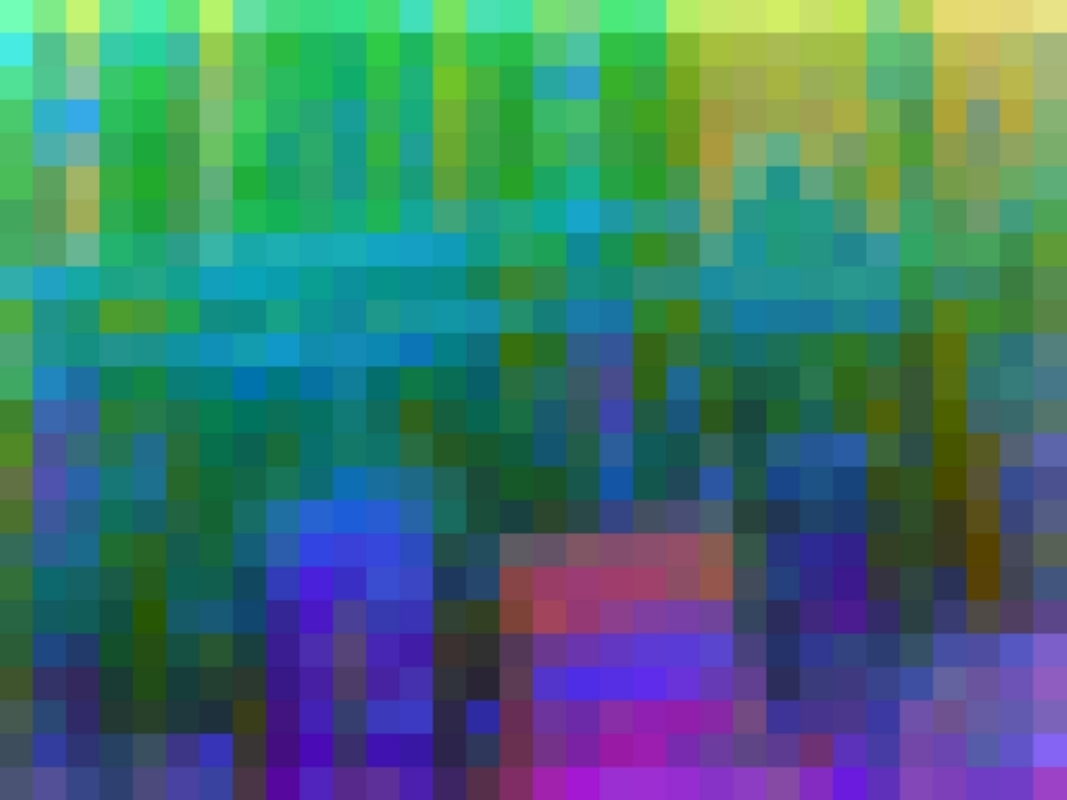} &
        \includegraphics[width=0.166\linewidth]{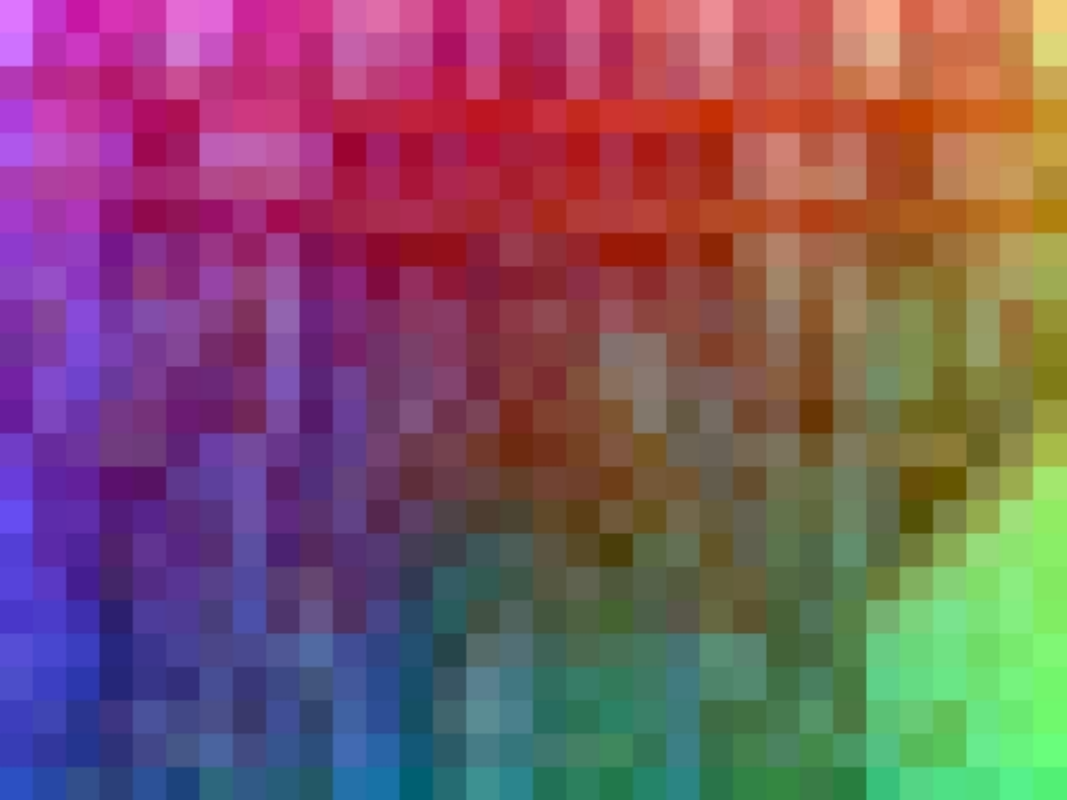} &
        \includegraphics[width=0.166\linewidth]{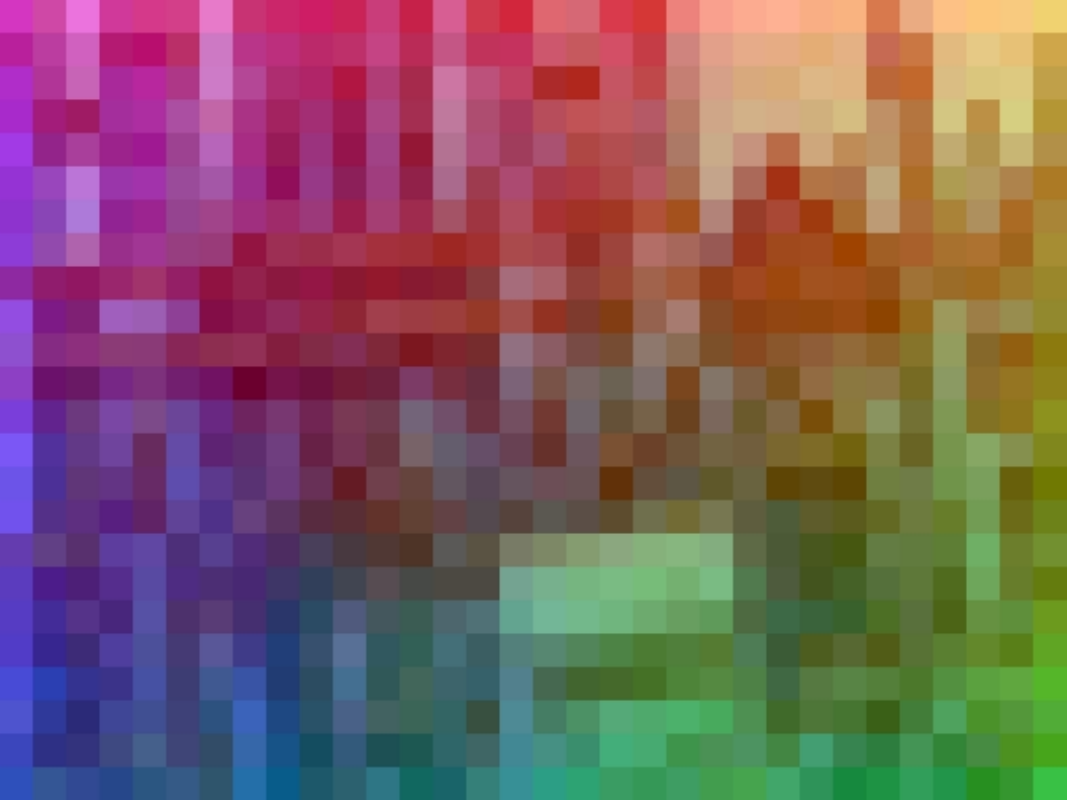} \\
        \includegraphics[width=0.166\linewidth]{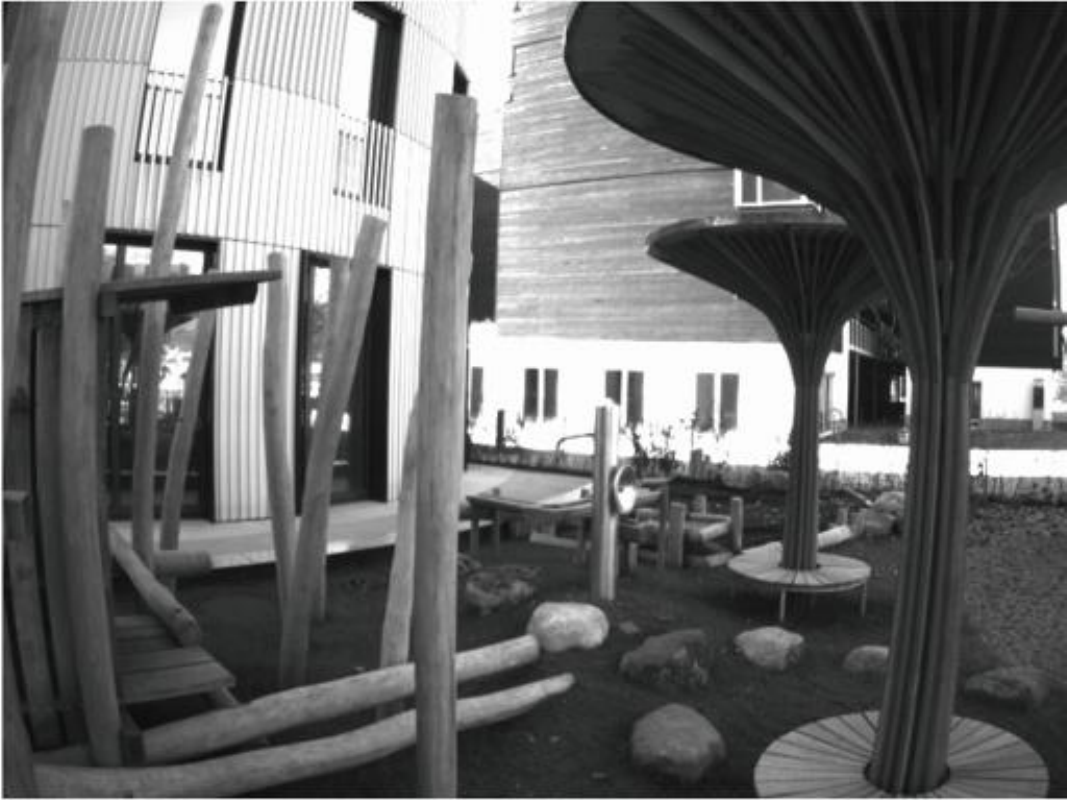} &
        \includegraphics[width=0.166\linewidth]{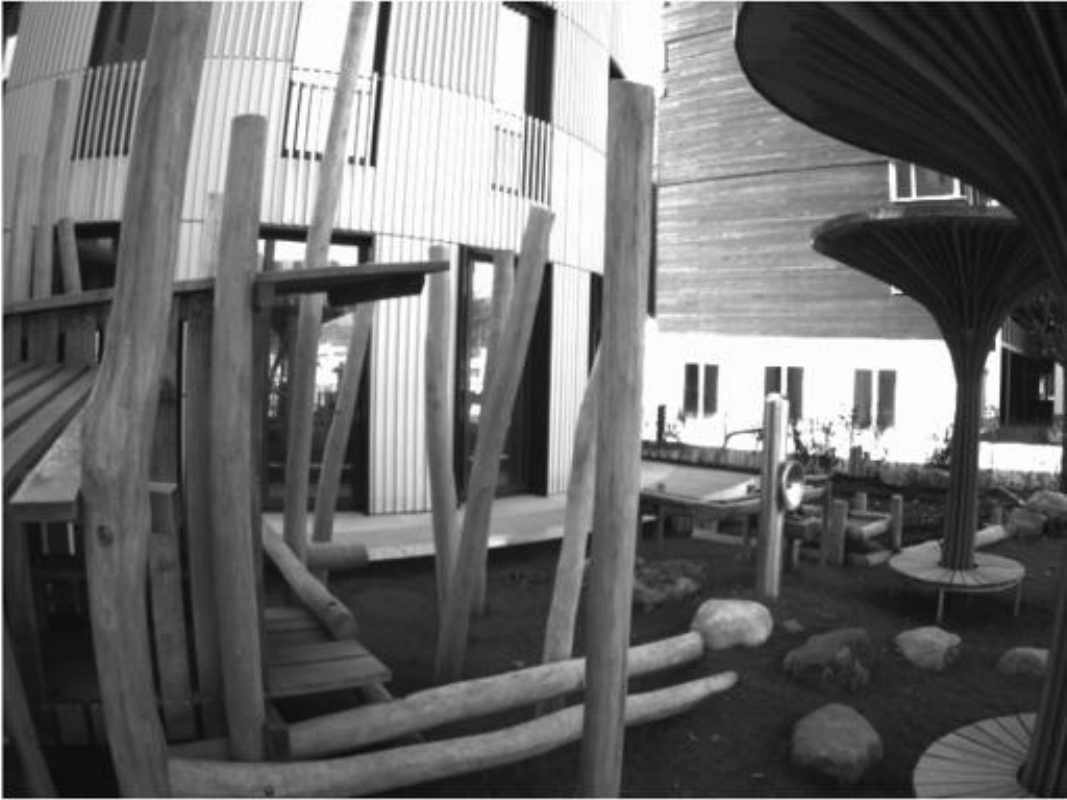} &
        \includegraphics[width=0.166\linewidth]{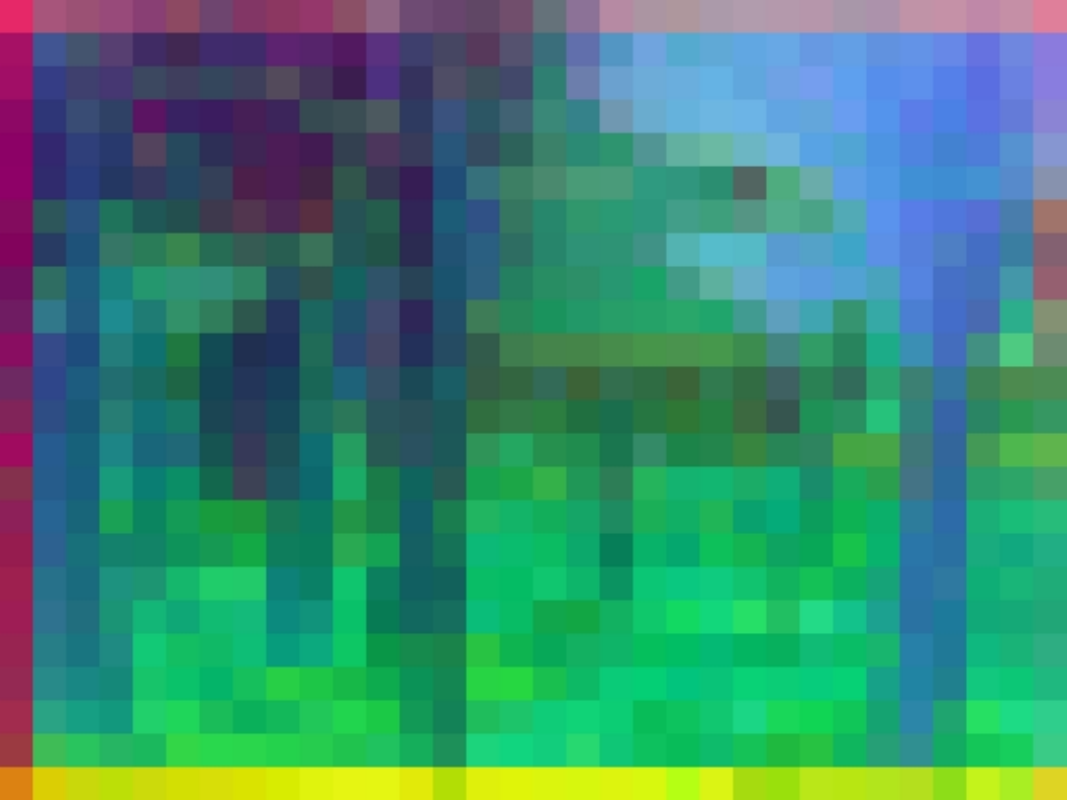} &
        \includegraphics[width=0.166\linewidth]{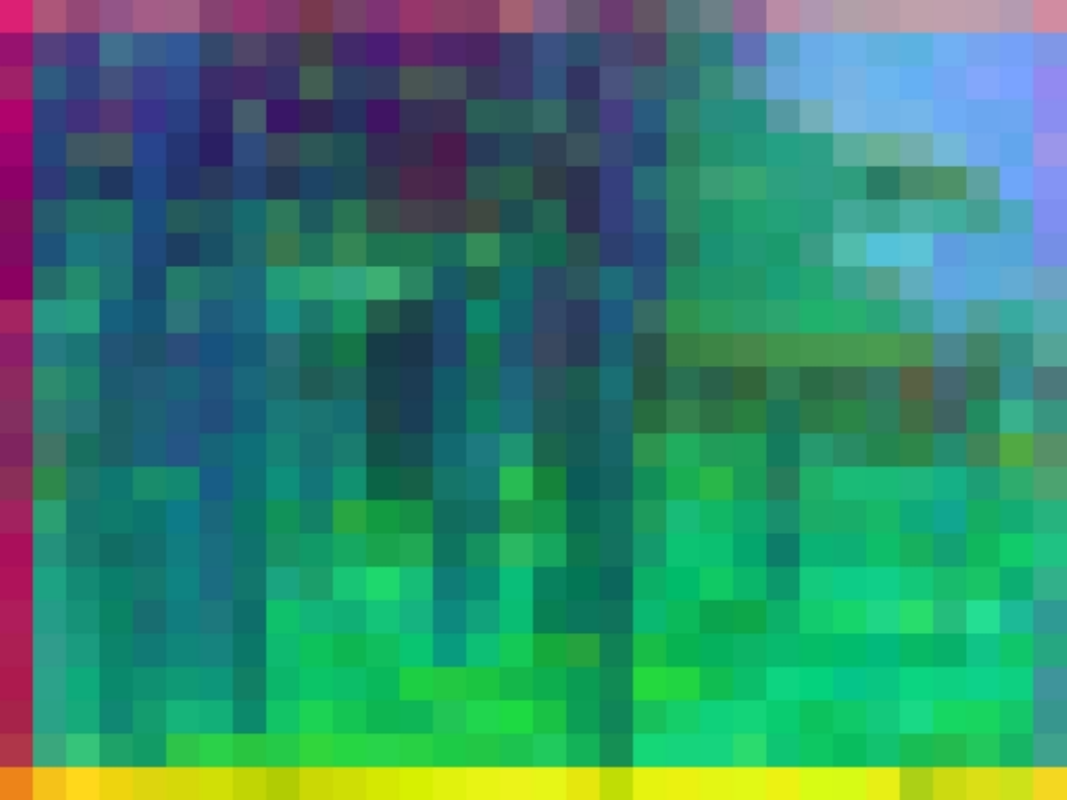} &
        \includegraphics[width=0.166\linewidth]{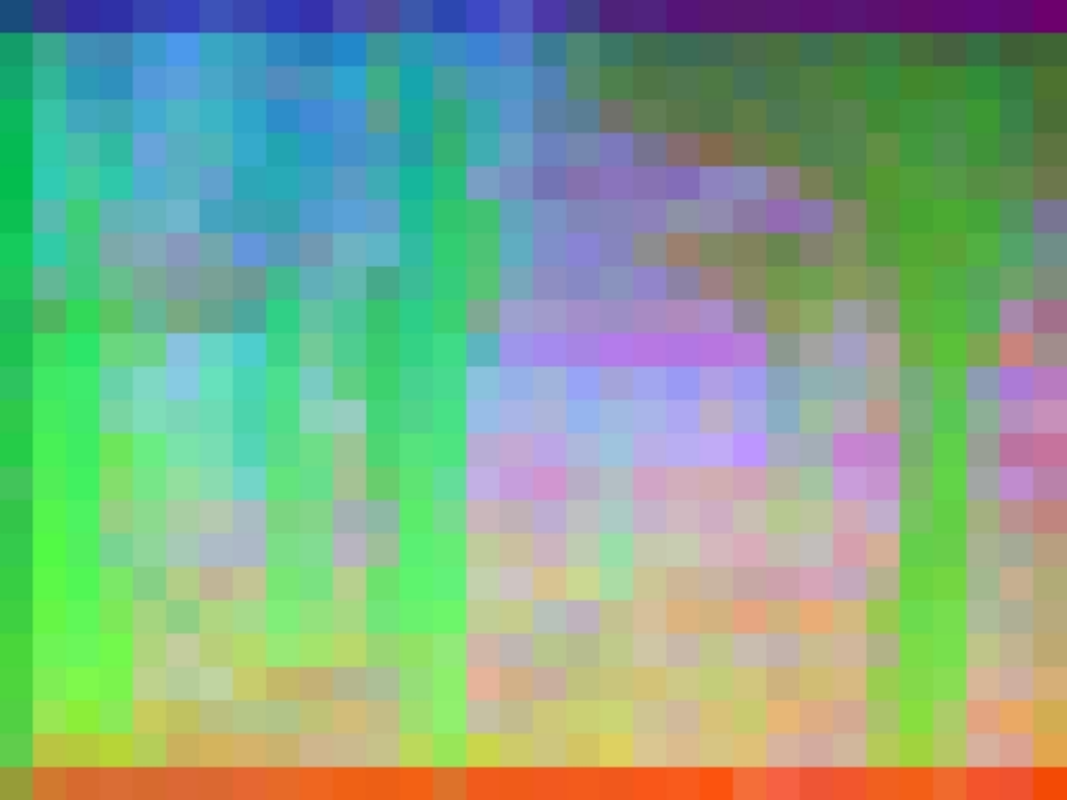} &
        \includegraphics[width=0.166\linewidth]{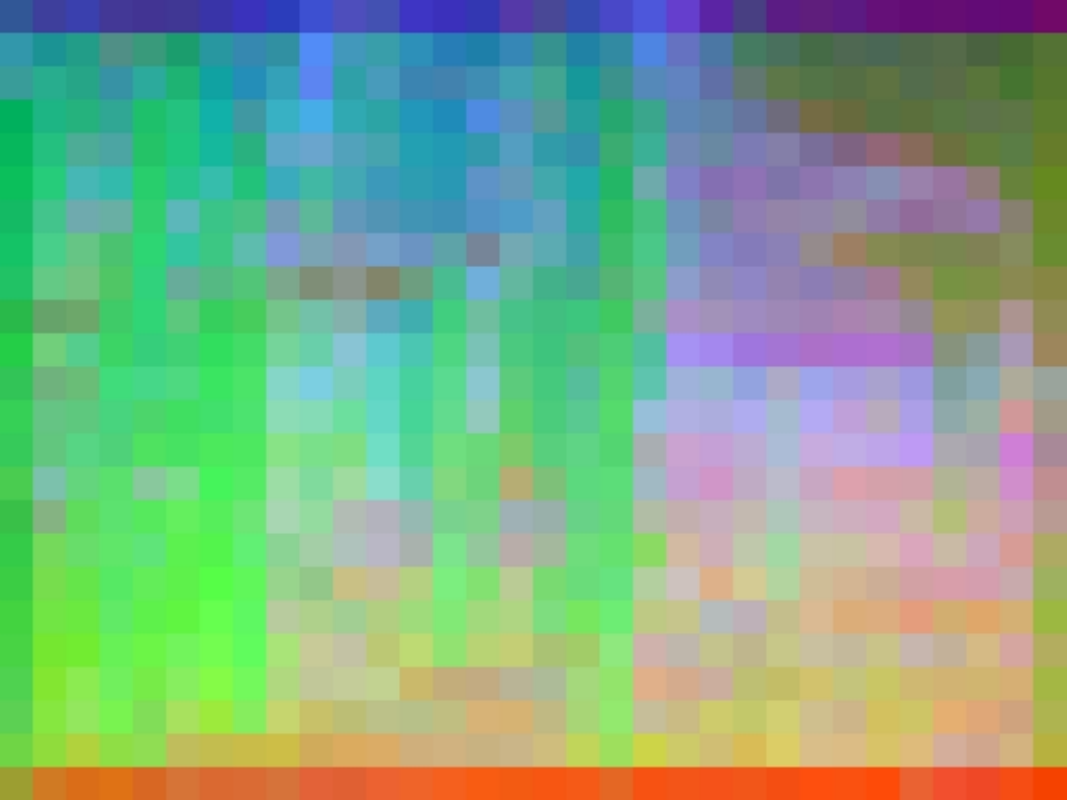} \\
        \includegraphics[width=0.166\linewidth]{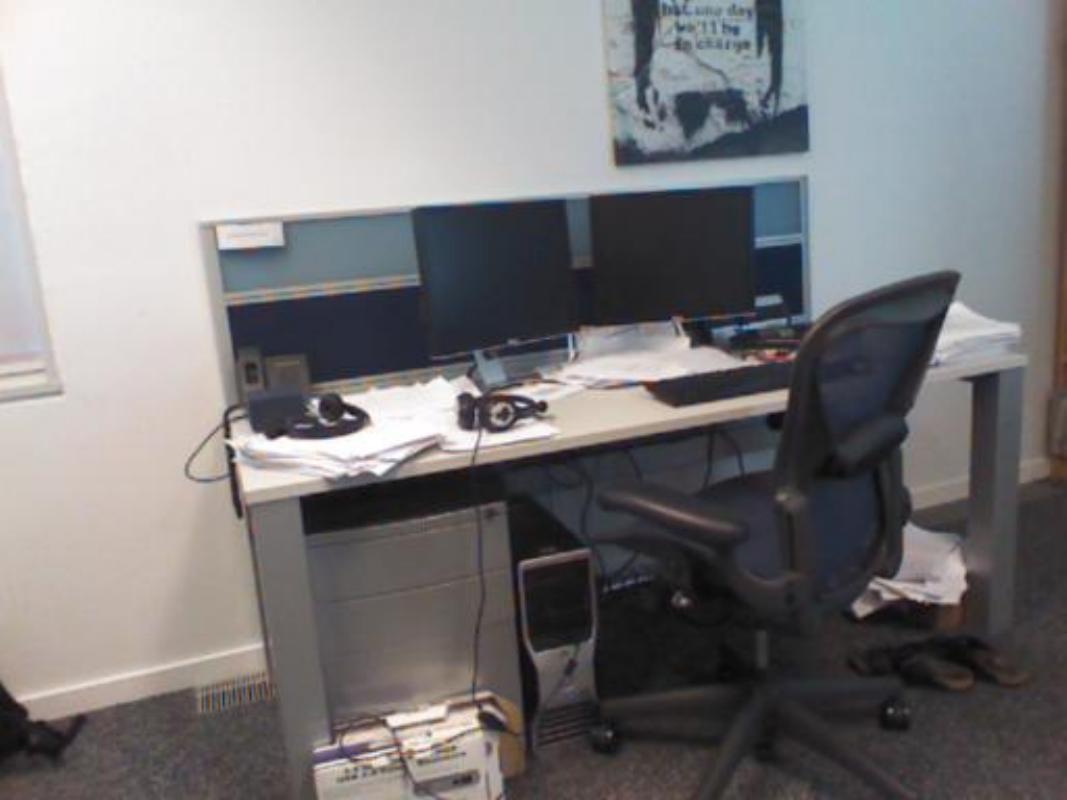} &
        \includegraphics[width=0.166\linewidth]{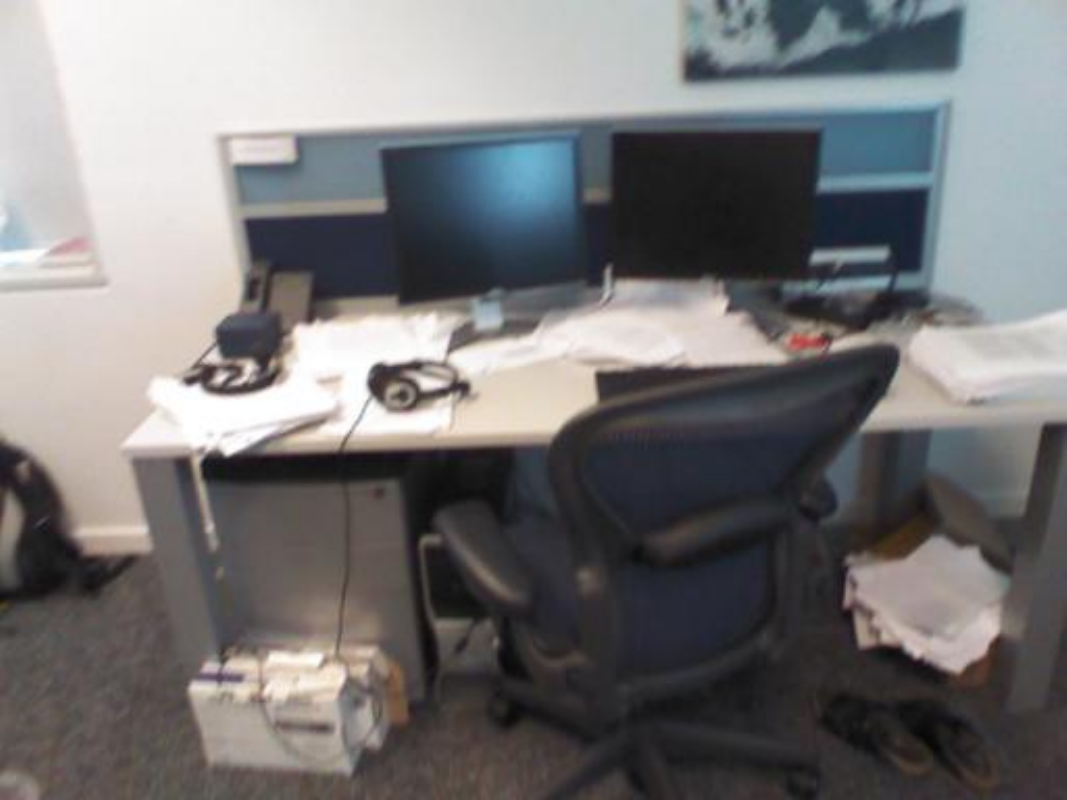} &
        \includegraphics[width=0.166\linewidth]{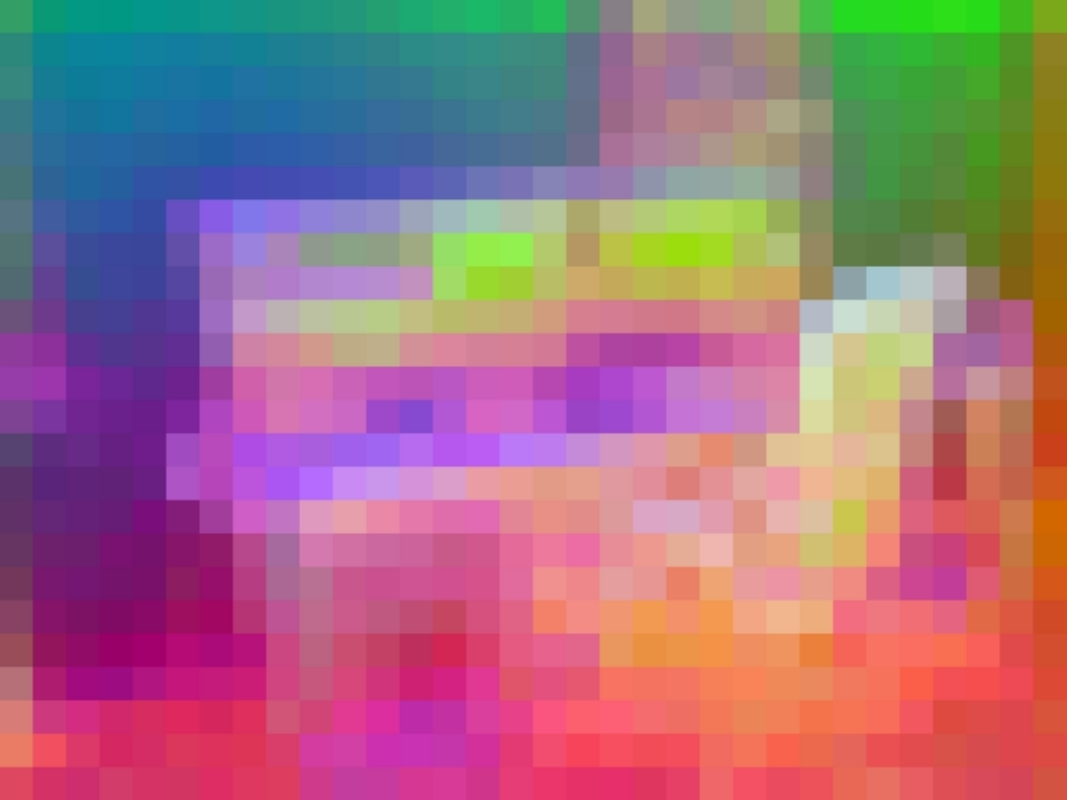} &
        \includegraphics[width=0.166\linewidth]{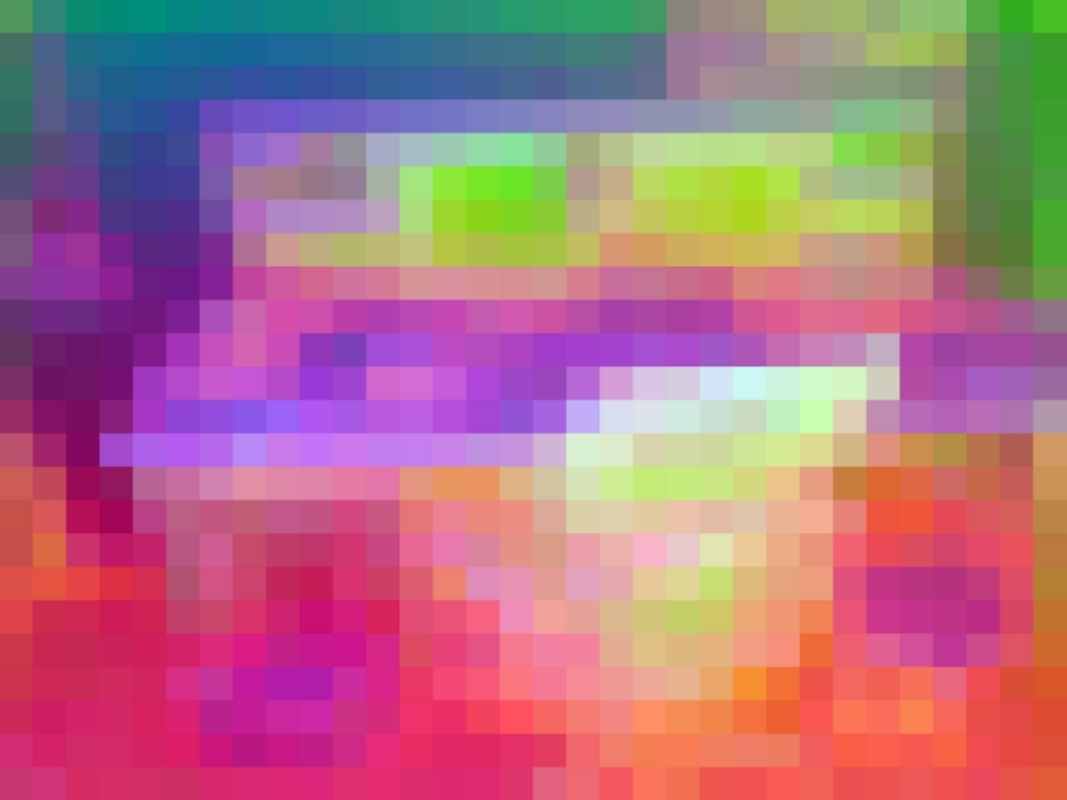} &
        \includegraphics[width=0.166\linewidth]{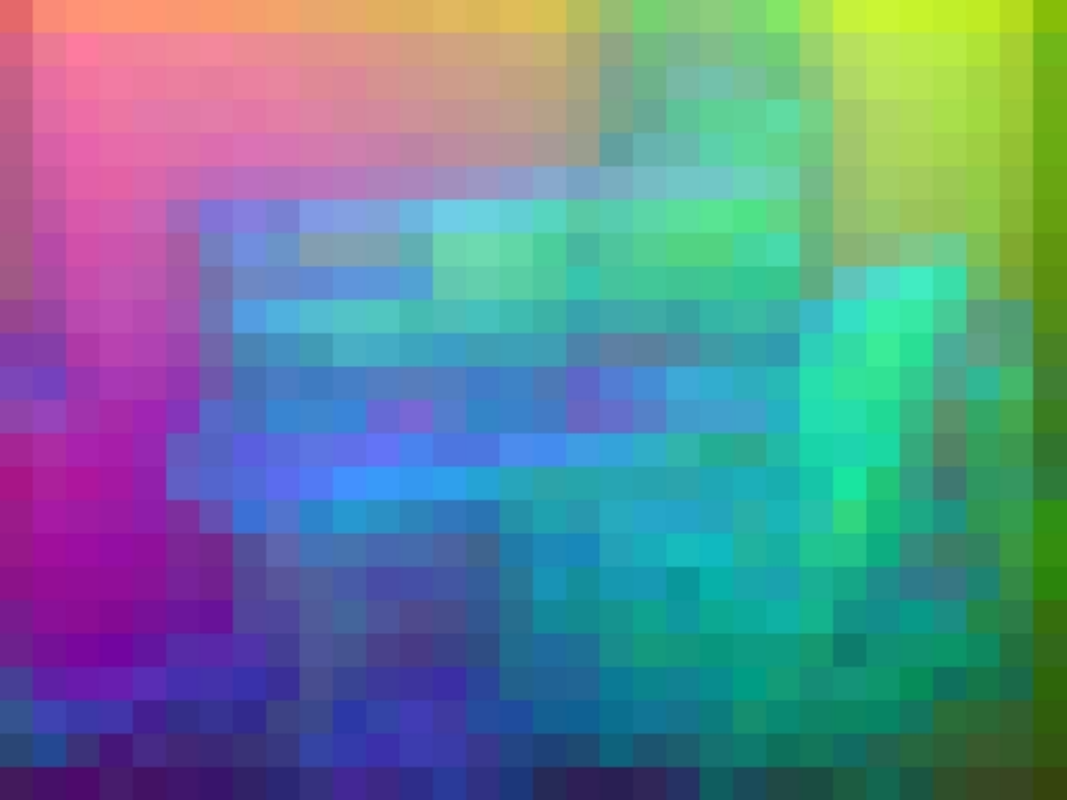} &
        \includegraphics[width=0.166\linewidth]{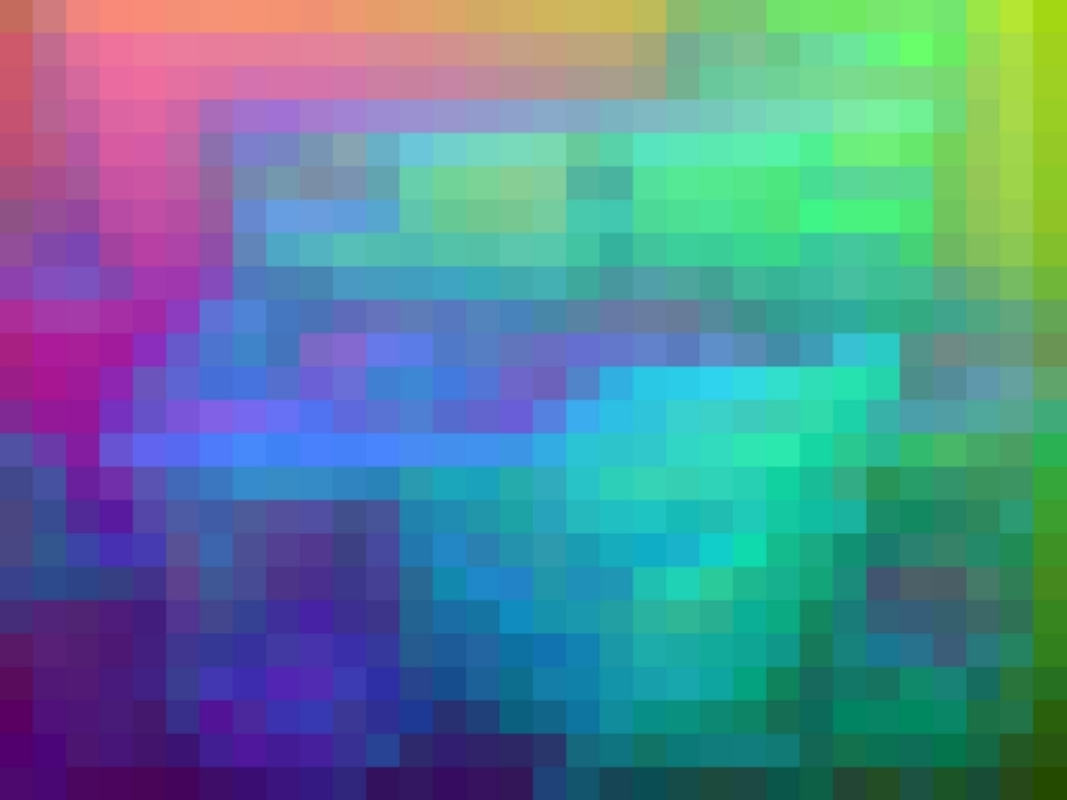} \\
        \includegraphics[width=0.166\linewidth]{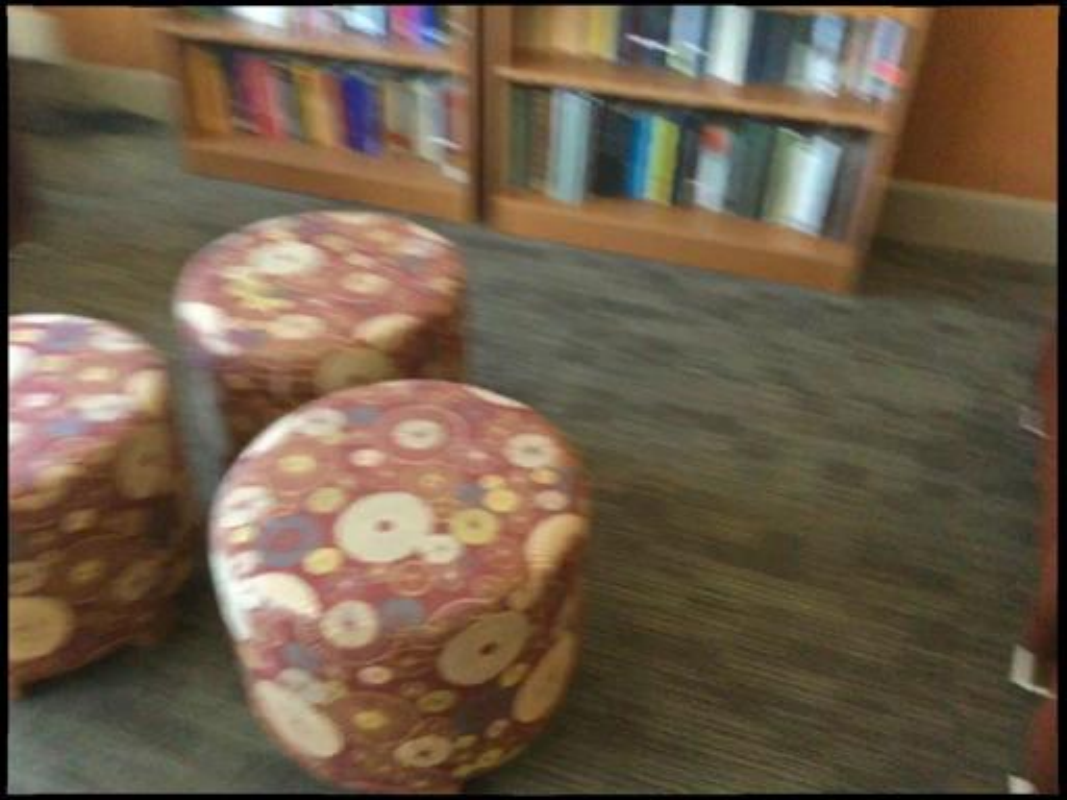} &
        \includegraphics[width=0.166\linewidth]{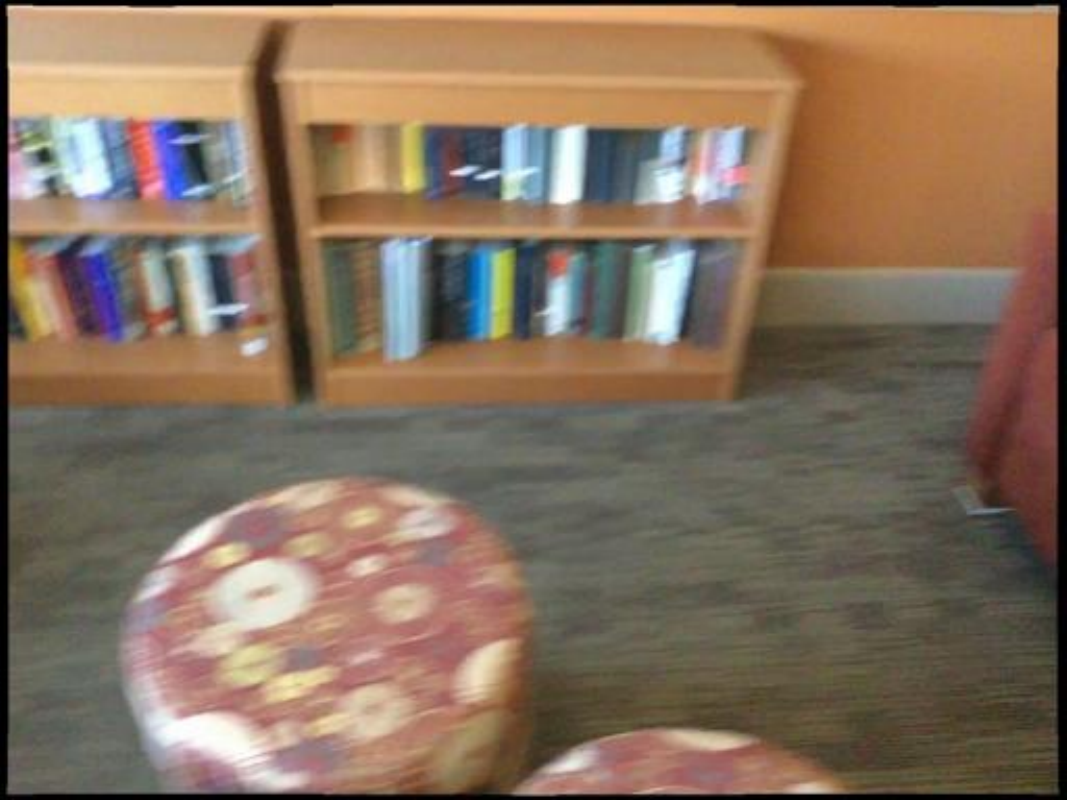} &
        \includegraphics[width=0.166\linewidth]{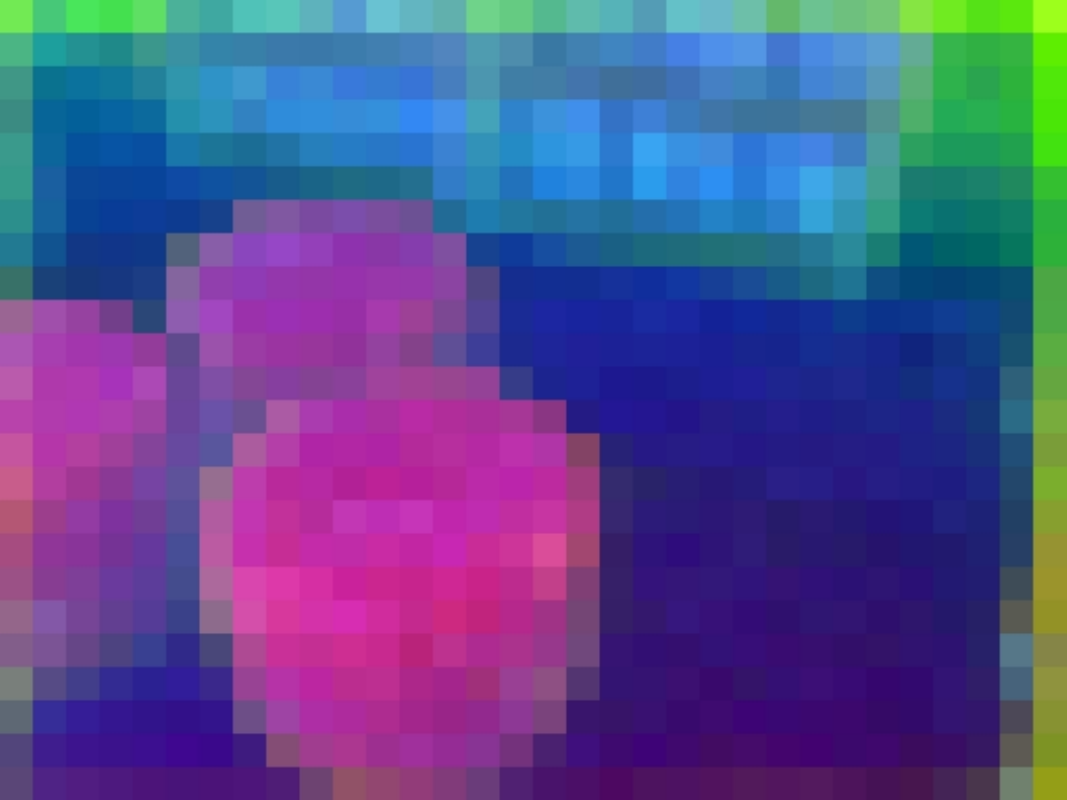} &
        \includegraphics[width=0.166\linewidth]{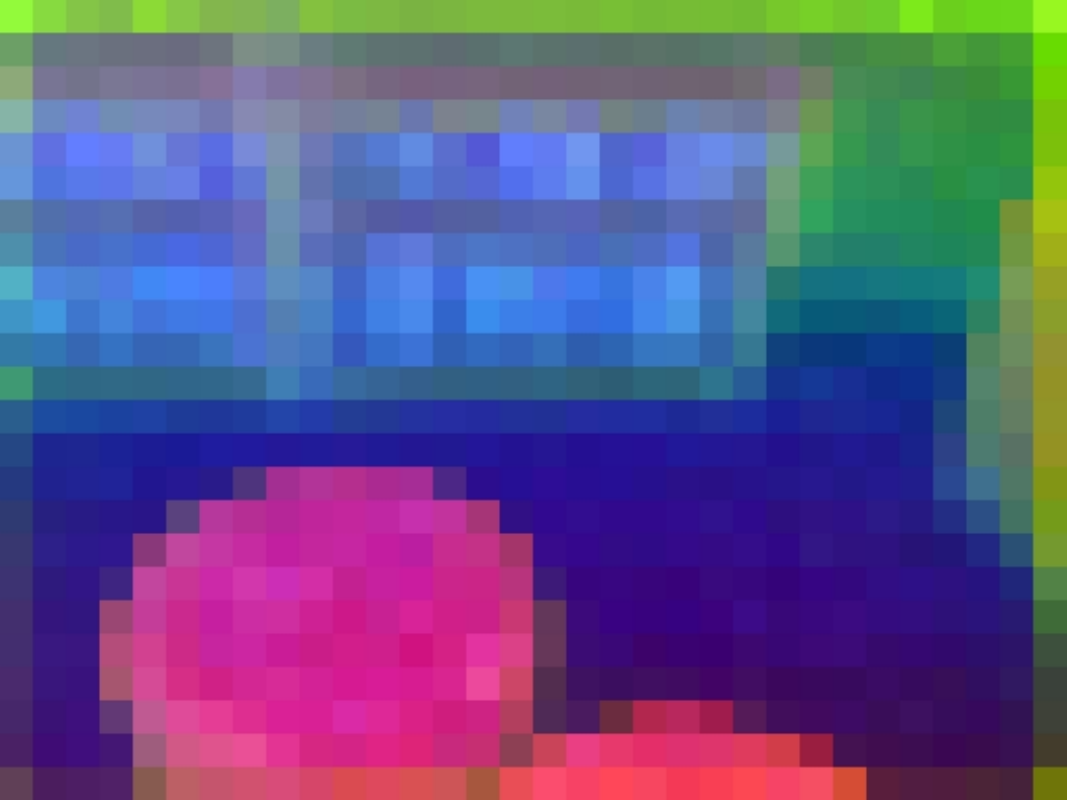} &
        \includegraphics[width=0.166\linewidth]{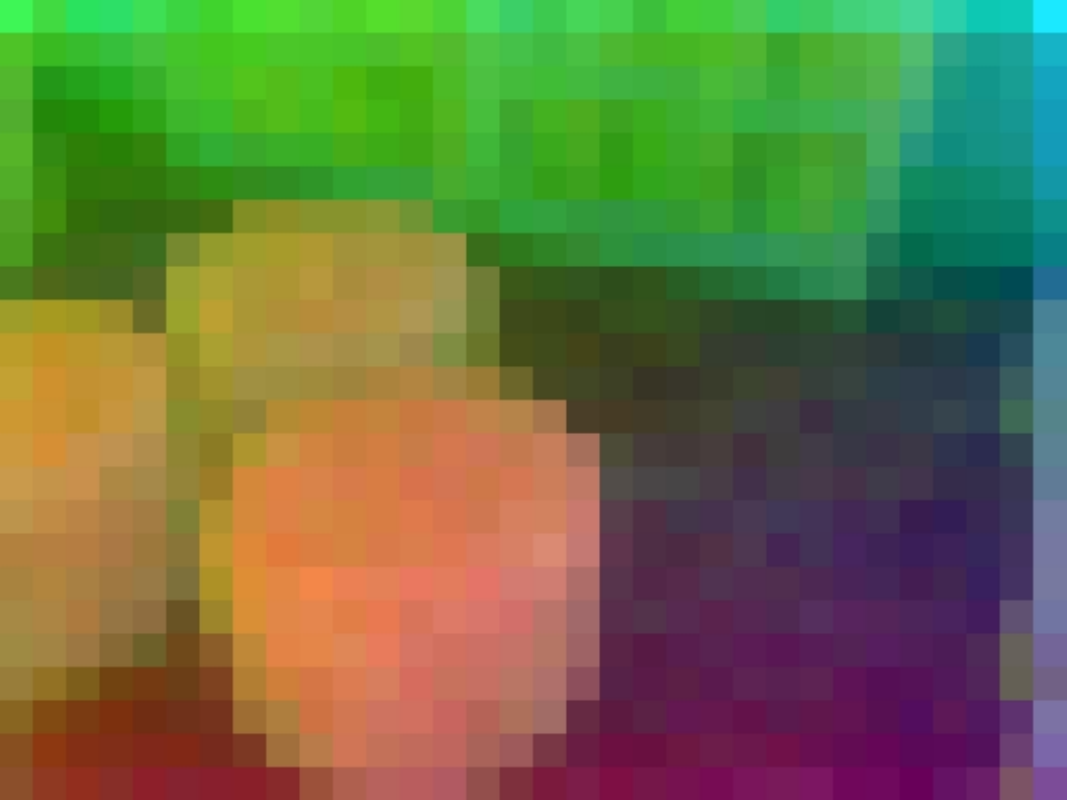} &
        \includegraphics[width=0.166\linewidth]{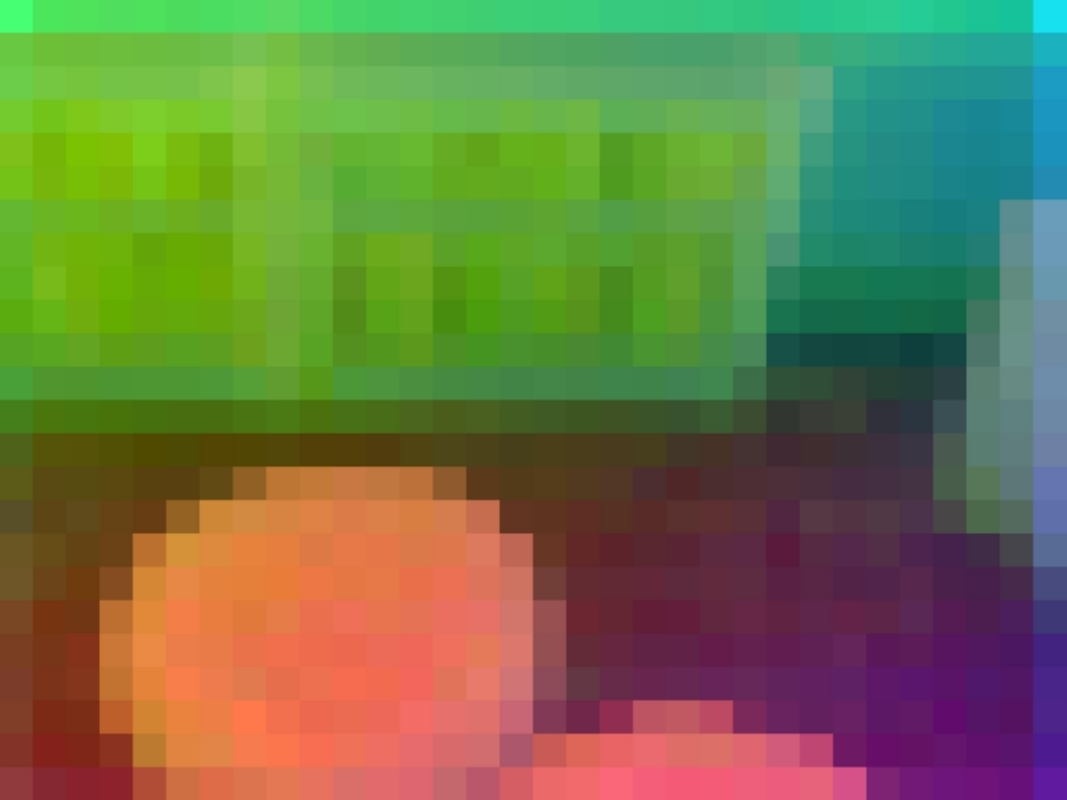} \\
        \includegraphics[width=0.166\linewidth]{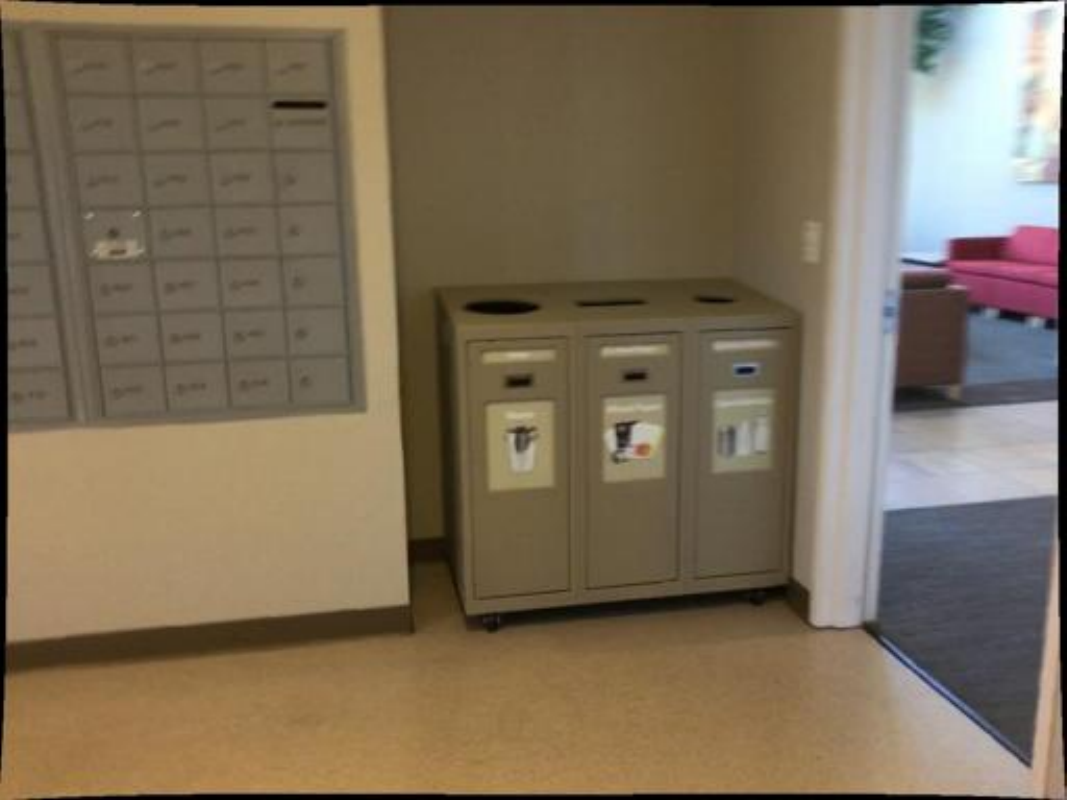} &
        \includegraphics[width=0.166\linewidth]{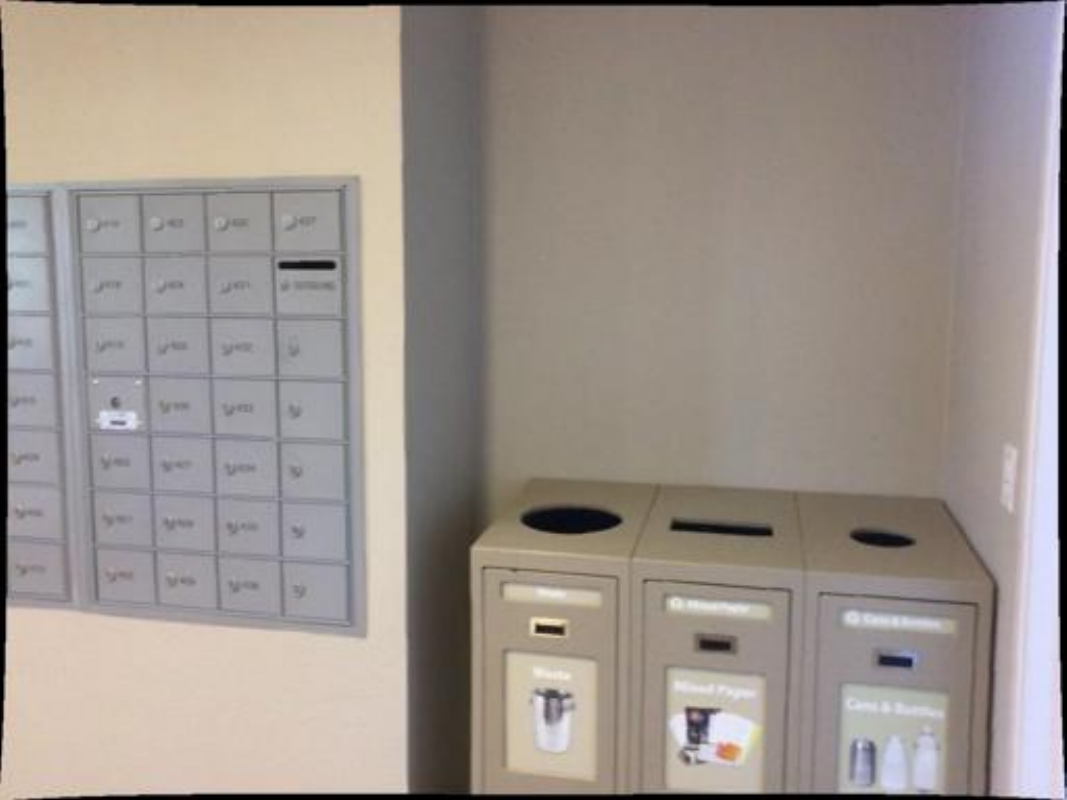} &
        \includegraphics[width=0.166\linewidth]{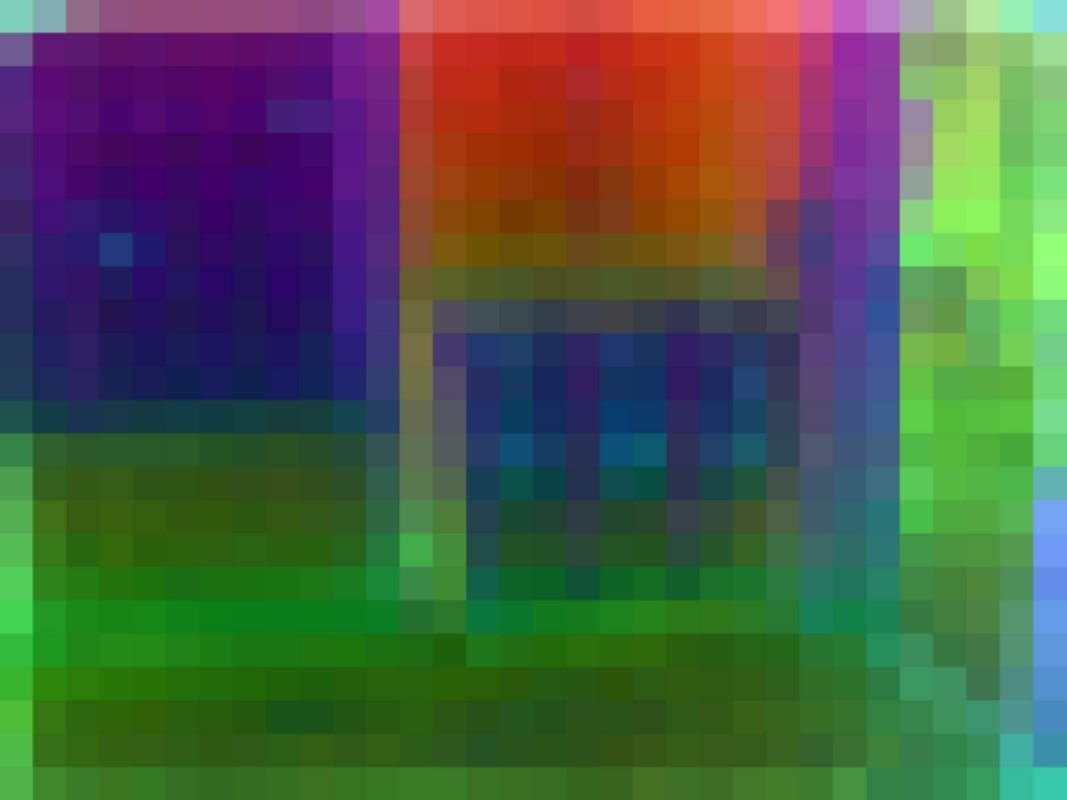} &
        \includegraphics[width=0.166\linewidth]{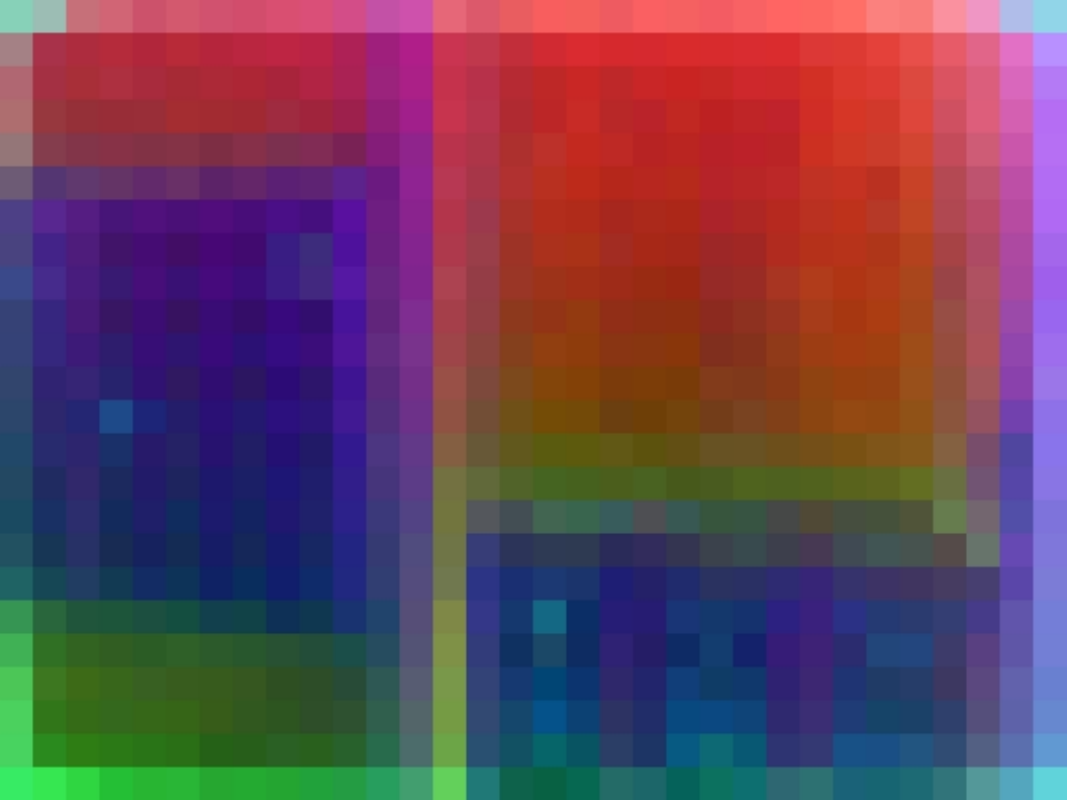} &
        \includegraphics[width=0.166\linewidth]{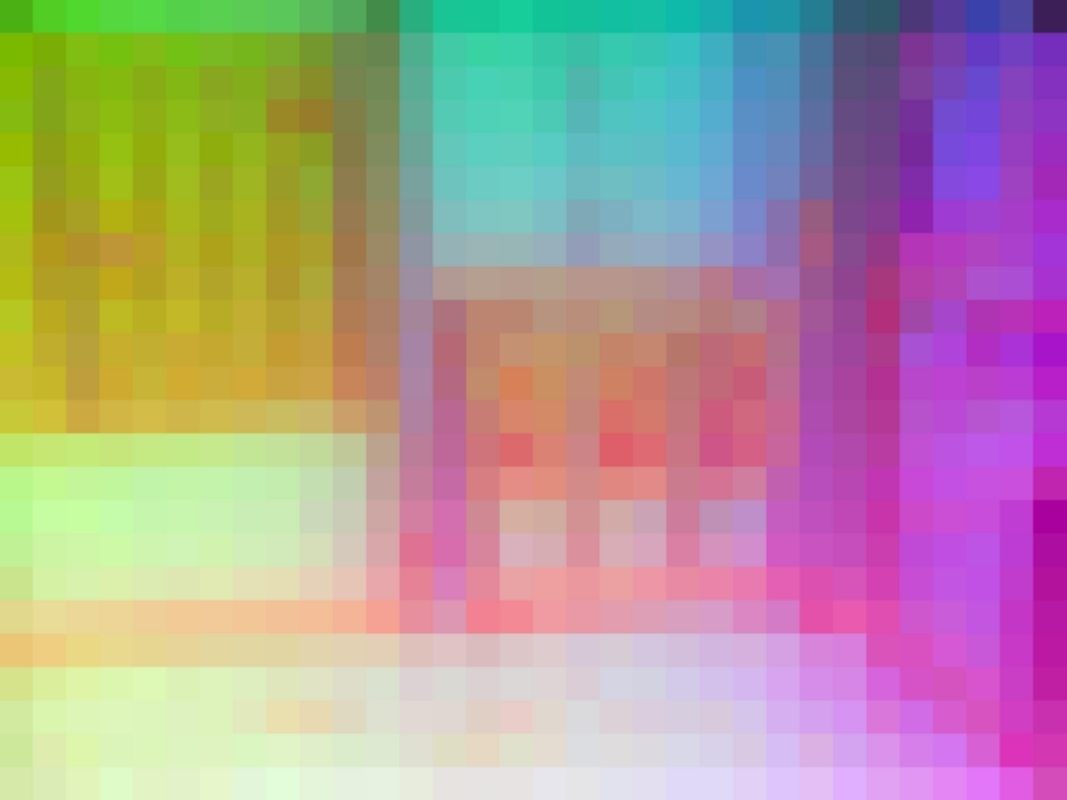} &
        \includegraphics[width=0.166\linewidth]{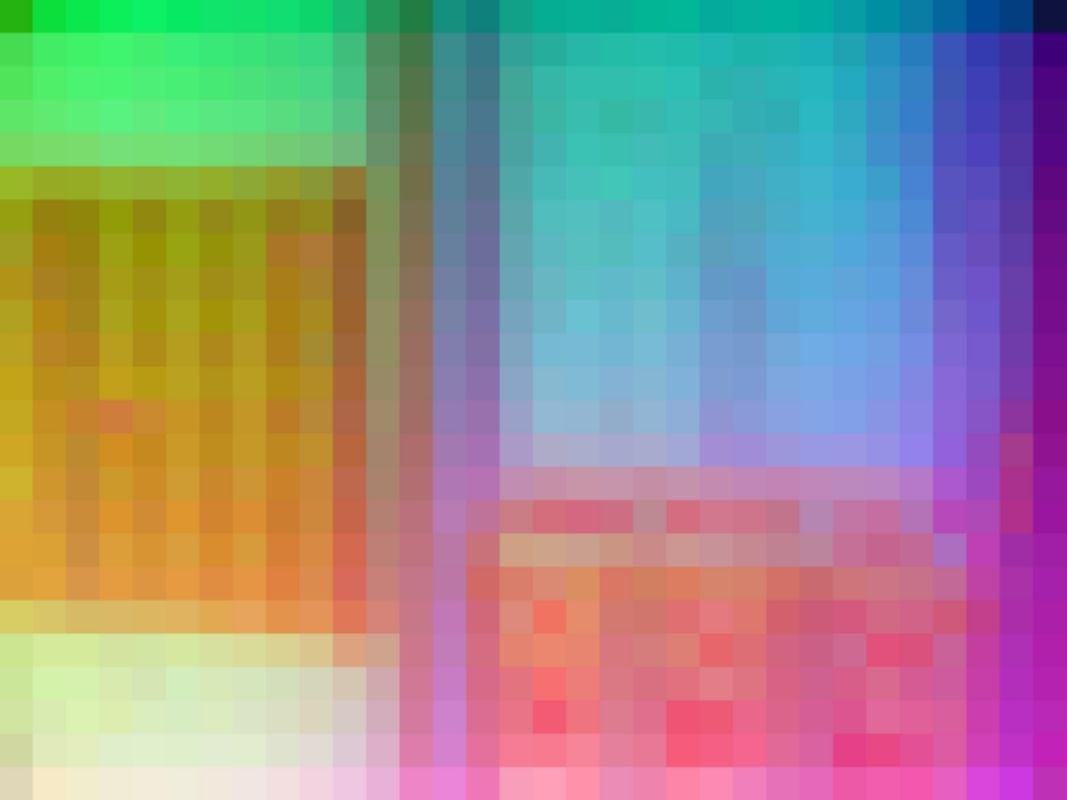} \\
        \includegraphics[width=0.166\linewidth]{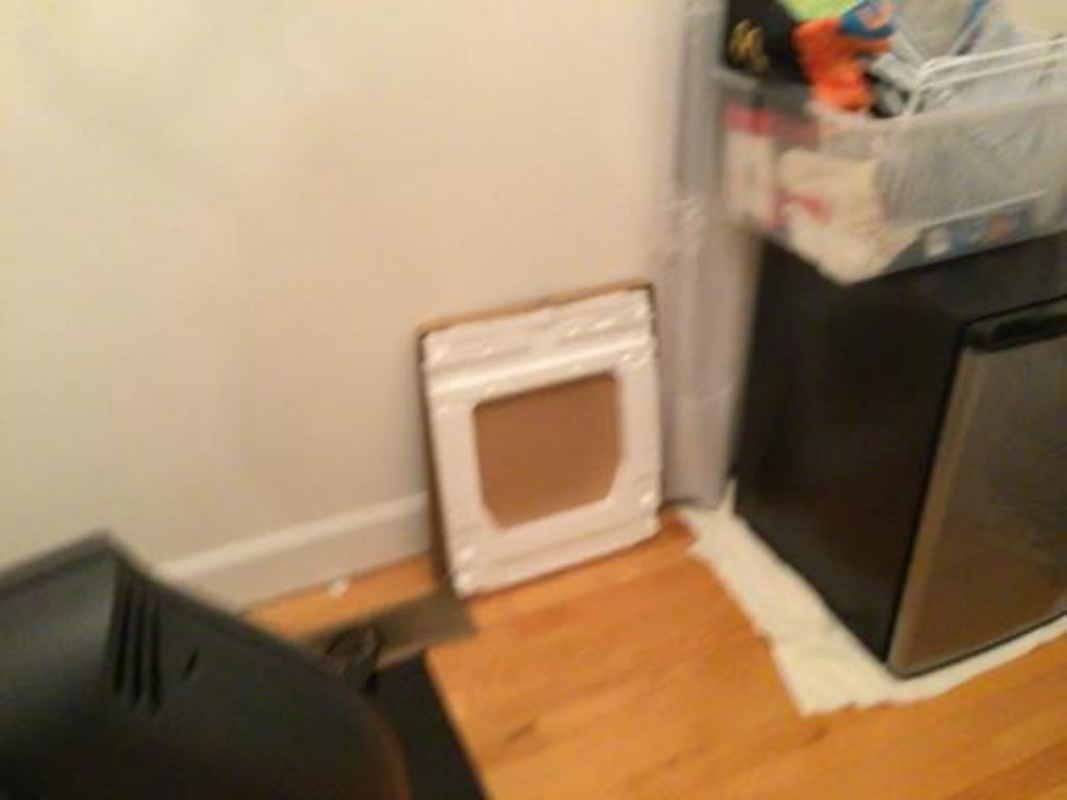} &
        \includegraphics[width=0.166\linewidth]{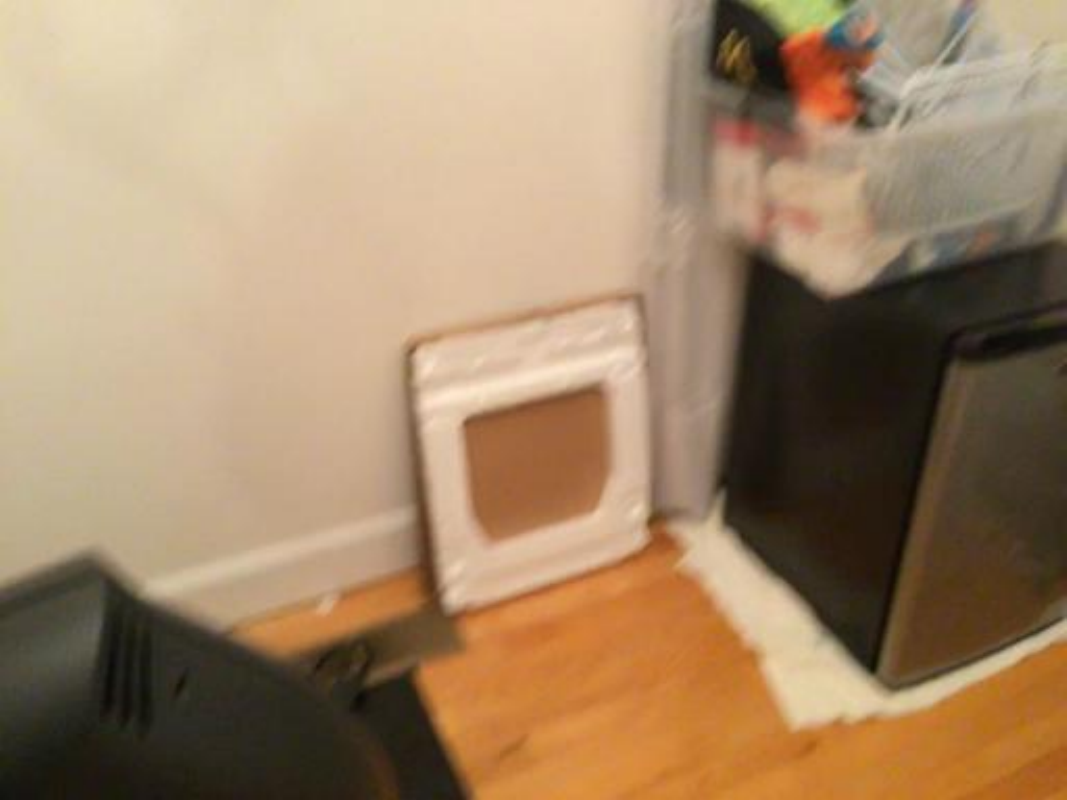} &
        \includegraphics[width=0.166\linewidth]{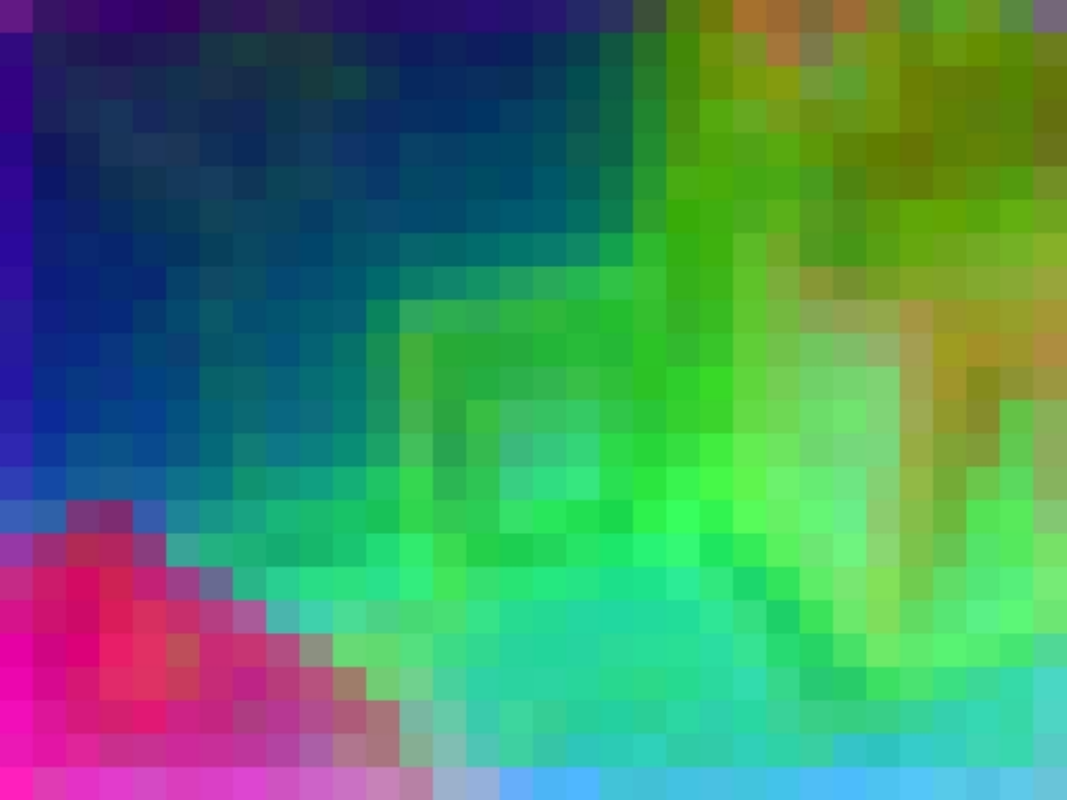} &
        \includegraphics[width=0.166\linewidth]{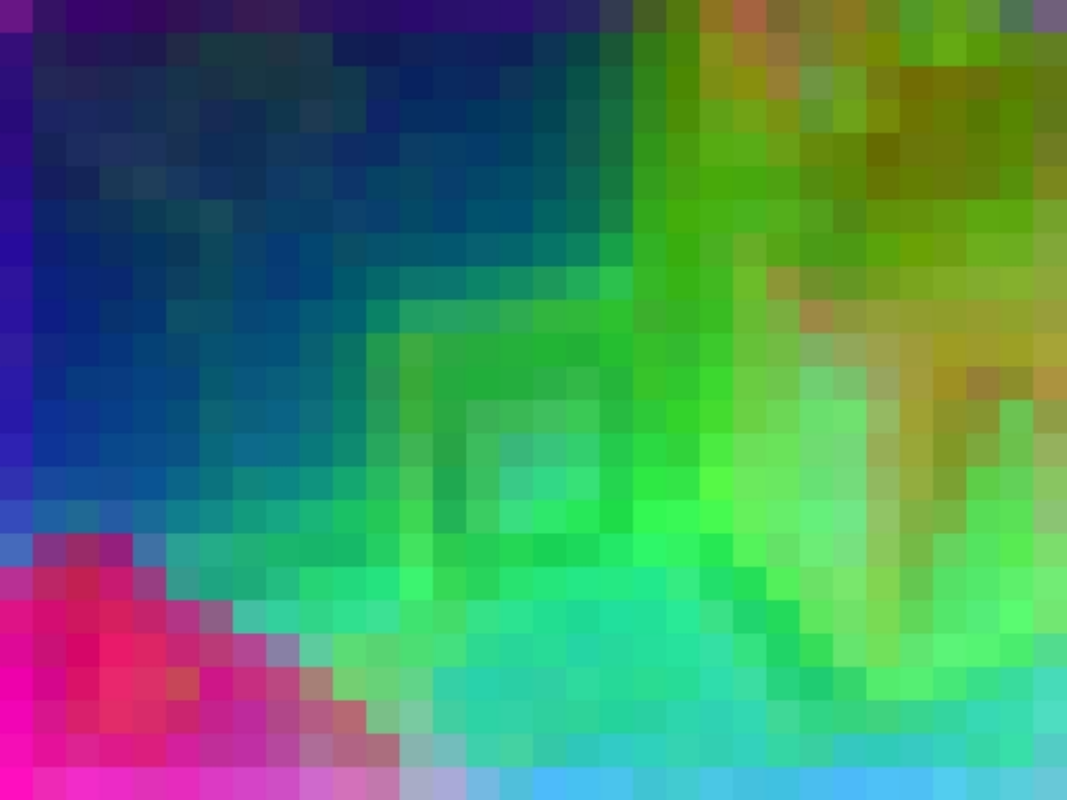} &
        \includegraphics[width=0.166\linewidth]{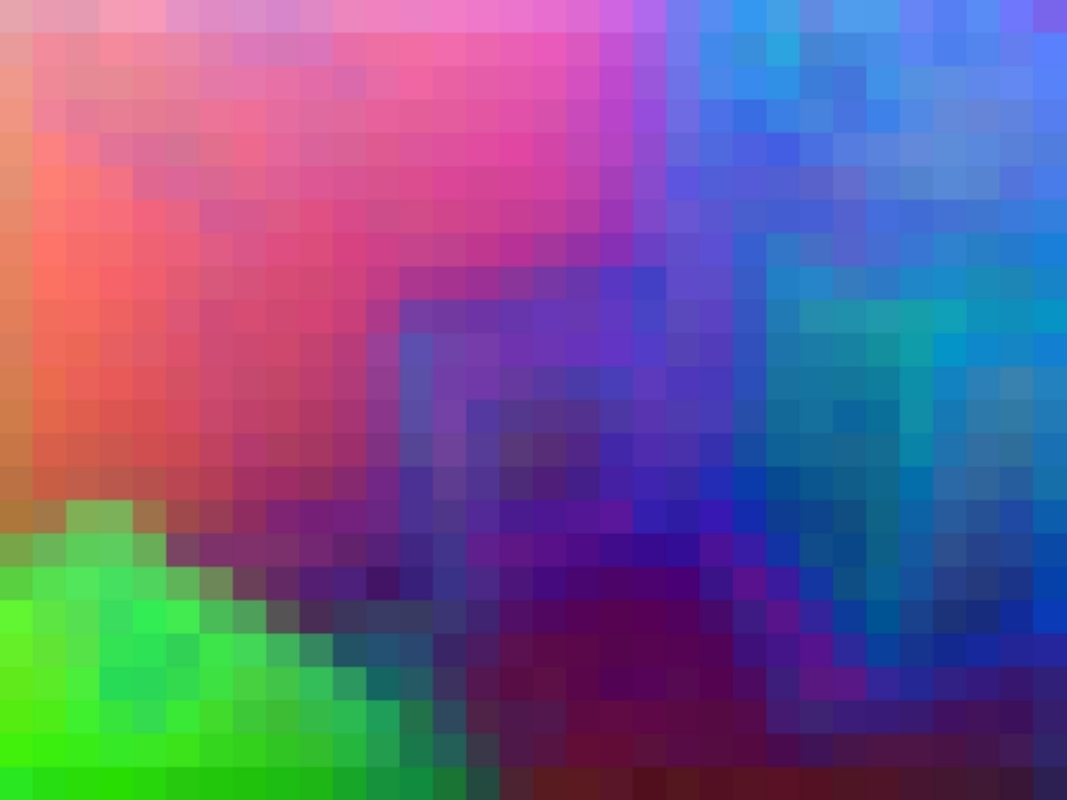} &
        \includegraphics[width=0.166\linewidth]{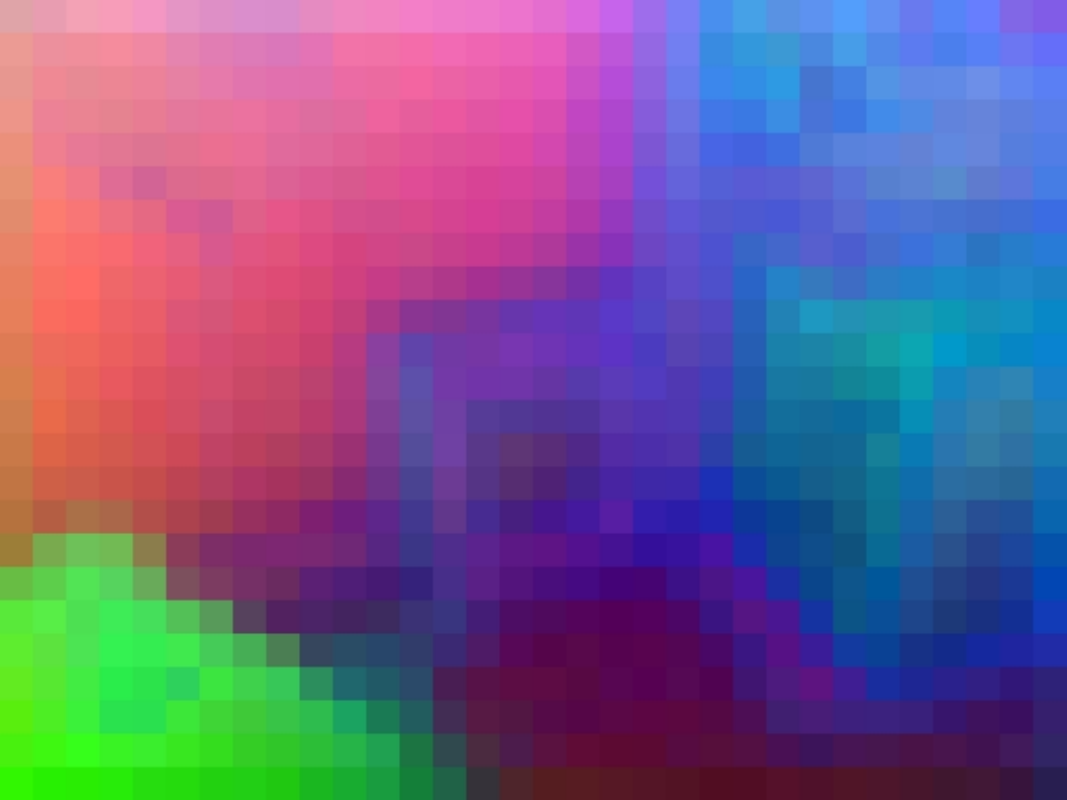} \\
        \includegraphics[width=0.166\linewidth]{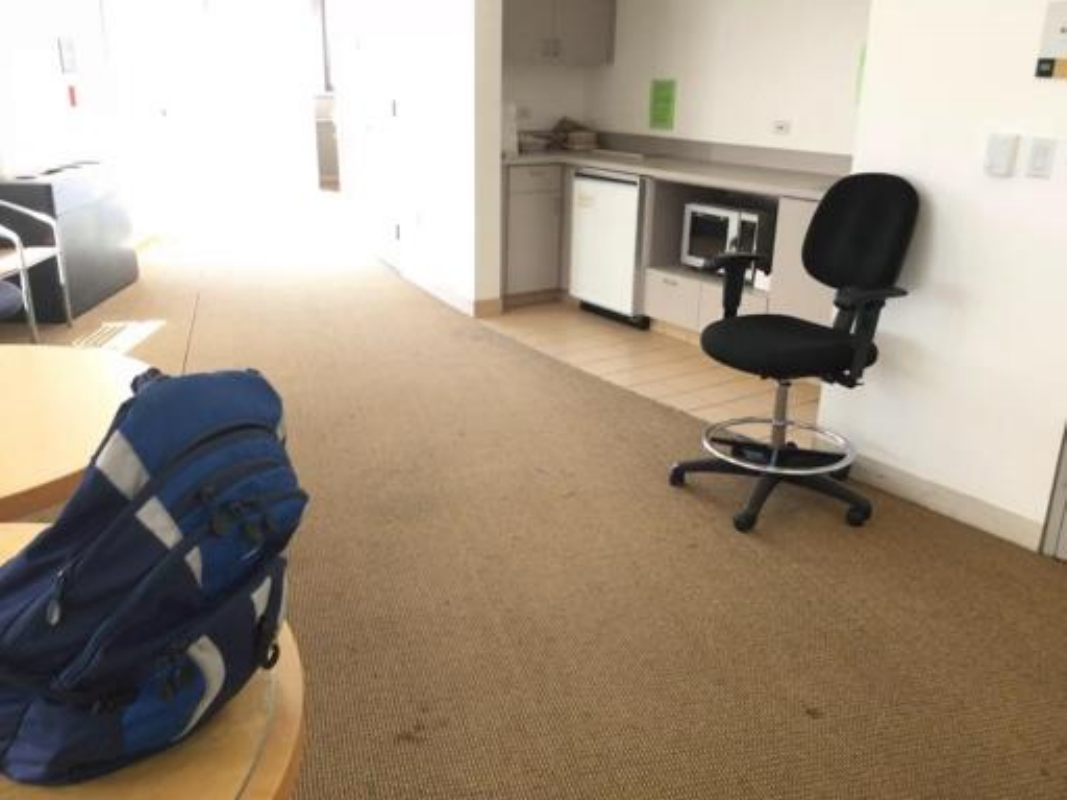} &
        \includegraphics[width=0.166\linewidth]{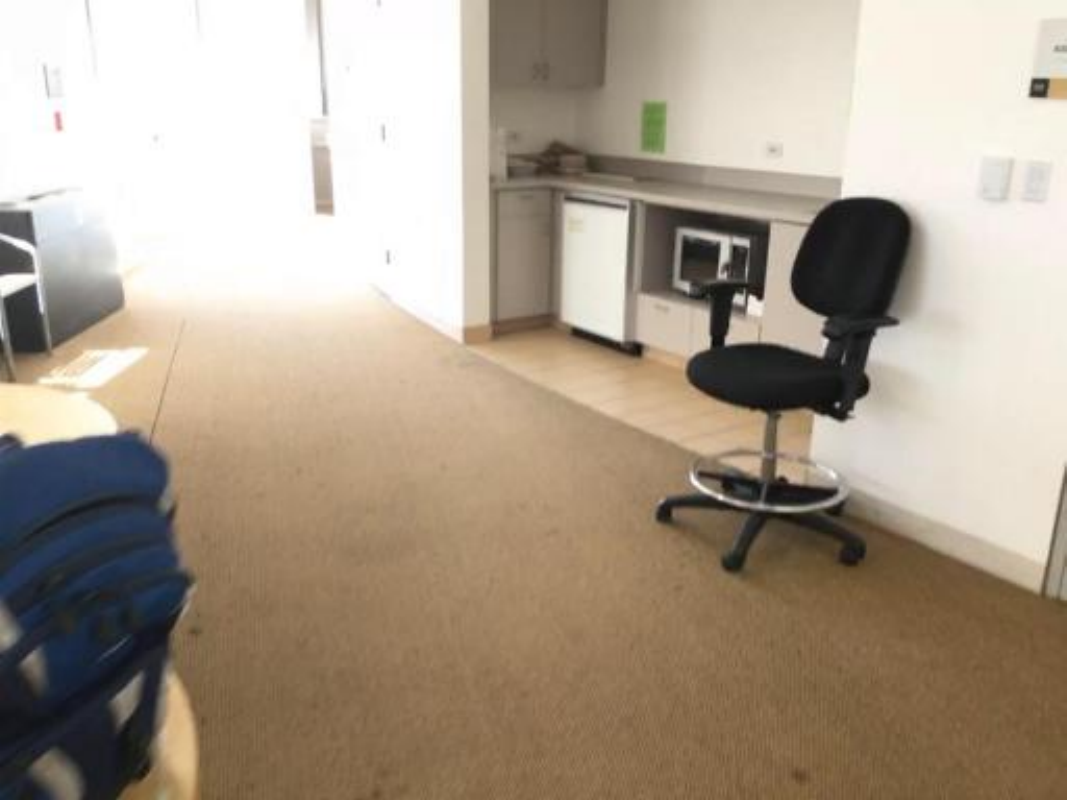} &
        \includegraphics[width=0.166\linewidth]{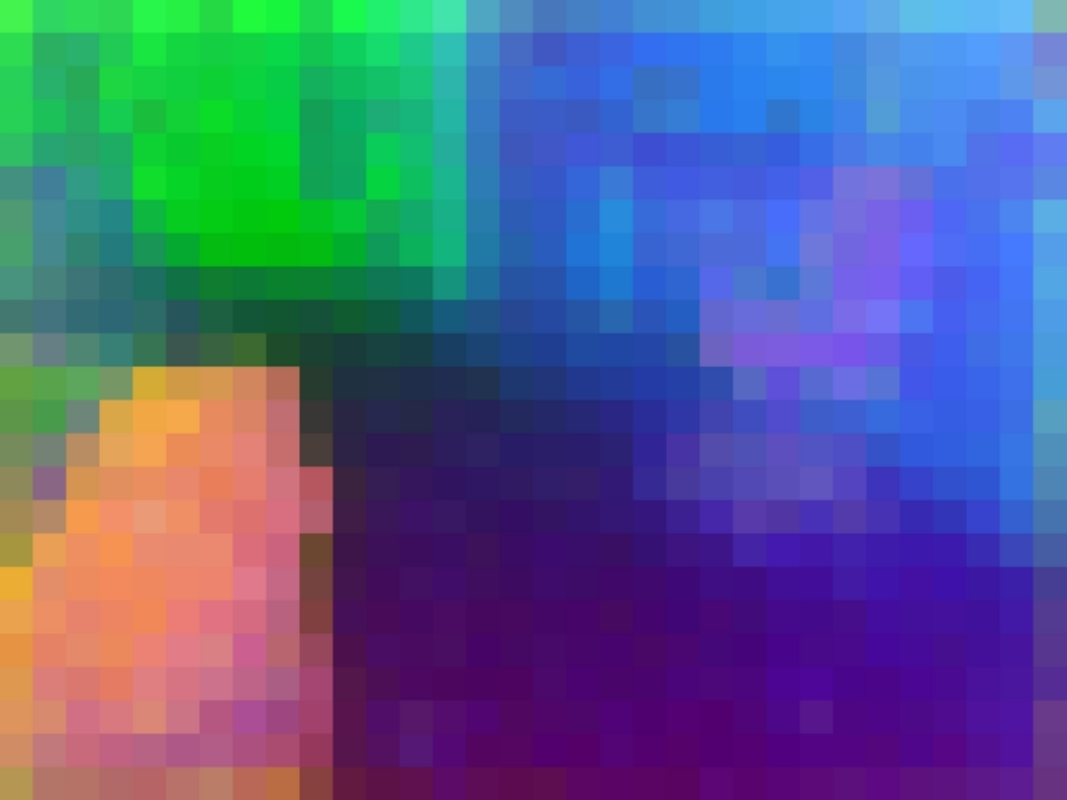} &
        \includegraphics[width=0.166\linewidth]{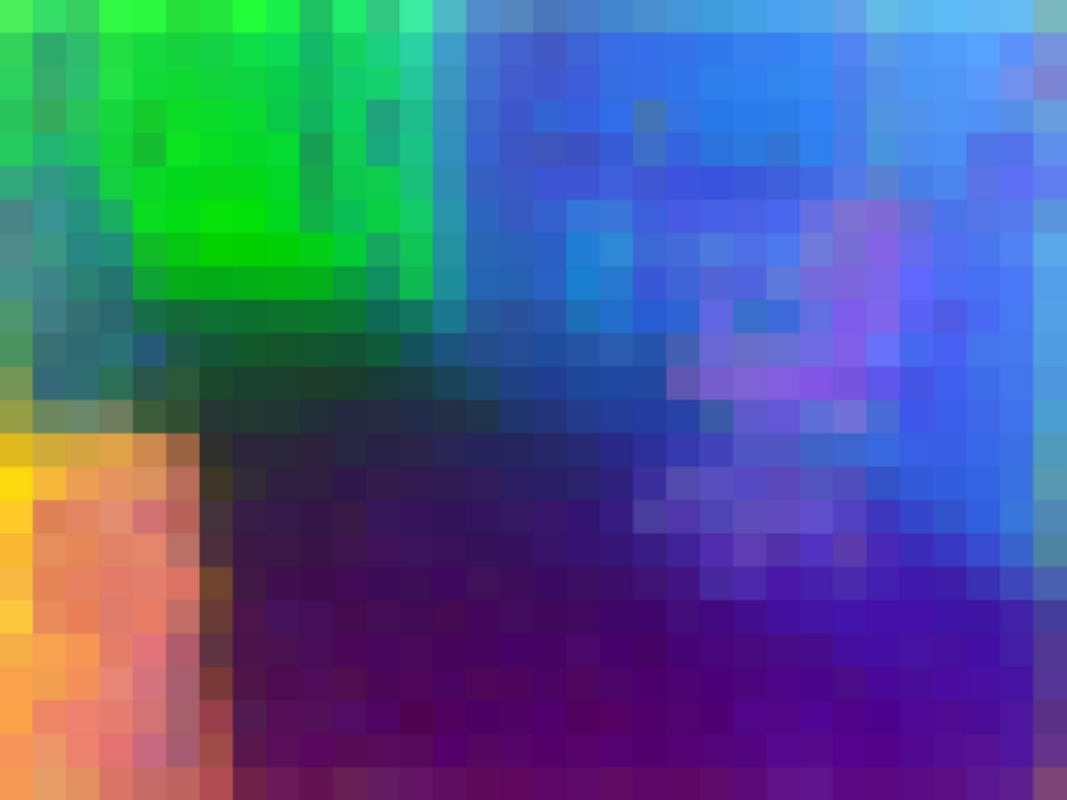} &
        \includegraphics[width=0.166\linewidth]{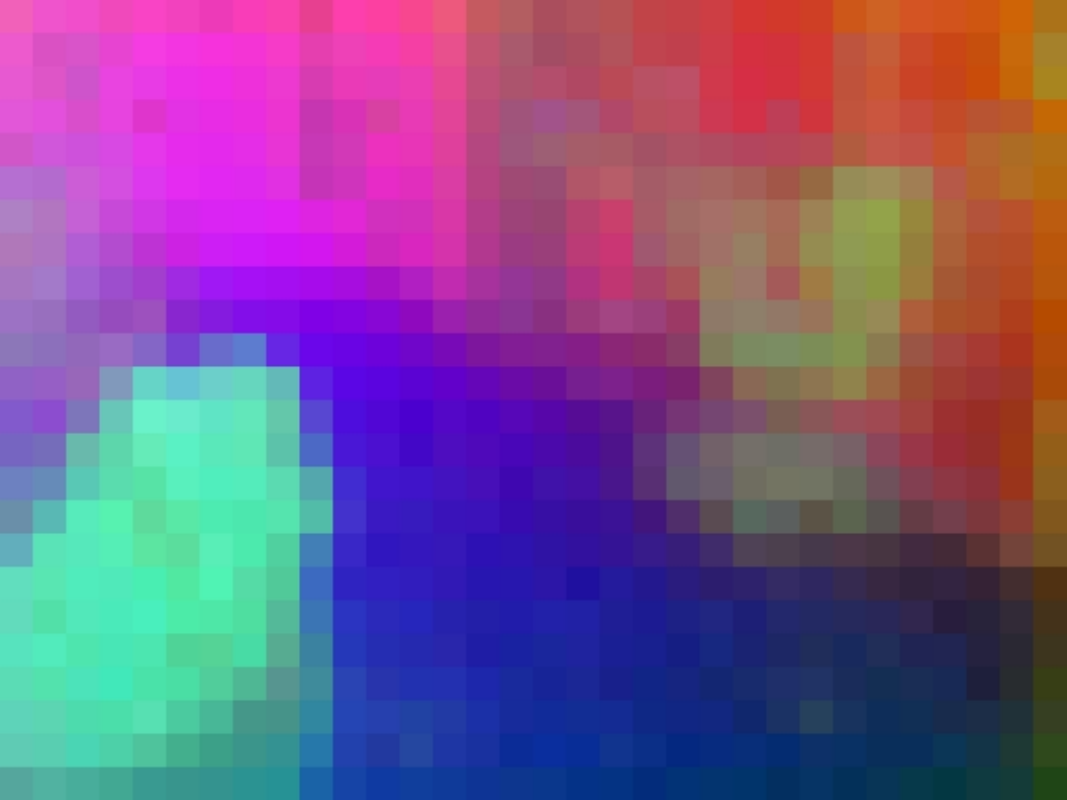} &
        \includegraphics[width=0.166\linewidth]{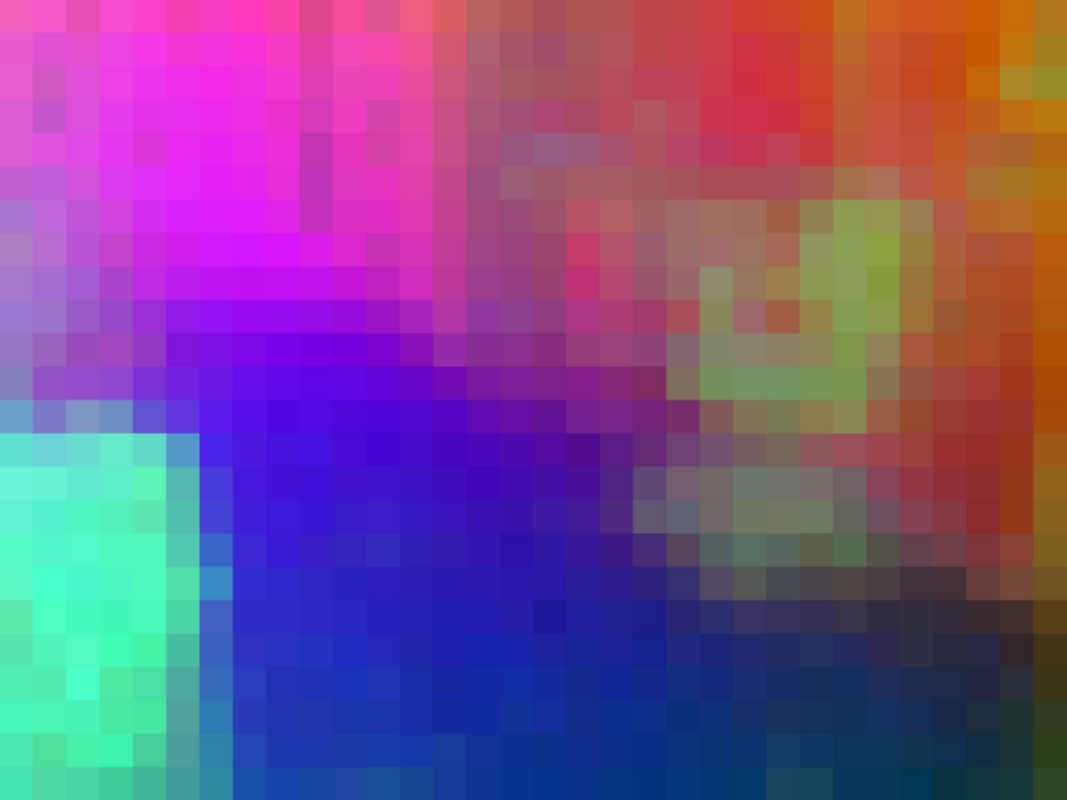} \\
        \includegraphics[width=0.166\linewidth]{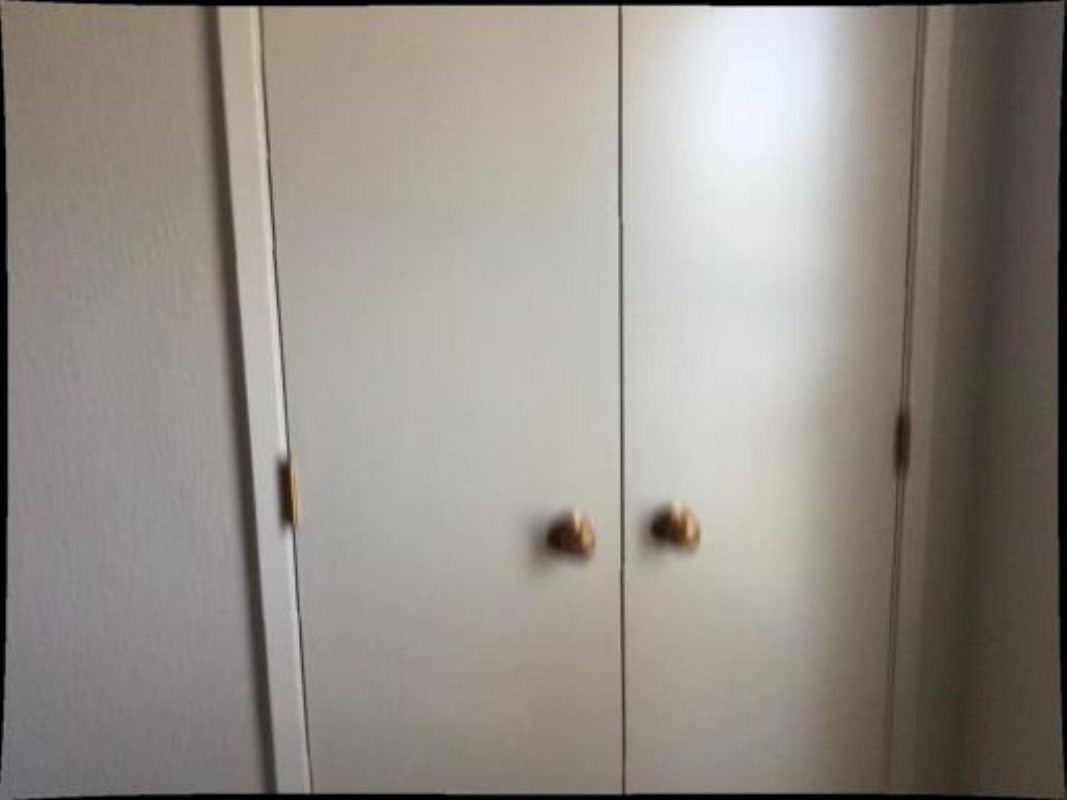} &
        \includegraphics[width=0.166\linewidth]{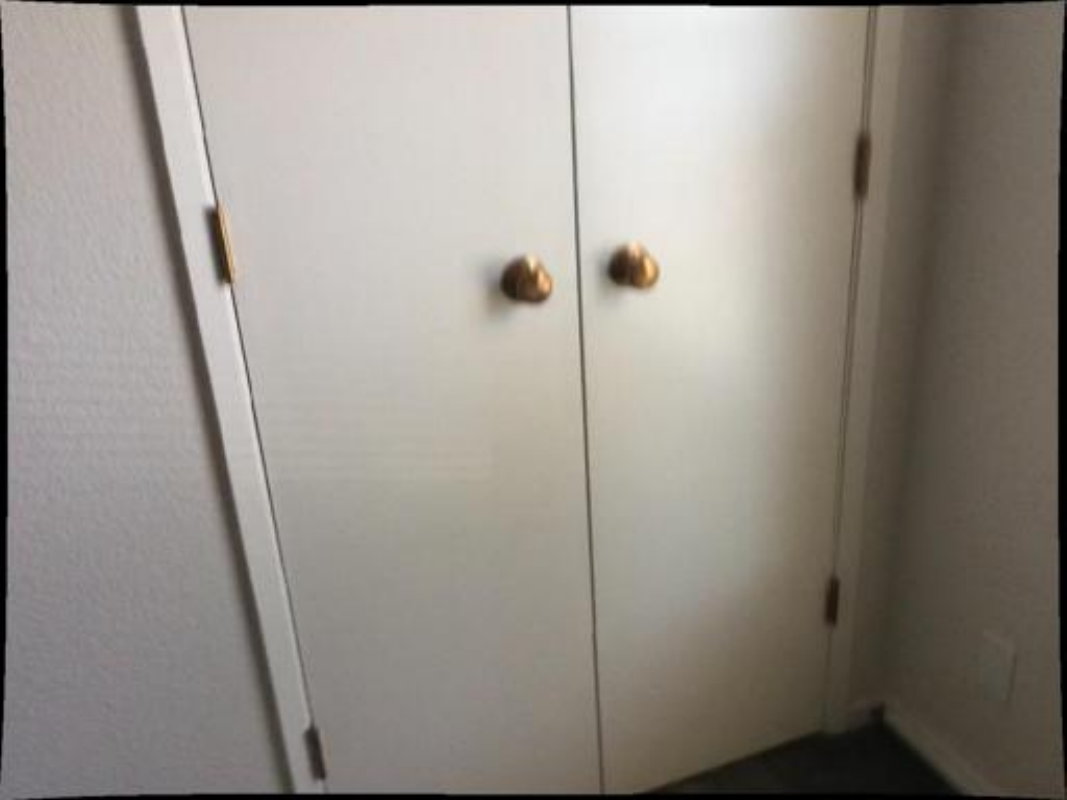} &
        \includegraphics[width=0.166\linewidth]{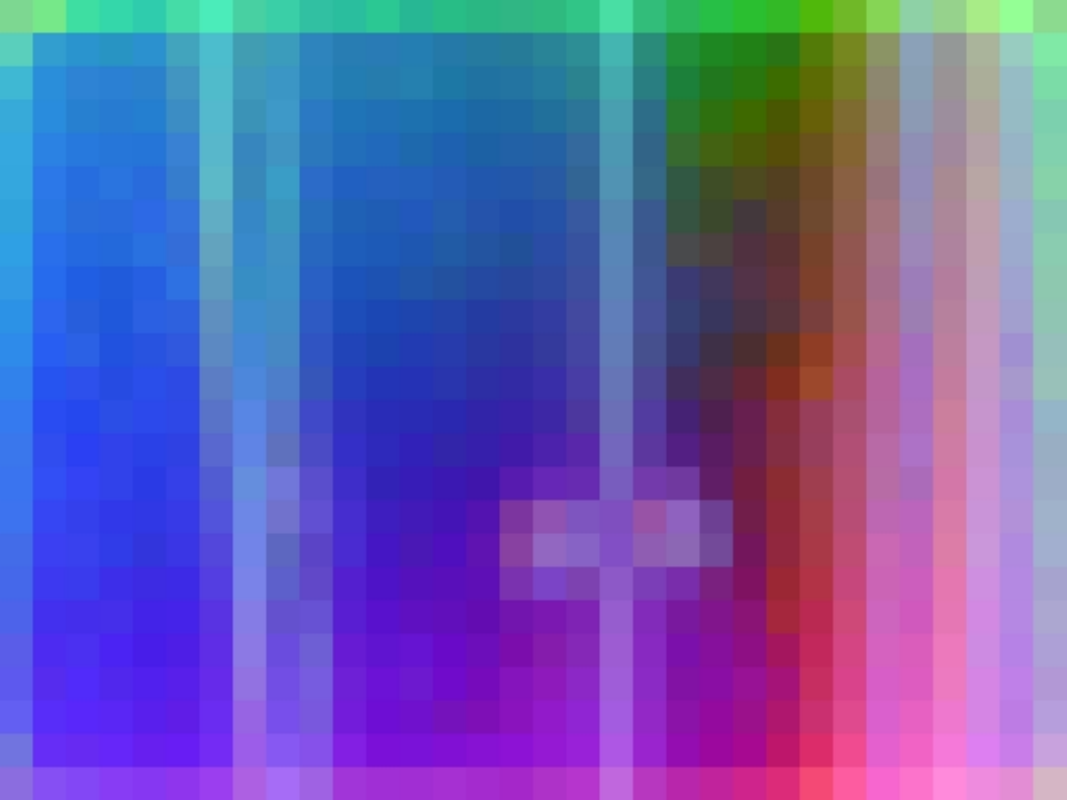} &
        \includegraphics[width=0.166\linewidth]{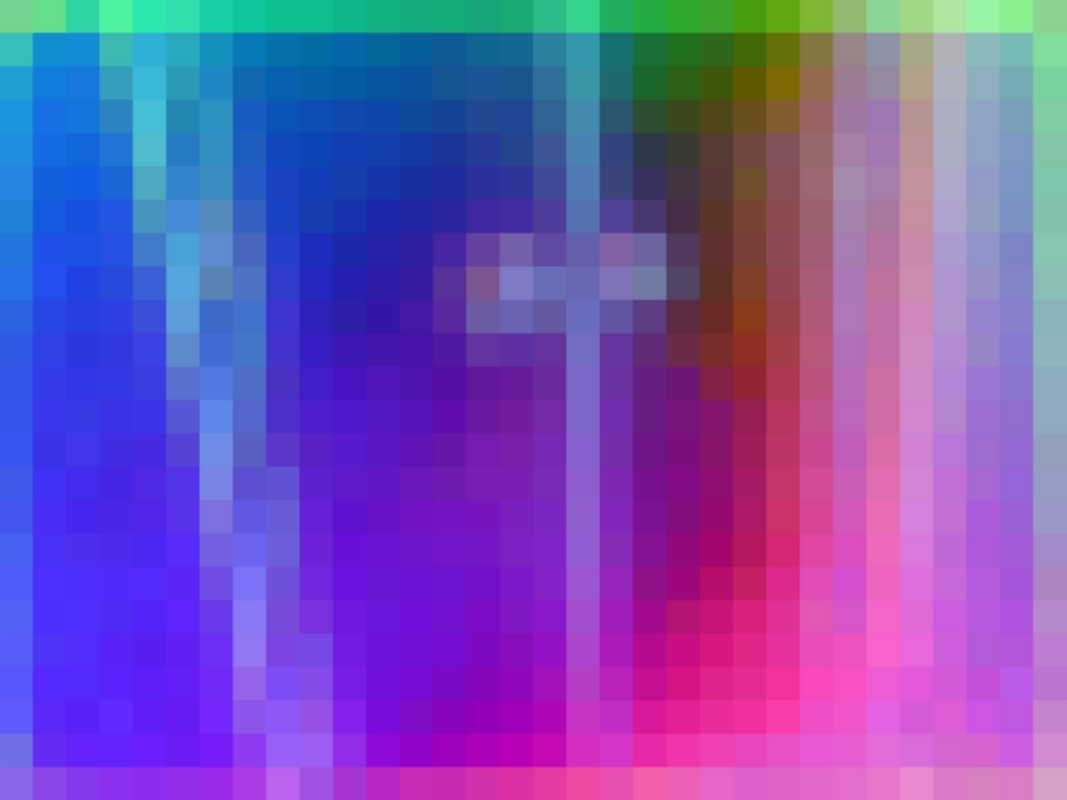} &
        \includegraphics[width=0.166\linewidth]{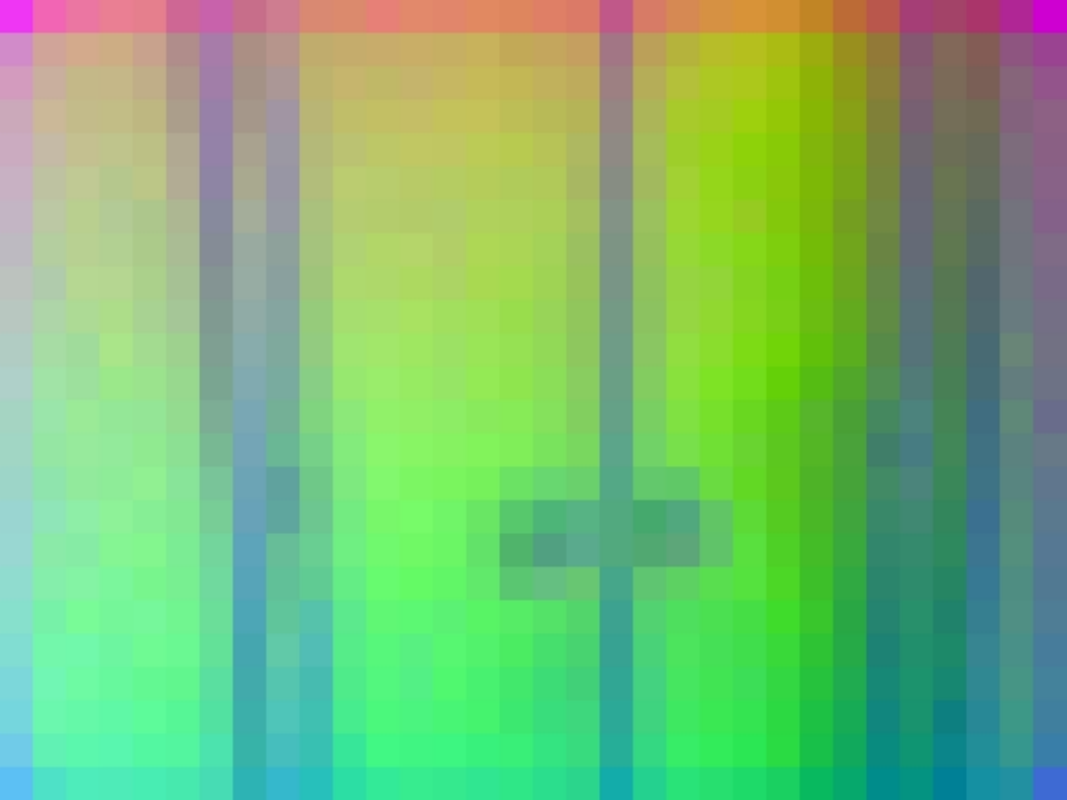} &
        \includegraphics[width=0.166\linewidth]{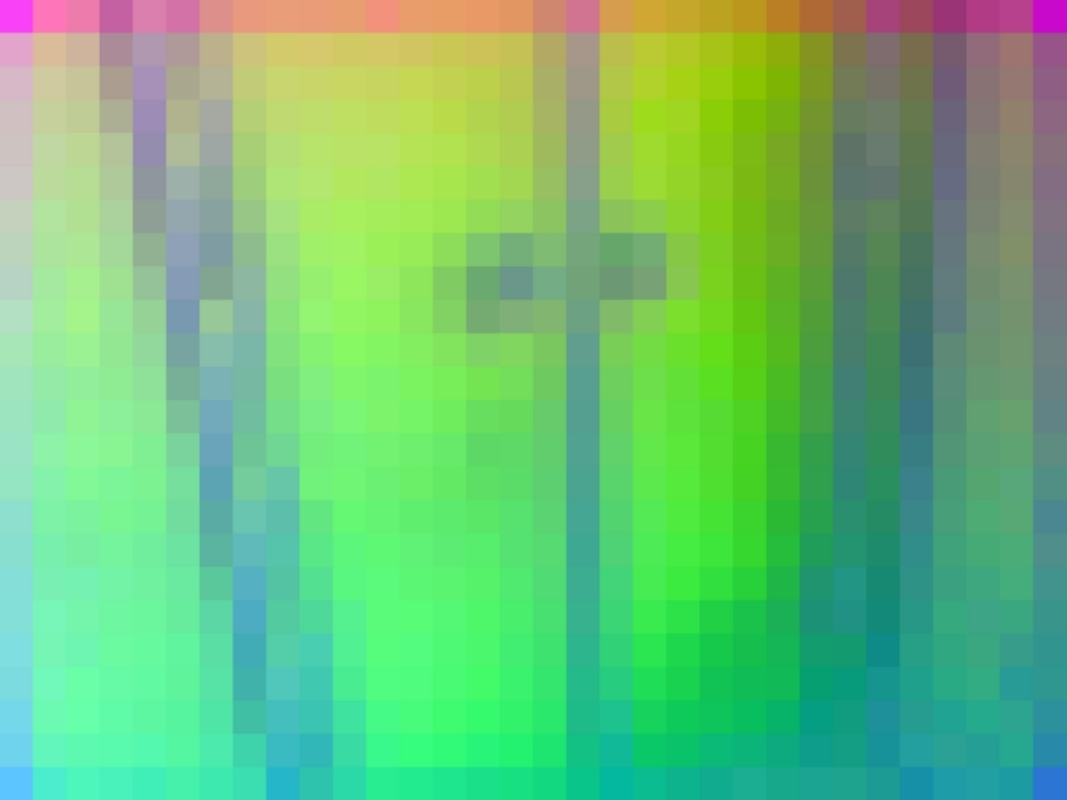} \\
    \end{tabular}
    \caption{\textbf{PCA visualization of encoder features on other domains.} Same visualization as~\Cref{fig:pca_features_1}. The first five rows are pairs from DL3DV (two outdoor scenes), Cambridge Landmarks, ETH3D and 7-Scenes; the remaining rows are ScanNet. Both models here are pre-trained on ScanNet alone, so the first five rows are out of domain for each of them, yet~\ourmethod still assigns matching colors to corresponding surfaces across the viewpoint change.}
    \label{fig:pca_features_2}
\end{figure*}

To qualitatively assess the learned representations, \Cref{fig:pca_features_1,fig:pca_features_2} visualize the first three principal components of the encoder output tokens, mapped to RGB channels, for both CroCo and~\ourmethod. The PCA is computed jointly over both views of each pair so that colors are directly comparable across images. Consistent colors in co-visible regions indicate that the encoder maps corresponding 3D points to similar feature vectors, a desirable property for downstream geometric tasks. Both models are the matched Base pair pre-trained on ScanNet-50. \Cref{fig:pca_features_1} shows ScanNet pairs; \Cref{fig:pca_features_2} adds pairs from DL3DV, Cambridge Landmarks, ETH3D and 7-Scenes, none of which either model saw during pre-training.

\section{Additional qualitative results}\label{sec:more_quali}
\begin{figure*}[t]
    \centering
    \setlength{\tabcolsep}{0pt}
    \begin{tabular}{@{}c@{}}
        \makebox[0.125\linewidth][c]{\small Target}\makebox[0.125\linewidth][c]{\small Reference}\makebox[0.125\linewidth][c]{$\ell_{\text{CroCo}}$~\eqref{eq:lossCroco}}\makebox[0.125\linewidth][c]{$\ell_{\text{conf}}$~\eqref{eq:loss_conf}}\makebox[0.125\linewidth][c]{$\ell_{\text{AI}}$~\eqref{eq:loss_rel_2}}\makebox[0.125\linewidth][c]{$\ell_{\text{AI}}$~\eqref{eq:loss_rel_2} \scriptsize{no norm.}}\makebox[0.125\linewidth][c]{$\ell_{\text{RI ratio}}$~\eqref{eq:loss_rel_3}}\makebox[0.125\linewidth][c]{$\ell_{\text{RI}}$~\eqref{eq:loss_rel} \ourmethod} \\
        \includegraphics[width=\linewidth]{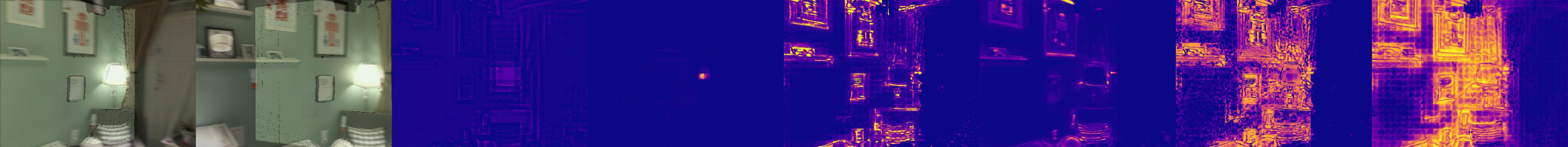} \\
        \includegraphics[width=\linewidth]{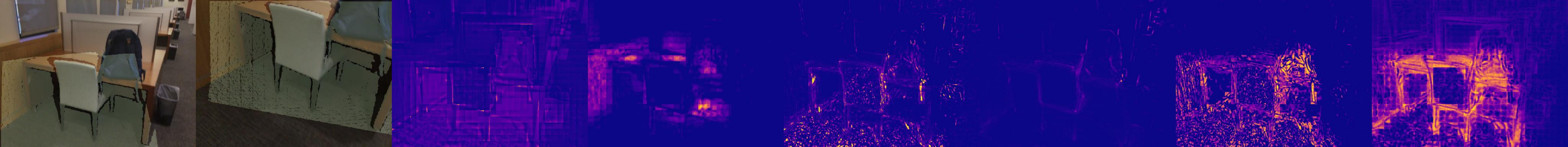} \\
        \includegraphics[width=\linewidth]{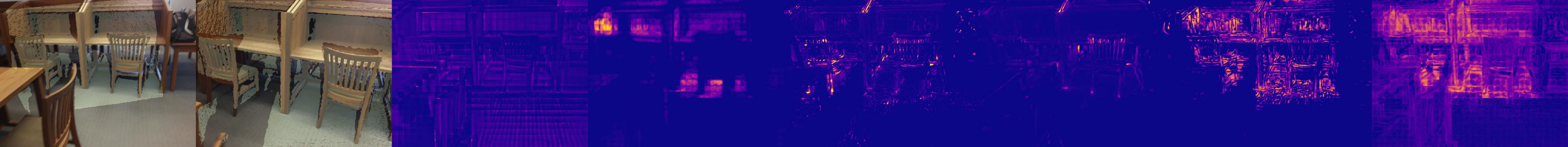} \\
        \includegraphics[width=\linewidth]{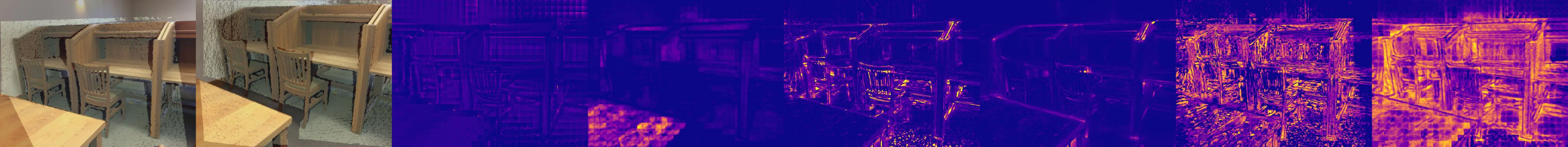} \\
        \includegraphics[width=\linewidth]{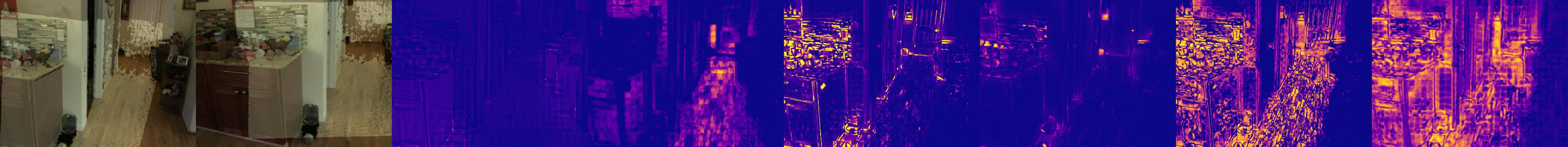} \\
        \includegraphics[width=\linewidth]{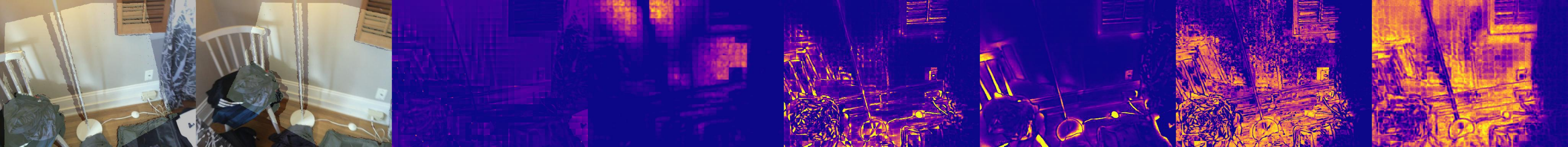} \\
        \includegraphics[width=\linewidth]{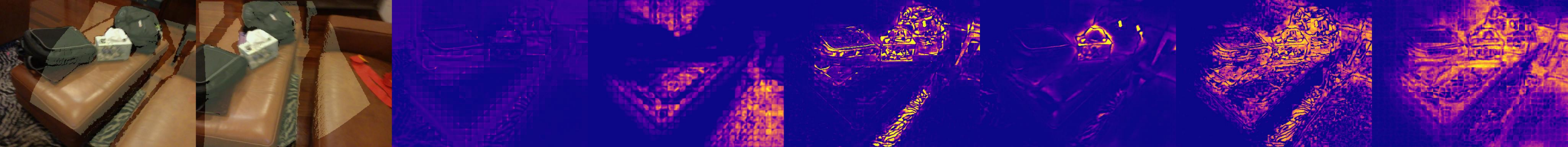} \\
        \includegraphics[width=\linewidth]{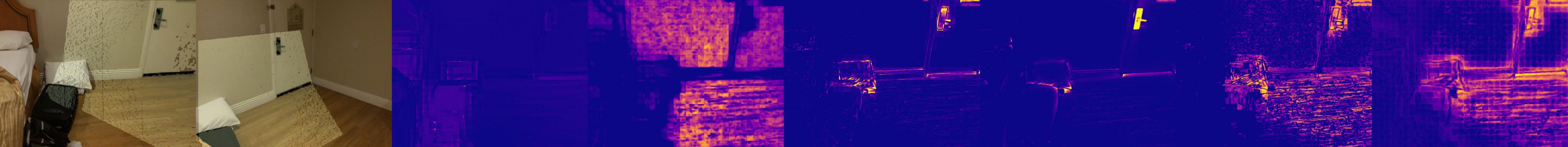} \\
        \includegraphics[width=\linewidth]{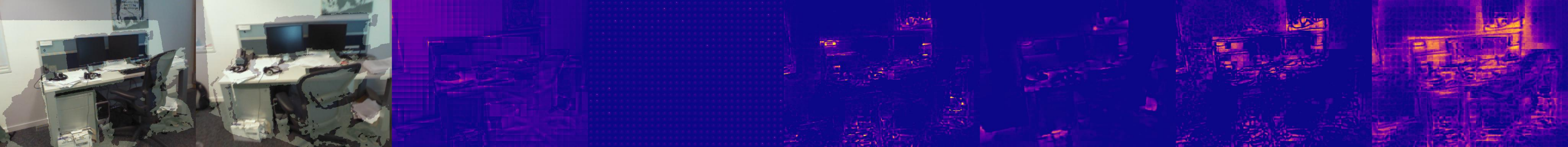} \\
        \includegraphics[width=\linewidth]{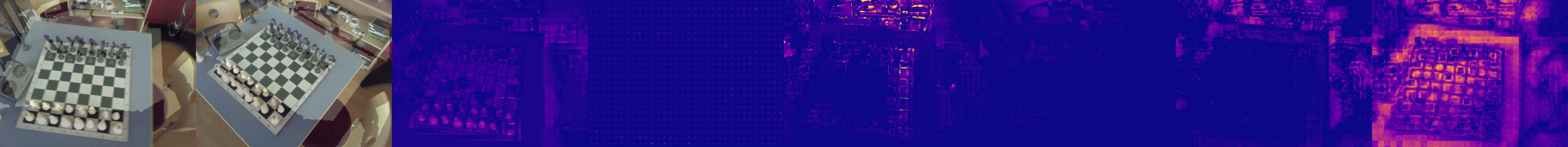} \\
        \includegraphics[width=\linewidth]{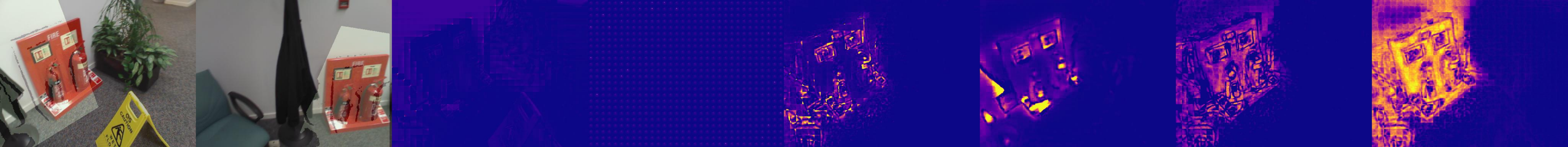}
    \end{tabular}
    \caption{\textbf{Qualitative comparison of relative improvement maps $\hat{\m{C}}$ across ablation configurations.} The relative improvement loss $\ell_{\text{RI}}$ (Eq.~\ref{eq:loss_rel}) produces sharper maps that better delineate \textcolor{mygreen}{co-visible} regions, while other configurations yield unreliable predictions.}
    \label{fig:more_heatmaps}
\end{figure*}


\end{document}